\documentclass[a4paper,10pt]{article}

\usepackage[a4paper, total={16cm, 26cm}]{geometry}
\usepackage[nottoc]{tocbibind}
\usepackage[width=.9\textwidth]{caption}
\usepackage{xspace}
\usepackage{bm}
\usepackage{amsmath}
\usepackage{amssymb}
\usepackage{longtable}
\usepackage{booktabs}
\usepackage{graphicx} 
\usepackage{textgreek}
\usepackage{iftex}
\usepackage[unicode]{hyperref}
\usepackage[utf8]{inputenc}
\usepackage{textcomp}
\usepackage{mathtools}
\usepackage{nicefrac}
\usepackage{tabularx}
\usepackage{adjustbox}
\usepackage{multirow}
\usepackage[dvipsnames,svgnames,x11names]{xcolor}
\usepackage[dvipsnames,svgnames,x11names]{colortbl}
\usepackage[linesnumbered,vlined,ruled]{algorithm2e}
\usepackage{algpseudocode}
\usepackage[english]{babel}
\usepackage{float}
\usepackage{orcidlink}
\usepackage[T1]{fontenc}
\usepackage{subcaption}
\usepackage{enumitem}

\usepackage[
backend=biber,
style=authoryear,
uniquename=true,
uniquelist=minyear,
maxcitenames=2,  % print all names if there are <= 2
]{biblatex}
\usepackage{csquotes}
\usepackage[english]{babel}

\usepackage[skip=10pt plus1pt, indent=0pt]{parskip}
\usepackage{lineno}
\nolinenumbers

\hypersetup
{
	pdftitle={Grammar as a Behavioral Biometric: Using Cognitively Motivated Grammar Models for Authorship Verification},
	pdfauthor={Andrea Nini, Oren Halvani, Lukas Graner, Sophie Titze, Valerio Gherardi and Shunichi Ishihara},
	pdfsubject={Authorship Verification},
	pdfcreator={},
	pdfproducer={},
	pdfkeywords={Authorship Verification, Forensic Linguistics, Likelihood Ratio, Cognitive Linguistics, Authorship Analysis, Stylometry, Biometrics},
	pdfpagemode=UseNone,
	pdfstartview={XYZ null null null},
	colorlinks = true,
	linkcolor  = blue,
	citecolor  = blue,
	urlcolor  = blue,
	anchorcolor = blue
}

\SetCommentSty{mycommfont} % color algorithm comments gray

\makeatletter
\newcommand*{\smallrel}[2][.8]{\mathrel{\mathpalette{\smallrel@{#1}}{#2}}}
\newcommand*{\smallrel@}[3]{\sbox0{$#2\vcenter{}$}\dimen@=\ht0  \raise\dimen@\hbox{\scalebox{#1}{\raise-\dimen@\hbox{$#2#3\m@th$}}}}
\makeatother % define a smaller variant of gtrapprox 

\newcommand*{\numberOfBaselines}   {seven\mbox{}\xspace}
\newcommand*{\numberOfEvalCorpora} {twelve\mbox{}\xspace}
\newcommand*{\ourApproach}  {\textsf{LambdaG}\mbox{}\xspace}

\newcommand*{\imOrg}        {\textsf{IM}\mbox{}\xspace}
\newcommand*{\imGen}        {\textsf{GenIM}\mbox{}\xspace}
\newcommand*{\spatium}      {\textsf{SPATIUM-L1}\mbox{}\xspace}
\newcommand*{\asgalf}       {\textsf{ASGALF}\mbox{}\xspace}
\newcommand*{\imdiff}       {\textsf{IM}$_{\text{diff}}$\mbox{}\xspace}
\newcommand*{\imrat}        {\textsf{IM}$_{\text{rat}}$\mbox{}\xspace}

\newcommand*{\unmasking}    {\textsf{Unmasking}\mbox{}\xspace}
\newcommand*{\taveer}       {\textsf{TAVeer}\mbox{}\xspace}

\newcommand*{\coav}         {\textsf{COAV}\mbox{}\xspace}

\newcommand*{\janithDiffVec}{\textsf{FeVecDiff}\mbox{}\xspace}
\newcommand*{\janithDiffVecLR}{\textsf{FeVecDiff (LR)}\mbox{}\xspace}
\newcommand*{\janithDiffVecMLP}{\textsf{FeVecDiff (MLP)}\mbox{}\xspace}
\newcommand*{\janithLR}     {\textsf{LR}\mbox{}\xspace}
\newcommand*{\janithMLP}    {\textsf{MLP}\mbox{}\xspace}
\newcommand*{\adhominem}    {\textsf{ADHOMINEM}\mbox{}\xspace}
\newcommand*{\adhominemOrg} {\textsf{ADHOMINEM}$^{\,(\bm{\dagger})}$\mbox{}\xspace}
\newcommand*{\hrsn}         {\textsf{HRSN}\mbox{}\xspace}  % Hierarchical recurrent Siamese neural network (HRSN).
\newcommand*{\luarAV}       {\textsf{LUAR}\mbox{}\xspace}

\newcommand*{\luarAVOrg}    {\textsf{LUAR}$^{\,(\bm{\dagger})}$\mbox{}\xspace}

\newcommand*{\bibertAV}     {\textsf{BiBERT-AV}\mbox{}\xspace}

\newcommand*{\A}            {\mathcal{A}}
\newcommand*{\B}            {\mathcal{B}}
\newcommand*{\notA}         {\neg\A}
\newcommand*{\unknown}      {\mathcal{U}}

\newcommand*{\D}            {\mathcal{D}}
\newcommand*{\DA}           {\D_{\A}}

\newcommand*{\Dref}         {\mathbb{D}_{\text{ref}}}
\newcommand*{\Dunk}         {\D_{\unknown}}

\newcommand*{\Dset}         {\mathbb{D}}

\newcommand*{\Arefset}      {\Dset_{\A}}
\newcommand*{\Brefset}      {\Dset_{\B}}

\newcommand*{\Threshold}    {\ensuremath{\theta}\xspace}
\newcommand*{\TFIDF}        {\emph{tf-idf}\mbox{}\xspace}
\newcommand*{\featVector}   {\ensuremath{\mathcal{F}}\mbox{}\xspace}
\newcommand*{\numberRepetitions}  {\ensuremath{r}\xspace}
\newcommand*{\modelOrder}  {\ensuremath{N}\xspace}

\newcommand*{\avBatch}   {\textrm{AV}_{\,\textrm{Batch}}}
\newcommand*{\avPAN}     {\textrm{AV}_{\,\textrm{Known}}}
\newcommand*{\avCore}    {\textrm{AV}_{\,\textrm{Core}}}

\newcommand*{\posNoise}       {\textsf{POSNoise}\mbox{}\xspace}
\newcommand*{\posNoiseBold}   {\textsf{POSNoise}\mbox{}\xspace} % Nothing works to make math symbols bold :-(   https://tex.stackexchange.com/questions/303568/sans-serif-bold-math-with-newtxmath
\newcommand*{\textDistortion} {\textsf{TextDistortion}\mbox{}\xspace}

\newcommand*{\Corpus}                 {\mathcal{C}}
\newcommand*{\CorpusTrain}            {\mathcal{C_{\,\textrm{train}}}}
\newcommand*{\CorpusTrainStar}        {\mathcal{C_{\,\textrm{train}*}}}
\newcommand*{\CorpusValidation}       {\mathcal{C_{\,\textrm{val}}}}
\newcommand*{\CorpusTest}             {\mathcal{C_{\,\textrm{test}}}}

\newcommand*{\CorpusEnron}            {\Corpus_\mathrm{Enron}}
\newcommand*{\CorpusStack}            {\Corpus_\mathrm{Stack}}
\newcommand*{\CorpusWiki}             {\Corpus_\mathrm{Wiki}}
\newcommand*{\CorpusACL}              {\Corpus_\mathrm{ACL}}
\newcommand*{\CorpusPJ}               {\Corpus_\mathrm{PJ}}
\newcommand*{\CorpusApricity}         {\Corpus_\mathrm{Apric}}
\newcommand*{\CorpusTripAdvisor}      {\Corpus_\mathrm{Trip}}
\newcommand*{\CorpusYelp}             {\Corpus_\mathrm{Yelp}}

\newcommand*{\CorpusIMDB}             {\Corpus_\mathrm{IMDB}}
\newcommand*{\CorpusBlogs}            {\Corpus_\mathrm{Blog}}
\newcommand*{\CorpusAmazon}           {\Corpus_\mathrm{Amazon}}
\newcommand*{\CorpusAllTheNews}       {\Corpus_\mathrm{AllNews}}

\renewcommand*{\AA}                   {AA\mbox{}\xspace}

\newcommand*{\AV}                     {AV\mbox{}\xspace}

\newcommand*{\Problem}                {\ensuremath{c}\xspace} 
\newcommand*{\ProblemNew}             {\ensuremath{\Problem_{\textrm{new}}}\xspace}

\newcommand*{\classY}                 {\texttt{Y}\mbox{}\xspace} 
\newcommand*{\classN}                 {\texttt{N}\mbox{}\xspace} 
\newcommand*{\classYdash}[1]          {\texttt{Y-}#1\mbox{}\xspace} 
\newcommand*{\classNdash}[1]          {\texttt{N-}#1\mbox{}\xspace}

\newcommand*{\auc}        {AUC\mbox{}\xspace}
\newcommand*{\fOne}       {\ensuremath{F_{\text{1}}}\mbox{}\xspace}
\newcommand*{\cllr}       {\ensuremath{C_{\text{llr}}}\mbox{}\xspace}
\newcommand*{\cllrmin}    {\ensuremath{C_{\text{llr}}^{\text{min}}}\mbox{}\xspace}
\newcommand*{\cllrcal}    {\ensuremath{C_{\text{llr}}^{\text{cal}}}\mbox{}\xspace}

\newcommand*{\sota}           {state-of-the-art\mbox{}\xspace}

\newcommand*{\posTag}         {POS tag\mbox{}\xspace}
\newcommand*{\posTags}        {POS tags\mbox{}\xspace}

\newcommand*{\ngram}          {$n$\texttt{-}gram\mbox{}\xspace}
\newcommand*{\ngrams}         {$n$\texttt{-}grams\mbox{}\xspace}

\newcommand*{\charNgrams}     {character $n$\texttt{-}grams\mbox{}\xspace}

\newcommand*{\token}          {\ensuremath{t}\mbox{}}
\newcommand*{\sentence}       {\ensuremath{S}\mbox{}}

\newcommand*{\SentenceSetUnk} {\ensuremath{\mathbb{S}_{\unknown}}\mbox{}}
\newcommand*{\SentenceSetA}   {\ensuremath{\mathbb{S}_{\A}}\mbox{}}
\newcommand*{\SentenceSetRef} {\ensuremath{\mathbb{S}_{\textsf{ref}}}\mbox{}}
\newcommand*{\vocabulary}     {\ensuremath{\mathcal{T}}\mbox{}}

\newcommand*{\ppm}            {\textsf{PPM}\mbox{}\xspace} %  Prediction by Partial Matching

\newcommand*{\eg}             {e.\,g.,\mbox{}\xspace}
\newcommand*{\ie}             {i.\,e.,\mbox{}\xspace}
\newcommand*{\vs}             {versus\mbox{}\xspace} 
\newcommand{\e}[1]{\emph{#1}} 

\renewcommand                {\quote}[1]{{``#1''}}
\newcommand                  {\quotetxt}[1]{{``\texttt{#1}''}}

\begin{document}
\title{\textbf{Grammar as a behavioral biometric: Using cognitively motivated grammar models for authorship verification}}
\author{
%-------------------------------------------------------------------------
\textbf{Andrea Nini}~\orcidlink{0000-0003-4218-5130}\\Linguistics and English Language\\University of Manchester\\Manchester,\ M13 9PL\\\href{mailto:andrea.nini@manchester.ac.uk}{Andrea.Nini@Manchester.ac.uk}
%-------------------------------------------------------------------------
\\\\\textbf{Oren Halvani}~\orcidlink{0000-0002-1460-9373}\\Fraunhofer Institute for Secure Information Technology SIT\\Darmstadt, 64295\\\href{mailto:oren.halvani@sit.fraunhofer.de}{Oren.Halvani@SIT.Fraunhofer.de}
%-------------------------------------------------------------------------
\\\\\textbf{Lukas Graner}~\orcidlink{0000-0002-0453-1146}\\Fraunhofer Institute for Secure Information Technology SIT\\Darmstadt, 64295\\\href{mailto:lukas.graner@sit.fraunhofer.de}{Lukas.Graner@SIT.Fraunhofer.de}
%-------------------------------------------------------------------------
\\\\\textbf{Sophie Titze}~\orcidlink{0009-0008-4034-7048}\\Fraunhofer Institute for Secure Information Technology SIT\\Darmstadt, 64295\\\href{mailto:sophie.titze@sit.fraunhofer.de}{Sophie.Titze@SIT.Fraunhofer.de}
%-------------------------------------------------------------------------
\\\\\textbf{Valerio Gherardi}~\orcidlink{0000-0002-8215-3013}\\Cavanilles Institute of Biodiversity and Evolutionary Biology \\ University of Valencia \\C/Catedrático José Beltrán 2, 46980 Paterna, Valencia, Spain\\\href{mailto:vgherard840@gmail.com}{VGherard840@GMail.com}
%-------------------------------------------------------------------------
\\\\\textbf{Shunichi Ishihara}~\orcidlink{0000-0001-6633-3316}\\Speech and Language Laboratory\\ANU College of Asia and the Pacific\\The Australian National University\\\href{mailto:Shunichi.Ishihara@anu.edu.au}{Shunichi.Ishihara@ANU.edu.au}
%-------------------------------------------------------------------------
}

\maketitle
\begin{abstract}
Authorship Verification (AV) is a key area of research in digital text forensics, which addresses the fundamental question of whether two texts were written by the same person. Numerous computational approaches have been proposed over the last two decades in an attempt to address this challenge. However, existing AV methods often suffer from high complexity, low explainability and especially from a lack of clear scientific justification. We propose a simpler method based on modeling the grammar of an author following Cognitive Linguistics principles. These models are used to calculate $\lambda_G$ (\ourApproach): the ratio of the likelihoods of a document given the candidate’s grammar \vs given a reference population's grammar. Our empirical evaluation, conducted on \numberOfEvalCorpora datasets and compared against \numberOfBaselines baseline methods, demonstrates that \ourApproach achieves superior performance, including against several neural network-based AV methods. \ourApproach is also robust to small variations in the composition of the reference population and provides interpretable visualizations, enhancing its explainability. We argue that its effectiveness is due to the method’s compatibility with Cognitive Linguistics theories predicting that a person’s grammar is a behavioral biometric.
\end{abstract}

\section{Introduction}\label{sec-introduction}
Texts serve numerous purposes and can be found in a variety of digital and non-digital forms such as books, social media posts, emails, letters, websites, academic papers, blogs, product reviews, poems and many others. These text types, in turn, can be categorized according to various criteria including language, genre, topic, readability or authorship. The latter is particularly important when it comes to investigating  questioned documents, for example, in forensic settings, \eg blackmail letters, suicide notes, letters of confession, wills or theses potentially written by ghostwriters. 
\\\\
Over time, various research disciplines have emerged that focus on analyzing the language of authors from different perspectives which has led to an interdisciplinary field of research often referred to as \e{Digital Text Forensics} (\eg \cite{DigitalTextForensicsECIR:2019,BevendorffBiasAV:2019,DigitalTextForensicsTrends:2013,DigitalTextForensicsPAN:2019,CyberstalkingDetection:2016}). Two well-known and closely related sub-areas of this branch of research are \textbf{Authorship Attribution} (\AA) and \textbf{Authorship Verification} (\AV). \AA is concerned with the task of attributing an anonymous text to the most likely author based on a set of sample documents of known authors. \AV, on the other hand, deals (in its simplest form) with the task of deciding whether two documents were written by the same author. In this paper, we focus on \AV, which can essentially be regarded as a reformulation of \AA, given the fact that any \AA problem can be decomposed into a series of \AV problems \parencite{KoppelFundamentalProblemAA:2012}.

\subsection{Decision Problems} \label{DecisionProblems}
According to \textcite{SteinMetaAnalysisAV:2008}, \textcite{PANOverviewAV:2013}, \textcite{KoppelFundamentalProblemAA:2012} and \textcite{HalvaniPhD:2021} the following decision problems can be answered by \AV:
\begin{enumerate}
    \item $\avCore$: Given two documents $\D_1$ and $\D_2$: Were both documents written by the same author?
    
    \item $\avBatch$: Given two sets of documents $\Arefset = \{ \D^{\,\A}_1, \D^{\,\A}_2, \ldots \}$ and $\Brefset = \{ \D^{\,\mathcal{B}}_1, \D^{\,\mathcal{B}}_2, \ldots\}$ each of which written by a single author: Were the documents in $\Arefset$ and $\Brefset$ written by the same author?
    
    \item $\avPAN$: Given a set of documents $\Arefset = \{ \D_1, \D_2, \ldots \}$ by the same author $\A$ as well as a document $\Dunk$ of an unknown author $\unknown$: Has $\A$ also written $\Dunk$ (\ie $\A = \unknown$)? 
\end{enumerate}
%---------------------------------------------------- 
In a forensic context, the process of assisting the trier-of-facts in finding answers to these problems is referred to as \textit{Forensic Text Comparison} (FTC) as described by \textcite{ishihara2021}. The focus in this paper is on all three of these decisions problems.
%---------------------------------------------------- 
% AV Notation
%---------------------------------------------------- 
For the rest of this paper, we adopt the following notation: a document denoted by $\Dunk$, represents a document of an \textbf{unknown author} $\unknown$, whereas a document denoted by $\DA$ represents a document of a \textbf{known author} $\A$. In all decision problems, including $\avCore$ where the denotation of which document is known or unknown can be arbitrary, an \AV method seeks to answer the question of whether $\A = \unknown$ ($\A$ authored $\Dunk$) or $\A \neq \unknown$ ($\A$ did not author $\Dunk$) holds. In this context, a \textbf{known} verification case with matching and non-matching authorship is denoted by \classYdash{} and \classNdash{case}, respectively. Moreover, we denote $\notA$ as any author other than $\A$. 

\subsection{Applications of AV}
%----------------------------------------------------------------------------------------------------------- 
\AV can be used in many different areas and offers a wide range of possible applications. In forensic linguistics, \AV plays an important role in the analysis of texts related to criminal activities such as threatening letters, ransom notes, blackmail letters, terrorist manifestos or other documents used as evidence in investigative, criminal or civil proceedings. For example, \textcite{IqbalAVForensicInvestigation:2010} used a NIST-based speaker recognition system to verify the authorship of malicious emails in forensic contexts. Similarly, \textcite{RemmideAVPhishingEmailDetection:2024} employed \AV to detect phishing attacks. 
%----------------------------------------------------------------------------------------------------------- 
In journalism and media, \AV plays a vital role in authenticating the authorship of news articles and opinion pieces, as well as detecting ghostwriting \parencite{AlKhatibAVOpinionArticlesNewspapersIdiolect:2021}.  
%----------------------------------------------------------------------------------------------------------- 
In academia, \AV is crucial for detecting academic misconduct, such as contract cheating \parencite{StavngaardAVGhostwriting:2019} and plagiarism \parencite{EnriquezAVHiredPlagiarismDetection:2023}. 
%----------------------------------------------------------------------------------------------------------- 
Social media platforms also benefit from \AV, especially when it comes to detecting fake accounts or multiple accounts operated by the same person. In this context, \textcite{IgawaAVCompromisedAccountsTwitter:2015} and \textcite{BarbonAVSocialNetworks:2017} demonstrated the effectiveness of \AV regarding the identification of compromised accounts on X/Twitter. \textcite{LundmarkAVSwedishViaAdhominem:2020}, on the other hand, used AV to link user accounts on discussion forums, whereas \textcite{WeerasingheMitigateAbuseOnlineCommunities:2022} applied \AV to mitigate abuse in online communities.
%----------------------------------------------------------------------------------------------------------- 
Finally, in the \textit{Digital Humanities}, \AV has also become an important instrument for clarifying questions about the authorship of historical documents and authenticating literary works. \textcite{KoppelAuthenticatingHistoricalWritings:2016}, for example, applied \AV to authenticate historical writings attributed to Julius Caesar, while \textcite{TuccinardiAuthenticityPlinyTrajan:2017} used \AV methods to investigate the authenticity of Pliny the Younger’s letter to Emperor Trajan. %-----------------------------------------------------------------------------------------------------------

\subsection{Categories of AV Methods}
\AV methods can be divided into three different categories: \textbf{unary}, \textbf{binary-intrinsic} and \textbf{binary-extrinsic} \parencite{HalvaniPAN:2020,HalvaniAssessingAVMethods:2019}. All three rely on a decision criterion $\Threshold$ in order to accept or reject $\A$ as the author of $\Dunk$. The way in which $\Threshold$ is determined is what primarily distinguishes these three categories:
\begin{itemize}
    \item \textbf{Unary} \AV methods rely solely on a set of documents $\Arefset = \{ \D_1, \D_2, \ldots \}$ of a known author $\A$ in order to determine $\Threshold$. These methods assume $\Dunk$ to be written by $\A$, if it is similar enough to the documents in $\Arefset$, where the decision is only based on these and $\Threshold$.

    \item \textbf{Binary-intrinsic} \AV methods require a given training corpus, consisting of verification cases with a ground truth regarding the \emph{same-authorship} (\classY) and \emph{different-authorship} (\classN) classes, in order to determine $\Threshold$. Binary-intrinsic \AV methods treat verification cases that include the unknown and known document(s) of $\A$, as a single unit $X$ (\eg a feature vector). If $X$ is more similar to the \classYdash{cases}, the method accepts $\A$ as the author of $\Dunk$. On the other hand, if $X$ is more similar to the \classNdash{cases}, $\Dunk$ is assumed to be written by another author. In any case, the decision is made solely on the basis of $X$ and the trained classification model that incorporates $\Threshold$. 
    
    \item \textbf{Binary-extrinsic} \AV methods require external, so-called \e{impostor} \parencite{KoppelWinter2DocsBy1:2014} documents (which have not been written by $\A$) in order to determine $\Threshold$. To obtain impostor documents $\mathbb{I} = \{ I_1, I_2, \ldots \}$, different strategies can be employed. \textcite{KoppelWinter2DocsBy1:2014}, for example, used two approaches: (1) collecting documents using search engine queries and (2) using a fixed set of already crawled documents. Binary-extrinsic \AV methods assume $\Dunk$ to be written by $\A$ only if $\Dunk$ is consistently more similar to $\DA$ (or alternatively, the documents in $\Arefset$) than any of a number of distractor impostor documents $\{ I_1, I_2, \ldots \}$ using a bootstrapping approach.
\end{itemize}

In a forensic context, the adherence to the \textit{likelihood ratio paradigm} for forensic science implies that an \AV method does not return a \classYdash{} or \classNdash{decision} but a likelihood ratio that quantifies the strength of the evidence in favor or against the two propositions. In addition to the three types above, therefore, it is also possible to classify an \AV method as \textit{forensic} when the method returns a likelihood ratio.

\subsection{Existing AV Solutions}

This section describes the advantages and limitations of some well-known existing \AV methods that have demonstrated a consistent degree of effectiveness over time. In particular, this section covers those methods that have found significant validation across the yearly competitions held at the PAN series of conferences focused on stylometry and digital text forensics. Since its launch in 2007, PAN has hosted 77 shared tasks\footnote{Past PAN shared tasks: \url{pan.webis.de/shared-tasks.html}}, in areas such as authorship analysis, computational ethics and evaluating the originality of writing, as well as curated 60 evaluation datasets\footnote{PAN datasets: \url{pan.webis.de/data.html}}, in addition to nine datasets contributed by the community \parencite{PANOverallOverview:2025}.

\subsubsection{The Impostors Method}

In a review of the field, \textcite{stamatatos_authorship_2016} concluded that binary-extrinsic methods tend to outperform other types of methods. A widely recognized and successful binary extrinsic \AV method that has been repeatedly validated in independent studies is the \e{Impostors Method} (\imOrg), which has been proposed by \textcite{KoppelWinter2DocsBy1:2014}. The basic idea behind \imOrg can be traced back to another successful \AV method called \unmasking, initially introduced by \textcite{KoppelAVOneClassClassification:2004} and later refined by \textcite{KoppelUnmasking:2007}. Given a verification case $\Problem = (\Dunk, \DA)$, the key insight of \imOrg is to collect a set of texts that are plausible impostors for $\DA$ and then calculate a similarity score for $\Dunk$ and $\DA$ not once but a large number of times while withholding a random sub-set of the features at every iteration. The rationale is that only the real author of the disputed text $\Dunk$ is so similar to it to withstand random permutations of the features set, which, for \imOrg, is the set of the most frequent character $4$\texttt{-}grams. \textcite{KoppelFundamentalProblemAA:2012} suggested that the key principle to choose the impostors should be to identify a relevant population, select linguistically similar texts, and then randomly sample from it. 

\imOrg found immediate external validation in the PAN 2013 competition, where a variant called \imGen (\e{General Impostors Method}), proposed by \textcite{SeidmanPAN13:2013}, outperformed all other submitted \AV methods. As suggested by \textcite{KoppelInTheWild:2011}, \imGen made use of collected impostors document from the web. The method then used the relative frequency of single words weighted by their document frequency (a quantity usually called \TFIDF) and the \e{Min-Max} (aka \e{Ru\v{z}i\v{c}ka} or fuzzy Jaccard) similarity measure. Successful results were then found by \textcite{KoppelWinter2DocsBy1:2014} for the $\avCore$ decision problem, again using \e{Min-Max} applied to vectors of character $4$\texttt{-}grams weighted by \TFIDF. They explored various configurations for impostors, including dynamically generating them through web searches or sampling from genre-compatible corpora. Koppel and Winter discovered that genre compatibility permits the use of fewer impostors but that, equally, a large number of impostors ($\approx 200$) generated through web searches can compensate for the absence of genre control. Their results demonstrated that it is possible to solve this challenging two-text only verification problem even for texts as short as 500 words.

After 2014, \imOrg saw the introduction of new variants that perfected other aspects of the original algorithm, with increasing degrees of success: 
\asgalf \parencite{KhonjiIraqiAV:2014}, which won the PAN competition in 2014, \spatium \parencite{KocherPANSpatium:2015,KocherSavoySpatiumL1:2017},
Kestemont's et al. variant \parencite{KoppelAuthenticatingHistoricalWritings:2016},
the \emph{Rank-Based Impostors} \parencite{StamatatosPothaImprovedIM:2017},
the \emph{Profile-Based Impostors} \parencite{PothaStamatatosExtrinsicAV:2019} and finally \imdiff and \imrat \parencite{khonji2022}. The success of \imOrg was also confirmed by two comparative review studies on \AV methods. \textcite{moreau2022} concluded that \imOrg outperformed all tested methods, including the meta-classifier. Similarly, \textcite{HalvaniPhD:2021} found that \imOrg ranked first or second in most cases across a series of robustness, topic masking and generalization experiments.

Although \imOrg's performance over the years is unquestionable, the four main drawbacks of the method are that it is not deterministic, that it suffers from a high run-time \parencite{HalvaniPhD:2021}, that it relies on choosing impostors from the same genre and that its results are often hard to interpret. The first issue is typically addressed by running several iterations and averaging the results but this additional procedure only leads to even longer run-times. The third issue is not strictly speaking an issue of \imOrg but a well-known problem for both \AV and \AA, as virtually all methods to solve these tasks tend to be highly affected by genre variation \parencite{argamon_computational_2018}. For \imOrg, \textcite{HalvaniPhD:2021} found that the performance (in terms of Accuracy\footnote{Note that we use the capitalized form "Accuracy" to clarify that we are talking about a specific machine learning evaluation metric, rather than the general concept of "accuracy".}) declines significantly when impostors are drawn from a different genre. For example, performing \AV on a corpus of chat logs achieves an Accuracy of 0.92 when using chat logs as impostors. However, when using business emails or academic articles as impostors, this performance drops to almost chance level at 0.54. Even with relatively similar genres, such as chat logs and blogs, the performance still decreases notably, though it remains above chance (0.74).

The last and most critical limitation of \imOrg and of all other \AV methods using the frequency of short \charNgrams is that the most important features of the analysis are not immediately obvious to the human analyst. In addition, unless some kind of content masking is implemented, the use of short \charNgrams can also lead to the contamination of the data from topic or content noise and other unwanted idiosyncrasies that could influence the results. For example, a character $4$\texttt{-}gram like \quotetxt{f th} is likely to capture grammatical information, such as the frequencies of grammatical sequences like \quotetxt{if the} or \quotetxt{of the} or \quotetxt{if that} but also information that has to do with the topic of a text, like \quotetxt{elf theme} or \quotetxt{grief therapy}. The danger of capturing content is that topics might correlate to authors and, especially in forensic settings, the \AV method could, for example, return a positive verification outcome only because in the corpus used for training there is a correlation between content and authors (\eg $\A$ is likely to be the author of $\Dunk$ because $\DA$ and $\Dunk$ are about \quotetxt{grief therapy}).

\subsubsection{Content-agnostic Approaches}
This content effect limitation of \imOrg is also one of a number of biases identified by \textcite{BevendorffBiasAV:2019} on the basis of the 2014 and 2015 PAN competitions. These biases were taken in consideration by PAN starting from its 2020 edition \parencite{PANOverviewAV:2020}. However, the problem of using features that can potentially capture topic or content instead of authorship remains in many verification systems, including more recent ones. 

One of the first approaches dealing with content bias in the context of authorship analysis was the topic-masking technique \textDistortion \parencite{StamatatosTextDistortion:2017}, which is simply based on replacing any word\footnote{As an alternative, \textcite{StamatatosTextDistortion:2017} proposed an additional variant of replacing not the entire word $w$ but instead each character in it with an asterisk.} $w$ that is not among the top most frequent words in a reference corpus with an asterisk. \textDistortion was found to actually improve the results in cross-topic and cross-genre analyses. In 2021, \textcite{HalvaniPOSNoise:2021} tested \textDistortion against another algorithm they introduced called \posNoise (see Section~\ref{sec-the-posnoise-algorithm}), which is a preprocessing method that consists in the replacement of words that are unlikely to be functional with their Part-Of-Speech (POS). They found that this method clearly improves the algorithms in cross-domain cases while often resulting in poorer performance in cases that are not cross-domain. These results are interpreted by them as evidence that this algorithm should be applied for forensic purposes because it is capable of removing the effect of content. Among the methods that benefited from this preprocessing was \imOrg, which was also one of the most successful ones in their evaluations.

As an alternative to these rule-based content removal methods, \textcite{BischoffSuppressingDomainStyleAA:2020} proposed a solution using a neural network trained to differentiate content from style. In their study, they confirmed that the frequency of short unprocessed \charNgrams is highly affected by content by analyzing a large corpus of cross-domain problems where they found that methods based on these features perform at random. Despite the success of their own neural network method, they discovered that a simpler approach such as \textDistortion works equally well at a fraction of the computational and data cost of a neural network.

%-------------------------------
% TAVeer
%-------------------------------
Another approach to counteract content bias, called \taveer\footnote{\taveer stands for \emph{\textbf{T}opic-agnostic \textbf{A}uthorship \textbf{V}erifier based on \textbf{e}qual \textbf{e}rror \textbf{r}ate}\cite{HalvaniARES:2020}.}, was proposed by \textcite{HalvaniARES:2020}. In contrast to \textDistortion and \posNoise, \taveer represents a stand-alone \AV method based on ensembles of topic-agnostic (TA) feature categories (namely punctuation \ngrams, TA sentence and sentence-initial tokens, TA sentence endings, TA token \ngrams and TA masked token \ngrams). The TA feature categories themselves cover a total of $\approx 1000$ words and phrases that originate from 20 linguistic categories such as punctuation marks, conjunctions, determiners, prepositions, contractions, auxiliary verbs, transition words and others, which are known in the literature \parencite{PavelecAAConjunctionsAndAdverbs:2008,BinongoAABookOfOz:2003,StolermanPhD:2015} to be content and topic independent. Given a verification case $\Problem = (\Dunk, \DA)$ and a set of selected feature categories, \taveer constructs two feature vectors for each category, which represent the $L^1$-normed occurrence frequencies of specific features (such as the conjunction word \quotetxt{and}) in the corresponding document. Afterwards, for each category, the Manhattan distance between two vectors is calculated and calibrated into the interval $[0,1]$, where $0.5$ represents a decision boundary. The median of the calibrated values over all selected categories, compared to $0.5$ then represents the final decision (\ie $\A = \unknown$ or $\A \neq \unknown$). The thresholds and the optimal subset of selected TA feature categories are optimized, by selecting the ones that yield the best accuracy on a given training corpus. 

\subsubsection{Language Models and Compression Methods} \label{sec-languagemodels-compression}
In addition to \imOrg and its variants, the most successful family of methods in \AV are the ones that are based on language modeling and/or compression, which are mathematically related. Here, a language model is defined as a generative statistical model that assigns a probability distribution to stretches of language. 

One of the very first successful examples for this class of methods was the winner of the PAN 2015 competition, \textcite{BagnallRNN:2015}, whose \AV method was based on a character-level recurrent neural network language model. Another noteworthy neural language modeling approach, based on a Siamese neural network, is \hrsn\footnote{\hrsn stands for \emph{\textbf{H}ierarchical \textbf{R}ecurrent \textbf{S}iamese \textbf{N}etwork} \parencite{BoenninghoffSimilarityLearningAV:2019}.}, proposed by \textcite{BoenninghoffSimilarityLearningAV:2019}. This approach was found to outperform \imOrg by reducing its error rate by half. However, as noted by \textcite{BoenninghoffExplainableAV:2019}, it has two key drawbacks: first, it does not utilize linguistically interpretable stylometric features, and second, it lacks a mechanism for directly visualizing the decision-making process. Additionally, \hrsn  is constrained by its reliance on a large and well-structured comparison dataset, which may not always be available. 

Another well-known \AV method that can be regarded as an improved version of \hrsn is \adhominem\footnote{\adhominem stands for \emph{\textbf{A}ttention-based \textbf{D}eep \textbf{H}ierarchical c\textbf{O}nvolutional sia\textbf{M}ese b\textbf{I}directional recurre\textbf{N}t n\textbf{E}ural-network \textbf{M}odel} \parencite{BoenninghoffExplainableAV:2019}.}, which was also proposed by \textcite{BoenninghoffExplainableAV:2019}.
\adhominem achieved the highest ranking in the PAN \AV competition 2020 \parencite{PANOverviewAV:2020} and, according to \textcite{LundmarkAVSwedishViaAdhominem:2020}, has shown great promise with respect to real \AV applications. The method can be summarized as follows: Given a verification case $\Problem = (\Dunk, \DA)$, \adhominem preprocesses both documents by hierarchically breaking them down into sentences, words, and characters. A Siamese neural network, incorporating a CNN and bi-directional LSTM layers with attention mechanisms, is then employed to derive a neural document embedding for each input. Specifically, the CNN layer extracts word representations from character embeddings, which are subsequently passed to an LSTM layer operating at the sentence level. The final LSTM layer synthesizes these sentence embeddings to construct a holistic document representation. Based on the document embeddings of $\Dunk$ and $\DA$, a nonlinear metric learning scheme is employed, which measures how closely $\Dunk$ and $\DA$ resemble each other.

Although language models are also inevitably affected by the issue of content bias, attempts to mitigate this problem are scarce. A notable mention is the approach by \textcite{fourkioti2019} consisting in training an \ngram language model on sequences of POS tags. The authors found that this kind of model is only valuable when fused with other features.

More recently, \textcite{RiveraSotoLUAR:2021} proposed a topic-shift robust \AV framework called \luarAV\footnote{\luarAV stands for \emph{\textbf{L}earning \textbf{U}niversal \textbf{A}uthorship \textbf{R}epresentations} \parencite{RiveraSotoLUAR:2021}.}, which essentially is a contrastively trained neural model based on an enhanced SBERT architecture. While SBERT consists of a BERT transformer followed by attention-weighted mean pooling, \luarAV extends this with a stylometric sampling strategy tailored for authorship. A text document is split into equal segments, from which one short token-sequence is randomly selected. \textcite{RiveraSotoLUAR:2021} report that this sampling method improves generalization, especially when training on longer texts. A self-attention layer processes the resulting chunk embeddings, which are then aggregated via max pooling and passed through a final linear layer. A similar approach called \bibertAV, proposed by \textcite{bibertav2024}, also leverages a pre-trained BERT model in a Siamese fashion, but combines it with Bi-LSTM layers. Although the evaluation of \bibertAV is limited to different variations of the Enron corpus and its less sophisticated design in contrast to \luarAV, it illustrates the effectiveness of Siamese architectures for \AV.

Related to language modeling, the other approach that has been found to be very successful is the use of compression algorithms. This family of methods is fundamentally very similar to language models because they both are probabilistic models of a training dataset. Language modeling can be thought of as a process of compression and one of the best compressor for text data, \e{Prediction by Partial Matching} (\ppm), could also be seen as a character-level language model. A particularly successful \AV method which makes use of \ppm as an underlying engine is \coav \parencite{HalvaniARES:2017, HalvaniOCCAV:2018}. The method was found to perform similarly to \imOrg, but in a much shorter time and without relying on impostors. Halvani's 2021 survey of the field \parencite{HalvaniPhD:2021} indeed concluded that \coav, together with \imOrg, are the best \AV methods. Both of them can be used together with \posNoise to avoid content bias. Although \imOrg tended to be more robust over the various tests it was subjected to, the advantage of \coav over \imOrg is that the method is deterministic and much faster (for comparison see, for example, \cite[Table~4]{HalvaniARES:2017}). Given a verification case $\Problem = (\Dunk, \DA)$, \coav compresses $\Dunk$, $\DA$ and the concatenation of both using a PPM variant called PPMd. Based on the length of the three compression results, a dissimilarity value is calculate using \e{Compression-based Cosine} (CBC). This dissimilarity is then compared to a threshold that is optimized on a training corpus to estimate whether either $\A = \unknown$ or $\A \neq \unknown$ holds. In addition,  \textcite{HalvaniPhD:2021} was able to visualize the importance of character sequences for the prediction by producing a color-coded version of the texts, thus making the method more qualitatively explorable by the analyst.

\subsubsection{Feature Vector Difference}

Finally, another \AV method that is also worth mentioning, and which was ranked second in the PAN AV competition 2020 \parencite{PANOverviewAV:2020}, is the so-called \emph{Feature Vector Difference} approach proposed by \textcite{WeerasingheFeVecDiff:2020}. Given a verification case $\Problem = (\Dunk, \DA)$, the first step of this method is to construct corresponding feature vectors $\featVector_{\unknown}, \featVector_{\A}$ using a set of features. Then, based on these two, a difference vector $\featVector = |\featVector_{\unknown} - \featVector_{\A}|$ is then calculated, which represents either a \classYdash{} or \classNdash{case}. This procedure is performed for all verification cases in a training corpus, and based on these a classifier is finally trained. For this, the authors considered two classification models: Logistic Regression (\janithLR) and Multi-Layer Perceptron (\janithMLP). Regarding the features, the authors chose \TFIDF values of: character \ngrams (for $n \in \{1, 2, \ldots, 6\}$), POS tag trigrams, 31 predefined special characters, trigrams of POS tag chunks (extracted from the parse tree representation of the document) as well as noun and verb phrase expansions. Moreover, they considered frequencies of 179 stop words (defined in the NLTK corpus package\footnote{\url{https://github.com/nltk/nltk}}), total number of characters and tokens in the document, average word length (characters per word) per document, distribution of token lengths (relative frequency of tokens with length $\ell$ (for $1 \leq \ell \leq 10$) and vocabulary richness measures.
\section{Cognitive Linguistics, Grammar, and Linguistic Individuality}\label{sec-cognitive}

After reviewing the \AV literature, this section introduces the theoretical grounding for the new method that we are proposing, which is motivated and inspired by the field of \e{Cognitive Linguistics}.

In Cognitive Linguistics, language and grammar are seen as complex adaptive systems \parencite{Beckner2009}, where regularities emerge from speakers interacting with each other, like a \e{phenomenon of the third kind} \parencite{DabrowskaLangPhenThirdKind:2020}. This is partially evidenced by decades of linguistics studies on corpora which have revealed that the traditional understanding of grammar as a set of phrase structure rules acting over basic Part-of-Speech categories (\eg nouns, adjectives) is untenable and that, instead, a better model of grammar is a \emph{lexicogrammar} continuum that goes from fixed lexicalized expressions (\eg~\quotetxt{I don't know}) to flexible templates (\eg~\quotetxt{the} X-er \quotetxt{the} Y-er) up to completely abstract structures (\eg~Agent Action Patient).

This continuum roughly corresponds to the continuum between \e{declarative memory}, which is semantic/encyclopedic memory for facts and which tends to be conscious, and \e{procedural memory}, which is instead memory for the probabilistic prediction of patterns of sequences and that deals with subconscious knowledge of habits and skills \parencite{ullman2004}. The roles that these two memories play in language is clearly evidenced in cases of aphasia. When an impairment affects circuits of declarative memory, this gives rise to \e{fluent aphasia}, resulting in speech that is mostly constituted by function words (articles, prepositions, pronouns) or morphological forms (\quotetxt{-ed}, \quotetxt{-s}) but with lack of content words that contain meaning, leading to fluent but nonsensical speech, almost resembling \posNoise preprocessed data such as the one in Table~\ref{table:ComparisonTopicMasking}. Instead, when the impairment is on procedural memory, then speech is only consisting of unconnected content words \parencite{ullman2015}. These two memory systems work together and, typically, once a habit is formed through repetition then procedural memory takes over, leading to fluency of expression. This transition explains why languages tend to be heavily formulaic. For example, in English one can say \quotetxt{big mistake} but not \quotetxt{large mistake}, even though a formal phrase structure rule based grammar would seemingly have no problems with the latter expression, which is an adjective modifying a noun. What explains this reliance on formulaic expressions is the cognitive tendency to rely as much as possible on habitual and ritualized patterns stored as templates \parencite{ullman2004, sinclair1991}.

The way these units of procedural memory are formed is through the process of \e{chunking}: when two items that often go together are found together, then these two items can be treated as one, thus saving space in working memory \parencite{Gobet2001, bybee2010, christiansen2016}. In all effects, chunking is a process of \e{information compression} and, if the units of grammar are chunks, then grammar itself can be thought of as a compressed version of a language or, equivalently, as a probability model over these units. According to Cognitive Linguistics, therefore, a probability model like a \textit{language model} is a better model of grammar than a list of phrase structure rules.

This way of conceiving language also explains why each individual has a different and possibly unique grammar, a statement encapsulated by the \e{Principle of Linguistic Individuality} \parencite{NiniTheoryLingAA:2023}. The status of \e{unithood} given by chunking is not discrete (\eg something is or is not a unit) but a continuum of strength of unithood, where this strength is called \e{entrenchment} \parencite{divjak2019, Schmid2015}. Because entrenchment is dependent on regular exposure and usage, it is fundamentally personal and idiosyncratic. It is for this reason that Langacker hypothesized that \quote{the set of units mastered by all members of a speech community might be a rather small proportion of the units constituting the linguistic ability of any given speaker} \parencite{langacker1987}, a hypothesis that is finding increasing empirical support \parencite{NiniTheoryLingAA:2023}. For example, for most if not all speakers of English the entrenchment for a sequence such as \quotetxt{of the} is maximal, meaning that they can produce it effortlessly and unconsciously as a single unit. In contrast, a sequence such as \quotetxt{the of} would not be a unit for speakers of English because it is virtually never encountered or used. If a speaker of English encounters this sequence, they would decompose it as a structure made up of two units, \quotetxt{the} and \quotetxt{of}. In between these two extreme examples there is a lot of freedom as to what counts as a unit for a person or another, depending on what each individual has automatized in their procedural memory. For example, an individual $\A_1$ might have a very entrenched sequence such as \quotetxt{it is known that} when writing academic articles while an individual $\A_2$ might have \quotetxt{it is well established that}. The higher the entrenchment, the more likely it is that $\A$ is using procedural memory and that it is therefore performing an unconscious routine that constitutes a unit for this individual. These sequences will then have open slots ready to be filled in with the appropriate meaning.

Although there have been other attempts at modeling the grammar of an author in the past for \AV or \AA purposes \parencite{chaski_empirical_2001,baayen_outside_1996,fourkioti2019,gormanUniversalDependenciesAuthor2022}, these methods relied too much on a conception of grammar in the traditional sense, for example using sequences of POS tags or syntactic trees. For the reasons outlined above, these ways of modeling grammar are not compatible with predictions made in Cognitive Science and Cognitive Linguistics. The picture of grammar painted by Cognitive Linguistics is much more lexically-based and word-idiosyncratic. For example, the preposition \quotetxt{of} in English behaves very oddly compared to other prepositions \parencite{sinclair1991}. Giving it the same POS tag as to other prepositions is therefore going to obscure a lot of important grammatical information.

With this theoretical grounding, it is now possible to present our proposed novel approach, \ourApproach.

\section{The Likelihood Ratio of Grammar Models}
\begin{figure}[h!]
    \centering
    \includegraphics[trim={4.8cm 25cm 15cm 9.6cm},clip,width=1\linewidth]{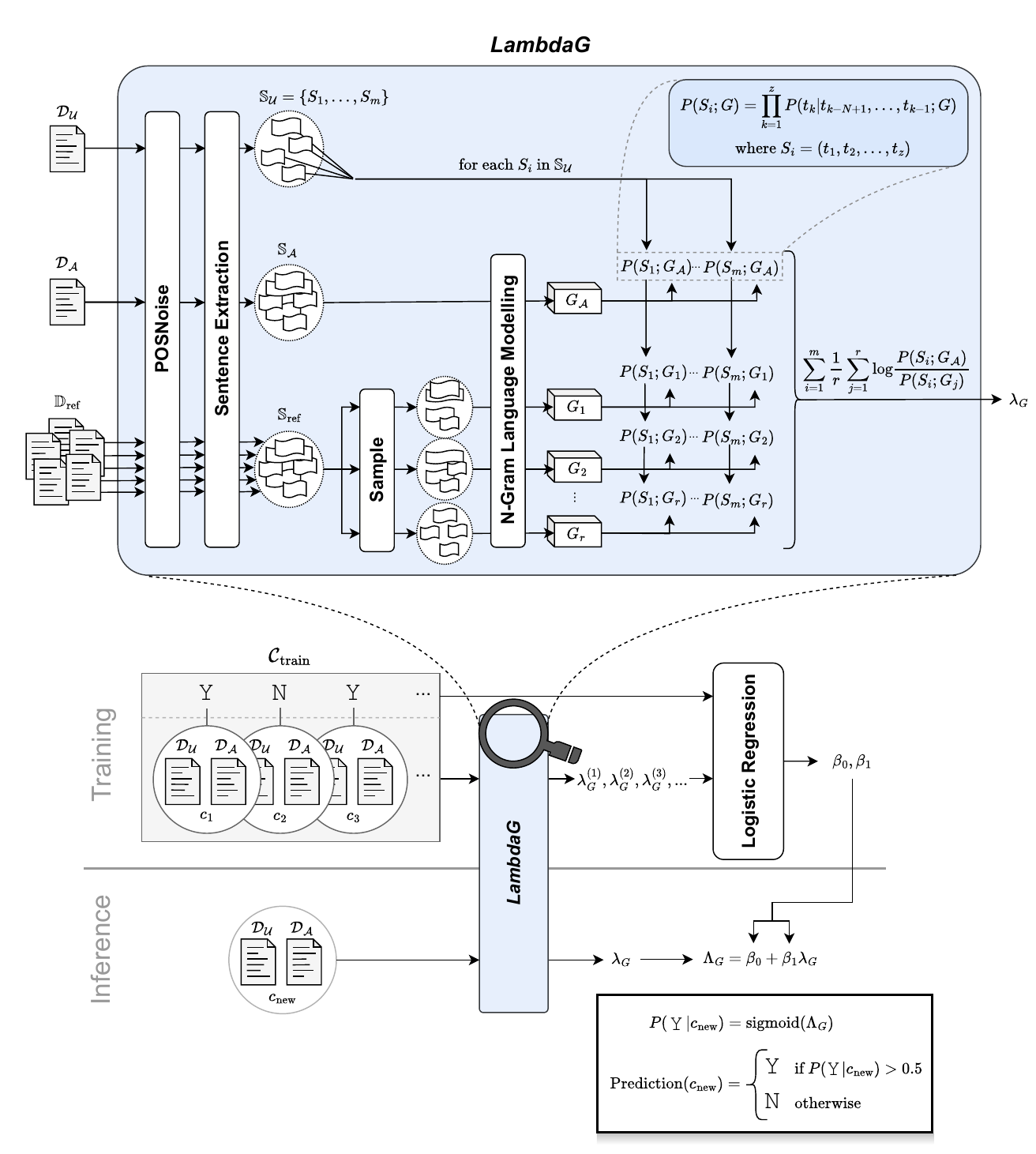}
    \caption{Schematic overview of our proposed \AV method \ourApproach.}
    \label{fig:overview}
\end{figure}

The algorithm presented in this paper is a \textbf{binary-extrinsic} \AV method. This section provides an overview of the algorithm, which is formally outlined in Algorithm \ref{AlgoGMComp} and schematically illustrated in Figure~\ref{fig:overview}. For a more detailed technical discussion of some of its components,  we refer the reader to Section~\ref{sec-methods}.

\begin{algorithm} [h!]
    \caption{Likelihood Ratio of Grammar Models (\ourApproach) \label{AlgoGMComp}}
	\SetNoFillComment
	\DontPrintSemicolon
    \SetKwInput{KwData}{Input}
	\SetKwInput{KwResult}{Output}
    \SetKwFunction{sample}{sample}
    \SetKwFunction{sent}{sent}
    \SetKwFunction{posnoise}{posnoise}
    \SetKwFunction{KN}{KN}
    \SetKwFunction{calibrate}{calibrate}
    \SetKwProg{Fn}{Function}{:}{}
    \KwData{$\Dunk$, $\DA$, $\Dref =\lbrace \mathcal{D}_{\text{ref}}^1, \mathcal{D}_{\text{ref}}^2, \dots \rbrace$, $\modelOrder$, $\numberRepetitions$}
    \tcp{where $\Dunk$, $\DA$ are two documents, $\Dref =\lbrace \mathcal{D}_{\text{ref}}^1, \mathcal{D}_{\text{ref}}^2, \dots \rbrace$ is a set of reference documents, $\modelOrder$ is the model order and $\numberRepetitions$ is the number of repetitions}
    \BlankLine
    
    \Fn{\sample{$s$, $\mathbb{S}$}}{
        Randomly select a subset of $s$ sentences from the given tokenized input sentence set $\mathbb{S}$\;
    }  
    \BlankLine

    \Fn{\posnoise{$\Dset$}}{
        Applies POSNoise to each document in $\Dset$ and returns a set of their \posnoise representations
    }  
    \BlankLine

    \Fn{\sent{$\Dset$}}{
        Construct a set $\mathbb{S}$ of tokenized sentences from all documents in $\Dset$\;
    }  
    \BlankLine
    \tcp{Apply \posNoise to all respective documents and construct tokenized sentences}
    $\SentenceSetUnk \leftarrow \sent(\posnoise(\lbrace \Dunk \rbrace))$\; 
    $\SentenceSetA \leftarrow \sent(\posnoise(\lbrace \DA \rbrace))$\; 
    $\SentenceSetRef \leftarrow \sent(\posnoise(\Dref))$\; 
    
    \vspace{1em}
    \tcp{Build Grammar Model for the known author $\A$ }
    $G_\A \leftarrow \KN(\mathbb{S}_\A, \modelOrder)$\;
    
    \vspace{1em}
    \tcp{Build reference Grammar Models}
    \For{$j \gets 1$ \KwTo $\numberRepetitions$}{
        \vspace{.2em}
        $\mathbb{S}_{\text{j}} \leftarrow$ \sample{$|\mathbb{S}_\A|$, $\mathbb{S}_{\text{ref}}$}\;
        $G_{j} \leftarrow \KN(\mathbb{S}_{\text{j}}, \modelOrder)$\; 
    }
    
    \vspace{1em}
    \tcp{Calculate $\lambda_G$ (the log-likelihood ratio of the Grammar Models) over $\SentenceSetUnk$}
    $\lambda_G(\SentenceSetUnk) \leftarrow 0$ \;

    \For{$\sentence_i \in \mathbb{S}_{\unknown}$}{
        \vspace{.2em}
        \tcp{Decompose $\sentence_i$ into a sequence of tokens}
        $(\token_1, \token_2, \ldots, \token_z) \leftarrow \sentence_i$ \;
        \vspace{1em}
        \For{$k \gets 1$ \KwTo $z$}{
            \tcp{Calculate mean $\lambda_G$ for $\token_k$ over reference Grammar Models}
            \For{$j \gets 1$ \KwTo $\numberRepetitions$}{
                \vspace{.2em}
                \vspace{.2em}
                $\lambda_G(\SentenceSetUnk) \leftarrow \lambda_G(\SentenceSetUnk) + \dfrac{1}{\numberRepetitions} \log \dfrac{P(\token_k|t_{<k};G_{\A})}{P(\token_k|t_{<k};G_j)} $\;
            }
        }
    }
    \vspace{0.2em}
    \Return $\lambda_G(\SentenceSetUnk)$\;
\end{algorithm}

We define the most basic grammatical unit of analysis as a function token $\token \in \vocabulary_L$, where $\vocabulary_L$ is the set all function tokens of a language \(L\), including all function words, punctuation marks and abstract grammatical categories. For instance, for English, $\vocabulary_{E}=\lbrace \texttt{the}, \texttt{of}, \texttt{at}, \dots, \allowbreak\texttt{NOUN}, \allowbreak\texttt{VERB}, \allowbreak\texttt{ADJECTIVE}, \dots, \texttt{!}, \texttt{:}, \texttt{;} \dots \rbrace$. We then define a Grammar Model \(G\) as a statistical model that generates a probability distribution over sentences $\mathbb{S}$, which are defined as complete sequences of function tokens. To obtain these representations, we first apply the \posNoise algorithm (see Section~\ref{sec-the-posnoise-algorithm}) to $\Dunk$, $\DA$, and each $\D$ in the reference corpus $\Dref$, retaining only function tokens. This is indicated in Algorithm~\ref{AlgoGMComp}, lines 7--9. Although Cognitive Linguistics would predict that grammatical categories contain more specificity than the broad ones used in the \posNoise algorithm, this is a better approximation than other category-free algorithms like \textDistortion. Then, we tokenize each $\D$ into sentences by adding sentence boundaries markers at an end of sentence punctuation mark or new line, thus turning them into their respective set of sentences $\SentenceSetA$, $\SentenceSetUnk$ and $\SentenceSetRef$ (also lines 7--9 in Algorithm~\ref{AlgoGMComp}). The boundaries of each sentence could also be set by a grammatical parsing algorithm. However, we observed that \ourApproach is robust to this variation.
%-----------------------------------------
We then employ \ngram models, which allow the probability of any grammatical token to depend only on a finite number of preceding tokens from the same sentence, thereby capturing only the information contained in short-term grammatical correlations. \ngram models assume that each sentence is independent and we maintain that this is not an unreasonable assumption since grammatical dependencies between tokens are known to be relatively local \parencite{christiansen2016}. The probability assigned to each sentence $\sentence_i = (\token_1, \token_2, \ldots, \token_z)$ by a Grammar Model $G$ would then be calculated as follows:
%-----------------------------------------
\begin{equation} \label{eq:chainprob}
P(\sentence_i;G) = \prod _{k = 1}^{z} P(\token_k \vert  \underbrace{\token_{k-N+1}, \token_{k-N+2}, \ldots, \token_{k-1}}_{\token_{<k}};G)
\end{equation}

%-----------------------------------------
For readability reasons, we rewrite the probability assigned by $G$ to each token $\token_k$ in its context as $P(\token_k|\token_{<k};G)$ (as in line 19 in Algorithm~\ref{AlgoGMComp}). We estimate Grammar Models adopting Kneser-Ney smoothing \parencite{kneser1995, ChenGoodmanLMSmoothing:1996}, represented in Algorithm~\ref{AlgoGMComp} as $\KN()$, with a fixed single default discount parameter $D = 0.75$, as explained in Section~\ref{sec-ngram-modelling}. We set $\modelOrder=10$ as an order of the model to ensure that most grammatical relations among the tokens of a sentence are captured, even in cases of long sentences. However, the order of the model $\modelOrder$ could be considered as a hyperparameter of the algorithm (lines 10 and 13). 

Assuming the \textit{Principle of Linguistic Individuality} that at any time $\tau$, for any language $L_\tau$, there do not exist two individuals, $\A$ and $\B$, for whom their grammar is identical \parencite{NiniTheoryLingAA:2023}, we expect that in a large number of cases the two Grammar Models $G_{\A}$ and $G_{\B}$ would assign two different probabilities to the same token in context. $P(\token_k|\token_{<k};G_{\A})$ can be interpreted as the extent to which token $\token_k$ is 'grammatical' or, according to Cognitive Linguistic theories, \textit{entrenched} for author $\A$ when in context $\token_{<k}$. For this reason, we would expect that, if the sequence $\sigma = (\token_1, \token_2, \token_3)$ was produced by $G_\A$, then $P(\sigma;G_{\A}) > P(\sigma;G_{\B})$ or, more generally according to the \textit{Principle of Linguistic Individuality}, for any $\A' \neq \A$, $P(\sigma;G_{\A}) > P(\sigma;G_{\A'})$. 
%-----------------------------------------
The model $G_\A$ is estimated from $\SentenceSetA$ (line 10 in Algorithm~\ref{AlgoGMComp}). To approximate a comparison population of $G_{\A'}$, we estimate a set of reference Grammar Models $\SentenceSetRef = \lbrace G_1, G_2, \dots, G_r \rbrace$ where each $G_r$ is generated by uniformly sampling randomly from $\SentenceSetRef$ a set of sentences $\mathbb{S}_{r}$ of the same size as the set of sentences in the candidate author's corpus, $\SentenceSetA$, so that $|\mathbb{S}_{\text{ref}}| = |\SentenceSetA|$ (line 12 in Algorithm~\ref{AlgoGMComp}). We set $\numberRepetitions = 100$ but this value can also be considered a hyperparameter of the algorithm. A comparison of hyperparameter settings is reported in Section~\ref{sec-results}.
%-----------------------------------------
Taking an approach consistent with the likelihood ratio framework adopted in forensic science, similarly to \textcite{ishihara_forensic_2011}, it is therefore possible to calculate (lines 17--19 in Algorithm~\ref{AlgoGMComp}) a mean ratio of the likelihoods assigned to each token in context by $G_\A$ and by each $G_r$, which we call $\lambda_G$:
%-----------------------------------------
\begin{equation}\label{eq:lambda-f}
\lambda_G(\token_k|\token_{<k})=\frac{1}{\numberRepetitions}\sum_{j = 1}^r{\log\frac{P(\token_k|\token_{<k};G_{\A})}{P(\token_k|\token_{<k};G_j)}}
\end{equation}
%-----------------------------------------
In Eq.~\ref{eq:lambda-f} the numerator can be seen as a measure of \textit{similarity}, how much the token is grammatical or entrenched for $G_{\A}$ given its context, while the denominator is a measure of \textit{typicality}, how atypical this token is in this context in the reference population. A value of $\lambda_G$ for each $\sentence_i \in \SentenceSetUnk = \lbrace \sentence_1, \sentence_2, \dots, \sentence_m \rbrace$, can then be calculated (line 15--19 in Algorithm~\ref{AlgoGMComp}) as:
%-----------------------------------------
\begin{equation}\label{eq:lambda-s}
\lambda_G(\sentence_i) = \sum_{k = 1}^{z}\lambda_G(\token_k|\token_{<k})
\end{equation}
%-----------------------------------------
Since we model each $\sentence_i \in \SentenceSetUnk$ as independent from each other, we can then calculate (line 14--20 in Algorithm~\ref{AlgoGMComp}):
\begin{align}\label{eq:lambda-d}
\begin{split}
\lambda_G(\SentenceSetUnk) = \sum_{i = 1}^{m}\hspace{-0.2em}\lambda_G(\sentence_i) = \sum\limits_{i=1}^{m}{\sum\limits_{k = 1}^{z}\lambda_G(\token_k|\token_{<k})} 
\end{split}
\end{align}
%-----------------------------------------
An advantage of this formula is that the final $\lambda_G$ can be decomposed into $\lambda_G$ values for each single sentence and each single token in each single sentence. This means that the analyst can reconstruct step by step what features contribute to the final score, for example, by ranking each sentence for importance and, for each sentence, color-code or highlight the most important tokens, as demonstrated in Section~\ref{sec-appendix-a}. 
%----------------------------------------- 
In addition to this property, the $\lambda_G$ value is also effectively an uncalibrated log-likelihood ratio score that includes information about both similarity and typicality and that can then be calibrated into a forensic log-likelihood ratio using a calibration method such as a logistic regression. We denote a calibrated $\lambda_G$ log-likelihood ratio that expresses the strength of evidence in a forensic context with the upper case lambda ($\Lambda_G$). To perform this calibration we use a training corpus containing \classYdash{} and \classNdash{cases} to estimate a logistic regression calibration model \parencite{vanleeuwen2015,ishihara2021} that turns each $\lambda_G$ into $\Lambda_G$. Here, $\Lambda_G$ quantifies the strength of the linguistic evidence for each of the two alternative forensic propositions. More details on the Likelihood Ratio Framework for forensic science and its evaluation is described in Section~\ref{sec-the-likelihood-ratio-framework}. Based on $\Lambda_G$, we classify a new verification case \ProblemNew as \classY if $\Lambda_G > 0$ (that is, if $\text{sigmoid}(\Lambda_G) > 0.5$), else as \classN.

\section{Results}\label{sec-results}

\begin{figure} % [h!]
    \centering
    \includegraphics[clip, trim=0 0 0 0.6em,width=.9\linewidth]{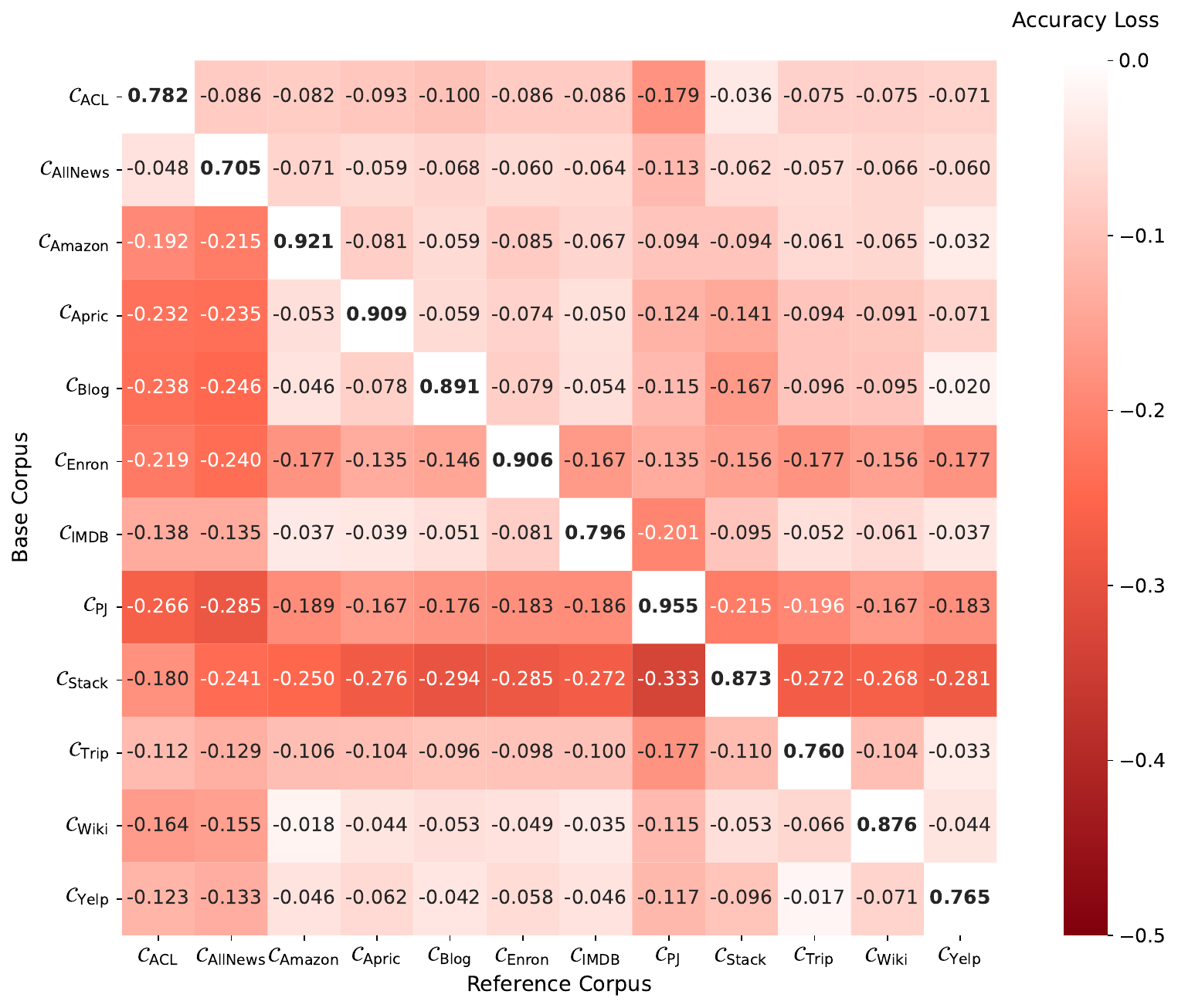}
    \includegraphics[clip, trim=0 0 0 0.6em,width=.9\linewidth]{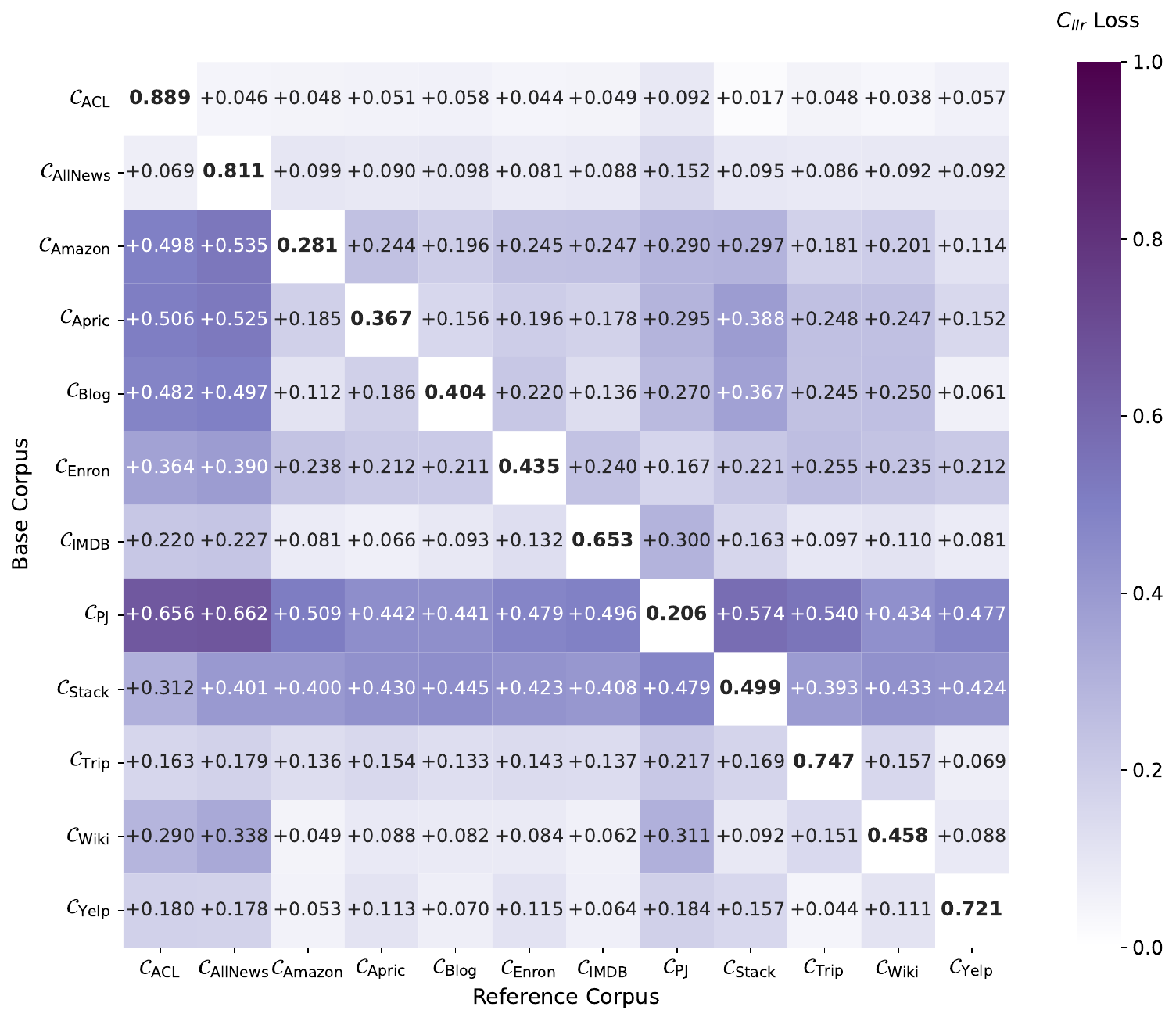} 
    \caption{The loss in Accuracy (top) and \cllr (bottom) results for cross-corpus comparison, \ie evaluating on Base Corpus while using reference texts $\Dref$ from Reference Corpus. Diagonal bold values denote the original Accuracy and \cllr, respectively. Darker shades denote a greater loss in performance.}
    \label{fig:crossgenre}
\end{figure}

Algorithm~\ref{AlgoGMComp} was applied to a set of twelve corpora that simulate various real-life \AV problems and scenarios, described in Section~\ref{sec-methods}. The complete results are summarized in Table~\ref{tab:EvalResults}, where the performance of all methods is quantified according to the five standard measures for classification: Accuracy, Area Under the ROC Curve (AUC), \fOne, Precision and Recall. These are values that were calculated using a threshold \Threshold, which is derived from $\Lambda_G$. The split between training and test datasets for each corpus tested is shown in Table~\ref{tab:CorpusStatistics}. To further assess the performance of $\Lambda_G$, the $\cllr$ measure is also used, which is a typical measure of accuracy for a likelihood ratio. This metric is explained in Section~\ref{sec-the-likelihood-ratio-framework}. When running the analysis on the training corpora, the reference corpus $\Dref$ was the remaining of the corpus after taking out the authors of $\DA$ and $\Dunk$. When analyzing the test corpora, the reference corpus was simply the corresponding training corpus. A robustness test using different corpora as reference data was also carried out (see Figure~\ref{fig:crossgenre}). Given the fact that \ourApproach, \imOrg, \adhominem and \luarAV are non-deterministic, each analysis was repeated five times, and the results presented in Table~\ref{tab:EvalResults} are the average of these five repetitions. 

In Table~\ref{tab:EvalResults}, \ourApproach is compared with \numberOfBaselines well-known and tested baseline \AV methods: \imOrg, \coav, \taveer, \adhominem, \janithDiffVecLR, \janithDiffVecMLP and \luarAV. \imOrg, \coav and \janithDiffVec (\janithLR and \janithMLP) were run on \posNoise preprocessed corpora and are therefore content-agnostic, similarly to \ourApproach. \taveer, on the other hand, was applied to the unprocessed corpora. However, since this method is inherently designed to consider only topic-agnostic features\footnote{Note that both \taveer and \posNoise rely on the same list of topic-agnostic words and phrases shown in Table~\ref{table:POSNoiseFeatures}.}, its results remain comparable to those of \ourApproach. With regard to \luarAV and \adhominem, we wish to point out that both methods are based on pre-trained neural network models, which were originally trained on \posNoise-unprocessed text data. Using pre-trained models is essential because training neural networks from scratch requires datasets that are larger than our \AV corpora. Instead, we therefore fine-tuned the original pre-trained models on our \AV corpora. Since pre-training was performed on \posNoise unprocessed text, the standard approach would be to fine-tune on similarly unprocessed \AV corpora (denoted as \luarAVOrg and \adhominemOrg in Table~\ref{tab:EvalResults}). However, this could introduce topic-related distortions, making the results less comparable with the other \AV methods used. To mitigate this, we also applied \luarAV and \adhominem on our \posNoise preprocessed corpora.

%-------------------------------------
\definecolor{lgray}{gray}{0.6}
\definecolor{lyellow}{HTML}{fff7bc}
\definecolor{lightgray}{rgb}{0.85, 0.85, 0.85}
%-------------------------------------
\setlength{\aboverulesep}{0pt}
\setlength{\belowrulesep}{0pt}
\setlength{\extrarowheight}{.15ex}
%-------------------------------------
\newcolumntype{g}{>{\columncolor{lightgray}}r}
%-------------------------------------
\begin{center} 
	  % \small % \footnotesize    
    \begin{longtable}{p{0.21cm}lrrrrrrrrr}
        \caption{Evaluation results of \ourApproach and the \numberOfBaselines introduced baselines. 
        $(\bm{\dagger})$ is marked next to a pre-trained neural network approach to indicate that it was fine-tuned using the original (\ie non-\posNoise-processed) corpora. In terms of Accuracy and AUC, bold and underlined values represent the best and second-best results, respectively. For clarity of presentation, the four confusion matrix outputs corresponding to each tested method have been shaded in gray. \label{tab:EvalResults}}\\\toprule
        %========================================================
        &  \textbf{Method} & \textbf{Acc.} & \textbf{AUC} & \textbf{F1} & \textbf{Prec.} & \textbf{Rec.} &\color{lgray}\textbf{TP} & \color{lgray}\textbf{FN} & \color{lgray}\textbf{FP} & \color{lgray}\textbf{TN} \\\midrule	
        %========================================================	
        & \cellcolor{lyellow}{\ourApproach} & \cellcolor{lyellow}{\textbf{0.906}} & \cellcolor{lyellow}{\textbf{0.938}} & \cellcolor{lyellow}{0.903} & \cellcolor{lyellow}{0.875} & \cellcolor{lyellow}{0.933} &\textcolor{lgray}{42} & \textcolor{lgray}{6} & \textcolor{lgray}{3} & \textcolor{lgray}{45} \\
        & \imOrg & \underline{0.823} & \underline{0.920} & 0.817 & 0.792 & 0.844 &\textcolor{lgray}{38} & \textcolor{lgray}{10} & \textcolor{lgray}{7} & \textcolor{lgray}{41} \\
        & \coav & 0.750 & 0.827 & 0.760 & 0.792 & 0.731 &\textcolor{lgray}{38} & \textcolor{lgray}{10} & \textcolor{lgray}{14} & \textcolor{lgray}{34} \\
        & \taveer & 0.729 & 0.842 & 0.735 & 0.750 & 0.720 &\textcolor{lgray}{36} & \textcolor{lgray}{12} & \textcolor{lgray}{14} & \textcolor{lgray}{34} \\ 
        & \janithDiffVecLR & 0.781 & 0.855 & 0.784 & 0.776 & 0.792 & \textcolor{lgray}{38} & \textcolor{lgray}{10} & \textcolor{lgray}{11} & \textcolor{lgray}{37} \\
        & \janithDiffVecMLP & 0.750 & 0.839 & 0.765 & 0.722 & 0.812 & \textcolor{lgray}{39} & \textcolor{lgray}{9} & \textcolor{lgray}{15} & \textcolor{lgray}{33} \\     
        & \adhominem & 0.583 & 0.594 & 0.474 & 0.643 & 0.375 & \textcolor{lgray}{18} & \textcolor{lgray}{30} & \textcolor{lgray}{10} & \textcolor{lgray}{38} \\
        \multirow{-5}{*}{\large\rotatebox{90}{$\bm{\CorpusEnron}$}}
        & \adhominemOrg & 0.573 & 0.607 & 0.438 & 0.640 & 0.333 & \textcolor{lgray}{16} & \textcolor{lgray}{32} & \textcolor{lgray}{9} & \textcolor{lgray}{39} \\
        & \luarAV & 0.750 & 0.810 & 0.755 & 0.740 & 0.771 & \textcolor{lgray}{37} & \textcolor{lgray}{11} & \textcolor{lgray}{13} & \textcolor{lgray}{35} \\
        & \luarAVOrg & 0.719 & 0.831 & 0.727 & 0.706 & 0.750 & \textcolor{lgray}{36} & \textcolor{lgray}{12} & \textcolor{lgray}{15} & \textcolor{lgray}{33} \\\midrule        
        %-------------------------------------------------------------------- 
        & \cellcolor{lyellow}{\ourApproach} & \cellcolor{lyellow}{\underline{0.876}} & \cellcolor{lyellow}{\underline{0.938}} & \cellcolor{lyellow}{0.872} & \cellcolor{lyellow}{0.841} & \cellcolor{lyellow}{0.905} &\textcolor{lgray}{95} & \textcolor{lgray}{18} & \textcolor{lgray}{10} & \textcolor{lgray}{103} \\
        & \imOrg & 0.854 & 0.927 & 0.862 & 0.912 & 0.817 &\textcolor{lgray}{103} & \textcolor{lgray}{10} & \textcolor{lgray}{23} & \textcolor{lgray}{90} \\
        & \coav & 0.801 & 0.912 & 0.793 & 0.761 & 0.827 &\textcolor{lgray}{86} & \textcolor{lgray}{27} & \textcolor{lgray}{18} & \textcolor{lgray}{95} \\
        & \taveer & 0.788 & 0.858 & 0.789 & 0.796 & 0.783 &\textcolor{lgray}{90} & \textcolor{lgray}{23} & \textcolor{lgray}{25} & \textcolor{lgray}{88} \\
        & \janithDiffVecLR & 0.739 & 0.840 & 0.678 & 0.886 & 0.549 & \textcolor{lgray}{62} & \textcolor{lgray}{51} & \textcolor{lgray}{8} & \textcolor{lgray}{105} \\
        & \janithDiffVecMLP & 0.735 & 0.809 & 0.737 & 0.730 & 0.743 & \textcolor{lgray}{84} & \textcolor{lgray}{29} & \textcolor{lgray}{31} & \textcolor{lgray}{82} \\   
        & \adhominem & 0.584 & 0.641 & 0.510 & 0.620 & 0.434 & \textcolor{lgray}{49} & \textcolor{lgray}{64} & \textcolor{lgray}{30} & \textcolor{lgray}{83} \\
        \multirow{-5}{*}{\large\rotatebox{90}{$\bm{\CorpusWiki}$}}  
        & \adhominemOrg & 0.575 & 0.618 & 0.575 & 0.575 & 0.575 & \textcolor{lgray}{65} & \textcolor{lgray}{48} & \textcolor{lgray}{48} & \textcolor{lgray}{65} \\
        & \luarAV & 0.761 & 0.833 & 0.755 & 0.776 & 0.735 & \textcolor{lgray}{83} & \textcolor{lgray}{30} & \textcolor{lgray}{24} & \textcolor{lgray}{89} \\ 
        & \luarAVOrg & \textbf{0.885} & \textbf{0.963} & 0.887 & 0.872 & 0.903 & \textcolor{lgray}{102} & \textcolor{lgray}{11} & \textcolor{lgray}{15} & \textcolor{lgray}{98} \\\midrule
        
        %--------------------------------------------------------------------  
        \pagebreak
        %--------------------------------------------------------------------         
        %========================================================	  
        % Add header for better readability due to table fragmentation...
        &  \textbf{Method} & \textbf{Acc.} & \textbf{AUC} & \textbf{F1} & \textbf{Prec.} & \textbf{Rec.} &\color{lgray}\textbf{TP} & \color{lgray}\textbf{FN} & \color{lgray}\textbf{FP} & \color{lgray}\textbf{TN} \\\midrule	
        %========================================================	
        
        %--------------------------------------------------------------------  
        & \cellcolor{lyellow}{\ourApproach} & \cellcolor{lyellow}{0.873} & \cellcolor{lyellow}{0.933} & \cellcolor{lyellow}{0.864} & \cellcolor{lyellow}{0.807} & \cellcolor{lyellow}{0.929} &\textcolor{lgray}{92} & \textcolor{lgray}{22} & \textcolor{lgray}{7} & \textcolor{lgray}{107} \\
        & \imOrg & 0.689 & 0.738 & 0.684 & 0.675 & 0.694 &\textcolor{lgray}{77} & \textcolor{lgray}{37} & \textcolor{lgray}{34} & \textcolor{lgray}{80} \\
        & \coav & 0.561 & 0.640 & 0.554 & 0.544 & 0.564 &\textcolor{lgray}{62} & \textcolor{lgray}{52} & \textcolor{lgray}{48} & \textcolor{lgray}{66} \\
        & \taveer & 0.693 & 0.763 & 0.685 & 0.667 & 0.704 &\textcolor{lgray}{76} & \textcolor{lgray}{38} & \textcolor{lgray}{32} & \textcolor{lgray}{82} \\
        & \janithDiffVecLR & \textbf{0.939} & \textbf{0.977} & 0.941 & 0.910 & 0.974 & \textcolor{lgray}{111} & \textcolor{lgray}{3} & \textcolor{lgray}{11} & \textcolor{lgray}{103} \\
        & \janithDiffVecMLP & \underline{0.882} & \underline{0.957} & 0.883 & 0.872 & 0.895 & \textcolor{lgray}{102} & \textcolor{lgray}{12} & \textcolor{lgray}{15} & \textcolor{lgray}{99} \\    
        & \adhominem & 0.504 & 0.516 & 0.464 & 0.505 & 0.430 & \textcolor{lgray}{49} & \textcolor{lgray}{65} & \textcolor{lgray}{48} & \textcolor{lgray}{66} \\
        \multirow{-5}{*}{\large\rotatebox{90}{$\bm{\CorpusStack}$}}
        & \adhominemOrg & 0.596 & 0.594 & 0.603 & 0.593 & 0.614 & \textcolor{lgray}{70} & \textcolor{lgray}{44} & \textcolor{lgray}{48} & \textcolor{lgray}{66} \\
        & \luarAV & 0.680 & 0.760 & 0.684 & 0.675 & 0.693 & \textcolor{lgray}{79} & \textcolor{lgray}{35} & \textcolor{lgray}{38} & \textcolor{lgray}{76} \\ 
        & \luarAVOrg & 0.614 & 0.662 & 0.596 & 0.625 & 0.570 & \textcolor{lgray}{65} & \textcolor{lgray}{49} & \textcolor{lgray}{39} & \textcolor{lgray}{75} \\\midrule 
        
        %-------------------------------------------------------------------- 
        & \cellcolor{lyellow}{\ourApproach} & \cellcolor{lyellow}{\textbf{0.782}} & \cellcolor{lyellow}{\textbf{0.861}} & \cellcolor{lyellow}{0.775} & \cellcolor{lyellow}{0.750} & \cellcolor{lyellow}{0.802} &\textcolor{lgray}{105} & \textcolor{lgray}{35} & \textcolor{lgray}{26} & \textcolor{lgray}{114} \\
        & \imOrg & \underline{0.775} & \underline{0.851} & 0.755 & 0.693 & 0.829 &\textcolor{lgray}{97} & \textcolor{lgray}{43} & \textcolor{lgray}{20} & \textcolor{lgray}{120} \\
        & \coav & 0.750 & 0.835 & 0.751 & 0.757 & 0.746 &\textcolor{lgray}{106} & \textcolor{lgray}{34} & \textcolor{lgray}{36} & \textcolor{lgray}{104} \\
        & \taveer & 0.707 & 0.748 & 0.689 & 0.650 & 0.734 &\textcolor{lgray}{91} & \textcolor{lgray}{49} & \textcolor{lgray}{33} & \textcolor{lgray}{107} \\
        & \janithDiffVecLR & 0.714 & 0.752 & 0.750 & 0.667 & 0.857 & \textcolor{lgray}{120} & \textcolor{lgray}{20} & \textcolor{lgray}{60} & \textcolor{lgray}{80} \\
        & \janithDiffVecMLP & 0.564 & 0.632 & 0.527 & 0.576 & 0.486 & \textcolor{lgray}{68} & \textcolor{lgray}{72} & \textcolor{lgray}{50} & \textcolor{lgray}{90} \\    
        & \adhominem & 0.554 & 0.575 & 0.449 & 0.586 & 0.364 & \textcolor{lgray}{51} & \textcolor{lgray}{89} & \textcolor{lgray}{36} & \textcolor{lgray}{104} \\
        \multirow{-5}{*}{\large\rotatebox{90}{$\bm{\CorpusACL}$}}
        & \adhominemOrg & 0.496 & 0.512 & 0.598 & 0.498 & 0.750 & \textcolor{lgray}{105} & \textcolor{lgray}{35} & \textcolor{lgray}{106} & \textcolor{lgray}{34} \\
        & \luarAV & 0.593 & 0.658 & 0.610 & 0.586 & 0.636 & \textcolor{lgray}{89} & \textcolor{lgray}{51} & \textcolor{lgray}{63} & \textcolor{lgray}{77} \\ 
        & \luarAVOrg & 0.704 & 0.780 & 0.700 & 0.708 & 0.693 & \textcolor{lgray}{97} & \textcolor{lgray}{43} & \textcolor{lgray}{40} & \textcolor{lgray}{100} \\\midrule
        
        & \cellcolor{lyellow}{\ourApproach} & \cellcolor{lyellow}{\textbf{0.955}} & \cellcolor{lyellow}{\textbf{0.990}} & \cellcolor{lyellow}{0.955} & \cellcolor{lyellow}{0.955} & \cellcolor{lyellow}{0.955} &\textcolor{lgray}{149} & \textcolor{lgray}{7} & \textcolor{lgray}{7} & \textcolor{lgray}{149} \\
        & \imOrg & 0.920 & 0.980 & 0.916 & 0.878 & 0.958 &\textcolor{lgray}{137} & \textcolor{lgray}{19} & \textcolor{lgray}{6} & \textcolor{lgray}{150} \\
        & \coav & 0.901 & 0.960 & 0.901 & 0.910 & 0.893 &\textcolor{lgray}{142} & \textcolor{lgray}{14} & \textcolor{lgray}{17} & \textcolor{lgray}{139} \\
        & \taveer & 0.885 & 0.939 & 0.889 & 0.923 & 0.857 &\textcolor{lgray}{144} & \textcolor{lgray}{12} & \textcolor{lgray}{24} & \textcolor{lgray}{132} \\
        & \janithDiffVecLR & 0.878 & 0.889 & 0.886 & 0.835 & 0.942 & \textcolor{lgray}{147} & \textcolor{lgray}{9} & \textcolor{lgray}{29} & \textcolor{lgray}{127} \\
        & \janithDiffVecMLP & 0.865 & 0.930 & 0.878 & 0.803 & 0.968 & \textcolor{lgray}{151} & \textcolor{lgray}{5} & \textcolor{lgray}{37} & \textcolor{lgray}{119} \\ 
        & \adhominem & 0.673 & 0.787 & 0.625 & 0.733 & 0.545 & \textcolor{lgray}{85} & \textcolor{lgray}{71} & \textcolor{lgray}{31} & \textcolor{lgray}{125} \\
        \multirow{-5}{*}{\large\rotatebox{90}{$\bm{\CorpusPJ}$}}
        & \adhominemOrg & 0.590 & 0.593 & 0.642 & 0.569 & 0.737 & \textcolor{lgray}{115} & \textcolor{lgray}{41} & \textcolor{lgray}{87} & \textcolor{lgray}{69} \\
        & \luarAV & 0.894 & 0.963 & 0.890 & 0.930 & 0.853 & \textcolor{lgray}{133} & \textcolor{lgray}{23} & \textcolor{lgray}{10} & \textcolor{lgray}{146} \\ 
        & \luarAVOrg & \underline{0.952} & \underline{0.987} & 0.951 & 0.967 & 0.936 & \textcolor{lgray}{146} & \textcolor{lgray}{10} & \textcolor{lgray}{5} & \textcolor{lgray}{151} \\\midrule
        %-------------------------------------------------------------------- 

        & \cellcolor{lyellow}{\ourApproach} & \cellcolor{lyellow}{\textbf{0.909}} & \cellcolor{lyellow}{\textbf{0.961}} & \cellcolor{lyellow}{0.911} & \cellcolor{lyellow}{0.929} & \cellcolor{lyellow}{0.893} &\textcolor{lgray}{158} & \textcolor{lgray}{12} & \textcolor{lgray}{19} & \textcolor{lgray}{151} \\
        & \imOrg & \underline{0.885} & 0.939 & 0.878 & 0.824 & 0.940 &\textcolor{lgray}{140} & \textcolor{lgray}{30} & \textcolor{lgray}{9} & \textcolor{lgray}{161} \\
        & \coav & 0.832 & 0.910 & 0.840 & 0.882 & 0.802 &\textcolor{lgray}{150} & \textcolor{lgray}{20} & \textcolor{lgray}{37} & \textcolor{lgray}{133} \\
        & \taveer & 0.829 & 0.891 & 0.822 & 0.788 & 0.859 &\textcolor{lgray}{134} & \textcolor{lgray}{36} & \textcolor{lgray}{22} & \textcolor{lgray}{148} \\
        & \janithDiffVecLR & 0.829 & 0.920 & 0.839 & 0.795 & 0.888 & \textcolor{lgray}{151} & \textcolor{lgray}{19} & \textcolor{lgray}{39} & \textcolor{lgray}{131} \\
        & \janithDiffVecMLP & 0.821 & 0.895 & 0.828 & 0.795 & 0.865 & \textcolor{lgray}{147} & \textcolor{lgray}{23} & \textcolor{lgray}{38} & \textcolor{lgray}{132} \\     
        & \adhominem    & 0.553 & 0.610 & 0.510 & 0.564 & 0.465 & \textcolor{lgray}{79} & \textcolor{lgray}{91} & \textcolor{lgray}{61} & \textcolor{lgray}{109} \\
        \multirow{-5}{*}{\large\rotatebox{90}{$\bm{\CorpusApricity}$}}
        & \adhominemOrg & 0.526 & 0.592 & 0.410 & 0.544 & 0.329 & \textcolor{lgray}{56} & \textcolor{lgray}{114} & \textcolor{lgray}{47} & \textcolor{lgray}{123} \\
        & \luarAV & 0.850 & 0.930 & 0.854 & 0.832 & 0.876 & \textcolor{lgray}{149} & \textcolor{lgray}{21} & \textcolor{lgray}{30} & \textcolor{lgray}{140} \\
        & \luarAVOrg & 0.879 & \underline{0.952} & 0.884 & 0.852 & 0.918 & \textcolor{lgray}{156} & \textcolor{lgray}{14} & \textcolor{lgray}{27} & \textcolor{lgray}{143} \\\midrule
        %--------------------------------------------------------------------

        & \cellcolor{lyellow}{\ourApproach} & \cellcolor{lyellow}{\textbf{0.760}} & \cellcolor{lyellow}{\textbf{0.822}} & \cellcolor{lyellow}{0.758} & \cellcolor{lyellow}{0.750} & \cellcolor{lyellow}{0.766} &\textcolor{lgray}{180} & \textcolor{lgray}{60} & \textcolor{lgray}{55} & \textcolor{lgray}{185} \\
        & \imOrg & 0.694 & 0.761 & 0.610 & 0.479 & 0.839 &\textcolor{lgray}{115} & \textcolor{lgray}{125} & \textcolor{lgray}{22} & \textcolor{lgray}{218} \\
        & \coav & 0.685 & 0.748 & 0.684 & 0.683 & 0.686 &\textcolor{lgray}{164} & \textcolor{lgray}{76} & \textcolor{lgray}{75} & \textcolor{lgray}{165} \\
        & \taveer & 0.690 & 0.729 & 0.655 & 0.588 & 0.738 &\textcolor{lgray}{141} & \textcolor{lgray}{99} & \textcolor{lgray}{50} & \textcolor{lgray}{190} \\
        & \janithDiffVecLR & 0.533 & 0.534 & 0.533 & 0.533 & 0.533 & \textcolor{lgray}{128} & \textcolor{lgray}{112} & \textcolor{lgray}{112} & \textcolor{lgray}{128} \\
        & \janithDiffVecMLP & 0.477 & 0.475 & 0.516 & 0.480 & 0.558 & \textcolor{lgray}{134} & \textcolor{lgray}{106} & \textcolor{lgray}{145} & \textcolor{lgray}{95} \\           
        & \adhominem & 0.494 & 0.497 & 0.475 & 0.493 & 0.458 & \textcolor{lgray}{110} & \textcolor{lgray}{130} & \textcolor{lgray}{113} & \textcolor{lgray}{127} \\
        \multirow{-5}{*}{\large\rotatebox{90}{$\bm{\CorpusTripAdvisor}$}} 
        & \adhominemOrg & 0.502 & 0.525 & 0.505 & 0.502 & 0.508 & \textcolor{lgray}{122} & \textcolor{lgray}{118} & \textcolor{lgray}{121} & \textcolor{lgray}{119} \\
        & \luarAV & 0.667 & 0.755 & 0.692 & 0.643 & 0.750 & \textcolor{lgray}{180} & \textcolor{lgray}{60} & \textcolor{lgray}{100} & \textcolor{lgray}{140} \\ 
        & \luarAVOrg & \underline{0.706} & \underline{0.773} & 0.707 & 0.705 & 0.708 & \textcolor{lgray}{170} & \textcolor{lgray}{70} & \textcolor{lgray}{71} & \textcolor{lgray}{169} \\\midrule
        %-------------------------------------------------------------------- 

        \pagebreak
        %========================================================	  
        % Add header for better readability due to table fragmentation...
        &  \textbf{Method} & \textbf{Acc.} & \textbf{AUC} & \textbf{F1} & \textbf{Prec.} & \textbf{Rec.} &\color{lgray}\textbf{TP} & \color{lgray}\textbf{FN} & \color{lgray}\textbf{FP} & \color{lgray}\textbf{TN} \\\midrule	
        %========================================================	
        
        & \cellcolor{lyellow}{\ourApproach} & \cellcolor{lyellow}{\textbf{0.765}} & \cellcolor{lyellow}{\textbf{0.839}} & \cellcolor{lyellow}{0.769} & \cellcolor{lyellow}{0.783} & \cellcolor{lyellow}{0.755} &\textcolor{lgray}{188} & \textcolor{lgray}{52} & \textcolor{lgray}{61} & \textcolor{lgray}{179} \\
        & \imOrg & 0.650 & 0.735 & 0.613 & 0.554 & 0.686 &\textcolor{lgray}{133} & \textcolor{lgray}{107} & \textcolor{lgray}{61} & \textcolor{lgray}{179} \\
        & \coav & 0.638 & 0.728 & 0.640 & 0.646 & 0.635 &\textcolor{lgray}{155} & \textcolor{lgray}{85} & \textcolor{lgray}{89} & \textcolor{lgray}{151} \\
        & \taveer & 0.671 & 0.749 & 0.686 & 0.721 & 0.655 &\textcolor{lgray}{173} & \textcolor{lgray}{67} & \textcolor{lgray}{91} & \textcolor{lgray}{149} \\
        & \janithDiffVecLR & 0.508 & 0.514 & 0.561 & 0.507 & 0.629 & \textcolor{lgray}{151} & \textcolor{lgray}{89} & \textcolor{lgray}{147} & \textcolor{lgray}{93} \\
        & \janithDiffVecMLP & 0.508 & 0.507 & 0.543 & 0.507 & 0.583 & \textcolor{lgray}{140} & \textcolor{lgray}{100} & \textcolor{lgray}{136} & \textcolor{lgray}{104} \\      
        & \adhominem & 0.537 & 0.568 & 0.456 & 0.554 & 0.388 & \textcolor{lgray}{93} & \textcolor{lgray}{147} & \textcolor{lgray}{75} & \textcolor{lgray}{165} \\
        \multirow{-5}{*}{\large\rotatebox{90}{$\bm{\CorpusYelp}$}}
        & \adhominemOrg & 0.521 & 0.588 & 0.498 & 0.523 & 0.475 & \textcolor{lgray}{114} & \textcolor{lgray}{126} & \textcolor{lgray}{104} & \textcolor{lgray}{136} \\
        & \luarAV & 0.706 & 0.782 & 0.706 & 0.707 & 0.704 & \textcolor{lgray}{169} & \textcolor{lgray}{71} & \textcolor{lgray}{70} & \textcolor{lgray}{170} \\
        & \luarAVOrg & \underline{0.746} & \underline{0.834} & 0.754 & 0.730 & 0.779 & \textcolor{lgray}{187} & \textcolor{lgray}{53} & \textcolor{lgray}{69} & \textcolor{lgray}{171} \\\midrule
        %--------------------------------------------------------------------  
        & \cellcolor{lyellow}{\ourApproach} & \cellcolor{lyellow}{\underline{0.796}} & \cellcolor{lyellow}{\underline{0.874}} & \cellcolor{lyellow}{0.791} & \cellcolor{lyellow}{0.772} & \cellcolor{lyellow}{0.811} &\textcolor{lgray}{618} & \textcolor{lgray}{182} & \textcolor{lgray}{144} & \textcolor{lgray}{656} \\
        & \imOrg & 0.778 & 0.858 & 0.783 & 0.798 & 0.768 &\textcolor{lgray}{638} & \textcolor{lgray}{162} & \textcolor{lgray}{193} & \textcolor{lgray}{607} \\
        & \coav & 0.730 & 0.802 & 0.730 & 0.729 & 0.731 &\textcolor{lgray}{583} & \textcolor{lgray}{217} & \textcolor{lgray}{215} & \textcolor{lgray}{585} \\
        & \taveer & 0.691 & 0.752 & 0.668 & 0.621 & 0.722 &\textcolor{lgray}{497} & \textcolor{lgray}{303} & \textcolor{lgray}{191} & \textcolor{lgray}{609} \\
        & \janithDiffVecLR & 0.618 & 0.644 & 0.639 & 0.606 & 0.675 & \textcolor{lgray}{540} & \textcolor{lgray}{260} & \textcolor{lgray}{351} & \textcolor{lgray}{449} \\
        & \janithDiffVecMLP & 0.563 & 0.593 & 0.566 & 0.562 & 0.569 & \textcolor{lgray}{455} & \textcolor{lgray}{345} & \textcolor{lgray}{354} & \textcolor{lgray}{446} \\     
        & \adhominem & 0.541 & 0.553 & 0.525 & 0.544 & 0.507 & \textcolor{lgray}{406} & \textcolor{lgray}{394} & \textcolor{lgray}{341} & \textcolor{lgray}{459} \\
        \multirow{-5}{*}{\large\rotatebox{90}{$\bm{\CorpusIMDB}$}}
        & \adhominemOrg & 0.581 & 0.638 & 0.552 & 0.594 & 0.515 & \textcolor{lgray}{412} & \textcolor{lgray}{388} & \textcolor{lgray}{282} & \textcolor{lgray}{518} \\
        & \luarAV & 0.771 & 0.859 & 0.764 & 0.788 & 0.743 & \textcolor{lgray}{594} & \textcolor{lgray}{206} & \textcolor{lgray}{160} & \textcolor{lgray}{640} \\
        & \luarAVOrg & \textbf{0.827} & \textbf{0.911} & 0.825 & 0.835 & 0.815 & \textcolor{lgray}{652} & \textcolor{lgray}{148} & \textcolor{lgray}{129} & \textcolor{lgray}{671} \\\midrule
        
        %-------------------------------------------------------------------- 
        & \cellcolor{lyellow}{\ourApproach} & \cellcolor{lyellow}{\underline{0.891}} & \cellcolor{lyellow}{\underline{0.954}} & \cellcolor{lyellow}{0.891} & \cellcolor{lyellow}{0.892} & \cellcolor{lyellow}{0.890} &\textcolor{lgray}{803} & \textcolor{lgray}{97} & \textcolor{lgray}{99} & \textcolor{lgray}{801} \\
        & \imOrg & 0.827 & 0.908 & 0.812 & 0.750 & 0.886 &\textcolor{lgray}{675} & \textcolor{lgray}{225} & \textcolor{lgray}{87} & \textcolor{lgray}{813} \\
        & \coav & 0.781 & 0.867 & 0.789 & 0.820 & 0.761 &\textcolor{lgray}{738} & \textcolor{lgray}{162} & \textcolor{lgray}{232} & \textcolor{lgray}{668} \\
        & \taveer & 0.781 & 0.867 & 0.778 & 0.767 & 0.789 &\textcolor{lgray}{690} & \textcolor{lgray}{210} & \textcolor{lgray}{185} & \textcolor{lgray}{715} \\
        & \janithDiffVecLR & 0.839 & 0.921 & 0.832 & 0.873 & 0.794 & \textcolor{lgray}{715} & \textcolor{lgray}{185} & \textcolor{lgray}{104} & \textcolor{lgray}{796} \\
        & \janithDiffVecMLP & 0.810 & 0.882 & 0.811 & 0.805 & 0.818 & \textcolor{lgray}{736} & \textcolor{lgray}{164} & \textcolor{lgray}{178} & \textcolor{lgray}{722} \\      
        & \adhominem & 0.601 & 0.647 & 0.550 & 0.630 & 0.488 & \textcolor{lgray}{439} & \textcolor{lgray}{461} & \textcolor{lgray}{258} & \textcolor{lgray}{642} \\
        \multirow{-5}{*}{\large\rotatebox{90}{$\bm{\CorpusBlogs}$}}
        & \adhominemOrg & 0.647 & 0.718 & 0.612 & 0.680 & 0.557 & \textcolor{lgray}{501} & \textcolor{lgray}{399} & \textcolor{lgray}{236} & \textcolor{lgray}{664} \\
        & \luarAV & 0.792 & 0.875 & 0.792 & 0.789 & 0.796 & \textcolor{lgray}{716} & \textcolor{lgray}{184} & \textcolor{lgray}{191} & \textcolor{lgray}{709} \\ 
        & \luarAVOrg & \textbf{0.902} & \textbf{0.962} & 0.904 & 0.888 & 0.921 & \textcolor{lgray}{829} & \textcolor{lgray}{71} & \textcolor{lgray}{105} & \textcolor{lgray}{795} \\\midrule
        %-------------------------------------------------------------------- 

        & \cellcolor{lyellow}{\ourApproach} & \cellcolor{lyellow}{\textbf{0.921}} & \cellcolor{lyellow}{\textbf{0.977}} & \cellcolor{lyellow}{0.921} & \cellcolor{lyellow}{0.922} & \cellcolor{lyellow}{0.919} &\textcolor{lgray}{1107} & \textcolor{lgray}{93} & \textcolor{lgray}{97} & \textcolor{lgray}{1103} \\
        & \imOrg & 0.856 & 0.932 & 0.855 & 0.851 & 0.859 &\textcolor{lgray}{1021} & \textcolor{lgray}{179} & \textcolor{lgray}{167} & \textcolor{lgray}{1033} \\
        & \coav & 0.801 & 0.884 & 0.805 & 0.820 & 0.790 &\textcolor{lgray}{984} & \textcolor{lgray}{216} & \textcolor{lgray}{261} & \textcolor{lgray}{939} \\
        & \taveer & 0.842 & 0.915 & 0.839 & 0.822 & 0.857 &\textcolor{lgray}{986} & \textcolor{lgray}{214} & \textcolor{lgray}{164} & \textcolor{lgray}{1036} \\
        & \janithDiffVecLR & \underline{0.901} & \underline{0.968} & 0.904 & 0.876 & 0.934 & \textcolor{lgray}{1121} & \textcolor{lgray}{79} & \textcolor{lgray}{159} & \textcolor{lgray}{1041} \\
        & \janithDiffVecMLP & 0.863 & 0.937 & 0.864 & 0.855 & 0.874 & \textcolor{lgray}{1049} & \textcolor{lgray}{151} & \textcolor{lgray}{178} & \textcolor{lgray}{1022} \\     
        & \adhominem & 0.586 & 0.619 & 0.578 & 0.590 & 0.568 & \textcolor{lgray}{681} & \textcolor{lgray}{519} & \textcolor{lgray}{474} & \textcolor{lgray}{726} \\
        & \adhominemOrg & 0.671 & 0.759 & 0.622 & 0.732 & 0.540 & \textcolor{lgray}{648} & \textcolor{lgray}{552} & \textcolor{lgray}{237} & \textcolor{lgray}{963} \\
        \multirow{-5}{*}{\large\rotatebox{90}{$\bm{\CorpusAmazon}$}}
        & \luarAV & 0.829 & 0.912 & 0.833 & 0.813 & 0.854 & \textcolor{lgray}{1025} & \textcolor{lgray}{175} & \textcolor{lgray}{235} & \textcolor{lgray}{965} \\ 
        & \luarAVOrg & 0.876 & 0.944 & 0.876 & 0.875 & 0.877 & \textcolor{lgray}{1052} & \textcolor{lgray}{148} & \textcolor{lgray}{150} & \textcolor{lgray}{1050} \\\midrule
        %-------------------------------------------------------------------- 
    
        & \cellcolor{lyellow}{\ourApproach} & \cellcolor{lyellow}{\underline{0.705}} & \cellcolor{lyellow}{\underline{0.777}} & \cellcolor{lyellow}{0.689} & \cellcolor{lyellow}{0.652} & \cellcolor{lyellow}{0.729} &\textcolor{lgray}{868} & \textcolor{lgray}{463} & \textcolor{lgray}{322} & \textcolor{lgray}{1009} \\
        & \imOrg & 0.676 & 0.739 & 0.616 & 0.520 & 0.756 &\textcolor{lgray}{692} & \textcolor{lgray}{639} & \textcolor{lgray}{223} & \textcolor{lgray}{1108} \\
        & \coav & 0.633 & 0.701 & 0.638 & 0.648 & 0.629 &\textcolor{lgray}{863} & \textcolor{lgray}{468} & \textcolor{lgray}{510} & \textcolor{lgray}{821} \\
        & \taveer & 0.635 & 0.682 & 0.614 & 0.581 & 0.651 &\textcolor{lgray}{773} & \textcolor{lgray}{558} & \textcolor{lgray}{414} & \textcolor{lgray}{917} \\
        & \janithDiffVecLR & 0.663 & 0.729 & 0.696 & 0.634 & 0.772 & \textcolor{lgray}{1028} & \textcolor{lgray}{303} & \textcolor{lgray}{593} & \textcolor{lgray}{738} \\
        & \janithDiffVecMLP & 0.634 & 0.693 & 0.637 & 0.632 & 0.643 & \textcolor{lgray}{856} & \textcolor{lgray}{475} & \textcolor{lgray}{499} & \textcolor{lgray}{832} \\       
        & \adhominem & 0.506 & 0.511 & 0.572 & 0.505 & 0.660 & \textcolor{lgray}{879} & \textcolor{lgray}{452} & \textcolor{lgray}{862} & \textcolor{lgray}{469} \\
        \multirow{-5}{*}{\large\rotatebox{90}{$\bm{\CorpusAllTheNews}$}}
        & \adhominemOrg & 0.551 & 0.549 & 0.610 & 0.539 & 0.702 & \textcolor{lgray}{935} & \textcolor{lgray}{396} & \textcolor{lgray}{799} & \textcolor{lgray}{532} \\
        & \luarAV & 0.583 & 0.624 & 0.576 & 0.586 & 0.566 & \textcolor{lgray}{753} & \textcolor{lgray}{578} & \textcolor{lgray}{531} & \textcolor{lgray}{800} \\
        & \luarAVOrg & \textbf{0.812} & \textbf{0.890} & 0.808 & 0.824 & 0.793 & \textcolor{lgray}{1055} & \textcolor{lgray}{276} & \textcolor{lgray}{225} & \textcolor{lgray}{1106} \\
        %========================================================
        \bottomrule
    \end{longtable}
\end{center}

\begin{figure}[h!]
    \centering
    \includegraphics[width=1\linewidth]{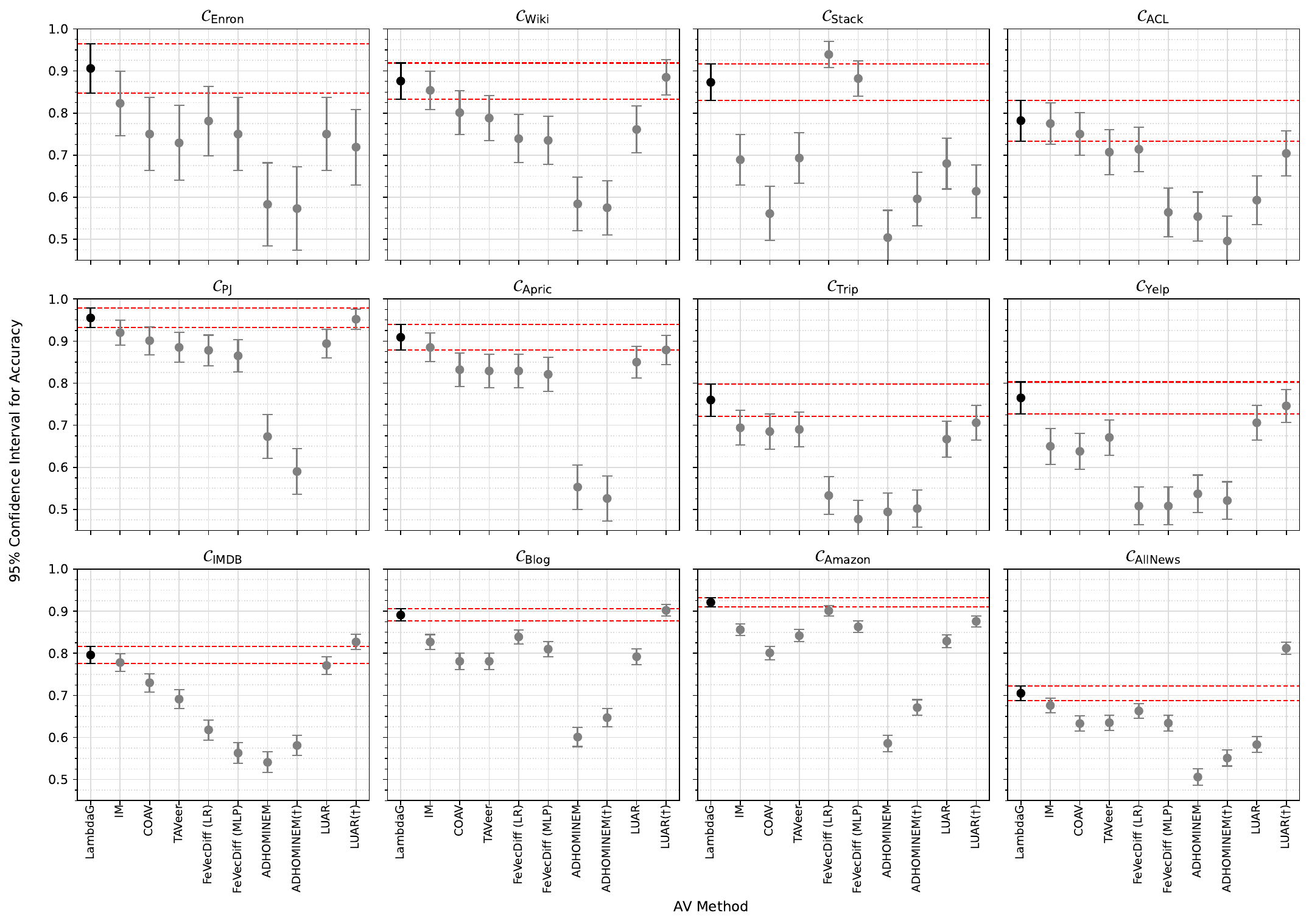} 
    \caption{Additional evaluation results showing the 95\% confidence intervals for Accuracy.}
    \label{fig:errorbars}
\end{figure}

In addition to Table~\ref{tab:EvalResults}, Figure~\ref{fig:errorbars} shows 95\% confidence intervals for Accuracy, calculated using the classical standard error estimate for a proportion, under the simplifying assumption of \textit{independent and identically distributed} (i.i.d.) test data:
\begin{equation} \label{eq:accuracySE}
\text{SE} = \sqrt{\frac{1}{n}\Bigl(\text{Accuracy} \times (1 - \text{Accuracy})}\Bigr)
\end{equation}

Here, $n$ indicates the total number of tests performed. Even though there is no obvious reason to take the i.i.d. assumption seriously due to the non-trivial sampling process of our test datasets (see Section~\ref{sec-data-sets}) and the relatively high number of potential comparisons involved, we can consider the resulting standard errors as a rough indication of the potential size of statistical fluctuations and an assessment of whether the observed performance improvements really suggest the presence of actual effects.

Considering the overall results, \ourApproach outperforms the other \AV baselines, ranking first on 7 of the 12 corpora and second on 4, with respect to Accuracy and AUC. This consistency across many corpora representing different \AV scenarios in terms of lengths and genres is a strong suggestion that this effect is robust and replicable. \ourApproach’s superiority is the most clear for $\CorpusEnron$, $\CorpusTripAdvisor$ and $\CorpusAmazon$. For all corpora, \ourApproach performs better than \adhominemOrg, even though this method could have exploited any correlation between author and topic in each corpus.
\ourApproach achieved similar performance to \luarAVOrg, that is \luarAV when applied to content-unmasked datasets, for $\CorpusWiki$, $\CorpusPJ$, $\CorpusYelp$, and $\CorpusBlogs$. Only for $\CorpusIMDB$ and $\CorpusAllTheNews$ the performance gap between \luarAVOrg and \ourApproach is considerable and probably statistically significant. It is important to note, however, that contrary to \ourApproach, \luarAVOrg also exploits the information in content words, which means that it could be affected by inherent topic biases. Sensitivity to topic-bias can be somewhat assessed by considering the results on $\CorpusStack$, which is designed to be cross-topic, and where \luarAVOrg's performance decreases significantly. Compared to \luarAV finetuned on \posNoise-processed data, \ourApproach's performance appears to be significantly greater in all cases except for $\CorpusIMDB$, $\CorpusYelp$, and possibly $\CorpusApricity$. 
In $\CorpusStack$, \janithDiffVecLR attains the highest accuracy, making it the only linear model to outperform \ourApproach outright in a corpus, albeit by a narrow margin.
Finally, it is worth mentioning that \imOrg appears as a strong baseline, in particular for $\CorpusEnron$, $\CorpusWiki$, $\CorpusACL$, $\CorpusApricity$, and $\CorpusIMDB$.

Table~\ref{tab:calib} instead shows the values of \cllr and \cllrmin for both $\lambda_G$ and $\Lambda_G$, since both are log-likelihood ratios. A significant discrepancy is observed between the \cllr and the \cllrmin values for $\lambda_G$, indicating that $\lambda_G$ is uncalibrated. This means that although higher values of $\lambda_G$ do correctly correspond to \classYdash{cases}, the scale of variation does not reflect the expectations of a perfectly calibrated system, where $\lambda_G = 0$ means an inconclusive result, a positive value suggests a \classYdash{case}, and a negative value suggests an \classNdash{case}. When $\lambda_G$ is turned into $\Lambda_G$ by fitting a logistic regression on training data, however, the lower values of $\cllr(\Lambda_G)$ are evidence that $\Lambda_G$ is well calibrated.

\begin{table}[h!]
    \centering
    \begin{tabular}{lrrr}
        \toprule
         Corpus &  $\cllr(\lambda_G)$ & $\cllr(\Lambda_G)$ & \cllrmin \\
         \midrule
         $\CorpusEnron$ & 3.630  & 0.432 & 0.315\\
         $\CorpusWiki$ & 2.026  & 0.457 & 0.405\\
         $\CorpusStack$ & 5.716 & 0.499 & 0.411\\
         $\CorpusACL$ & 18.038  & 0.885 & 0.636\\
         $\CorpusPJ$ & 2.000  & 0.254 & 0.143\\
         $\CorpusApricity$ & 2.811 & 0.397 & 0.337\\
         $\CorpusTripAdvisor$ & 2.666  & 0.752 & 0.715\\
         $\CorpusYelp$ & 1.872  & 0.726 & 0.691\\
         $\CorpusIMDB$ & 2.714  & 0.648 & 0.626\\
         $\CorpusBlogs$ & 2.969 & 0.424 & 0.403\\
         $\CorpusAmazon$ & 1.604 & 0.302 & 0.286\\
         $\CorpusAllTheNews$ & 9.920 & 0.812 & 0.797\\
         \bottomrule
    \end{tabular}
    \caption{Values of \cllr and \cllrmin for $\lambda_G$ and for $\Lambda_G$ generated by calibrating $\lambda_G$ with the logistic regression calibration method. \cllrmin is identical for both $\lambda_G$ and $\Lambda_G$.}
    \label{tab:calib}
\end{table}

\begin{figure}[h!]
    \centering
    \includegraphics[width=1\linewidth]{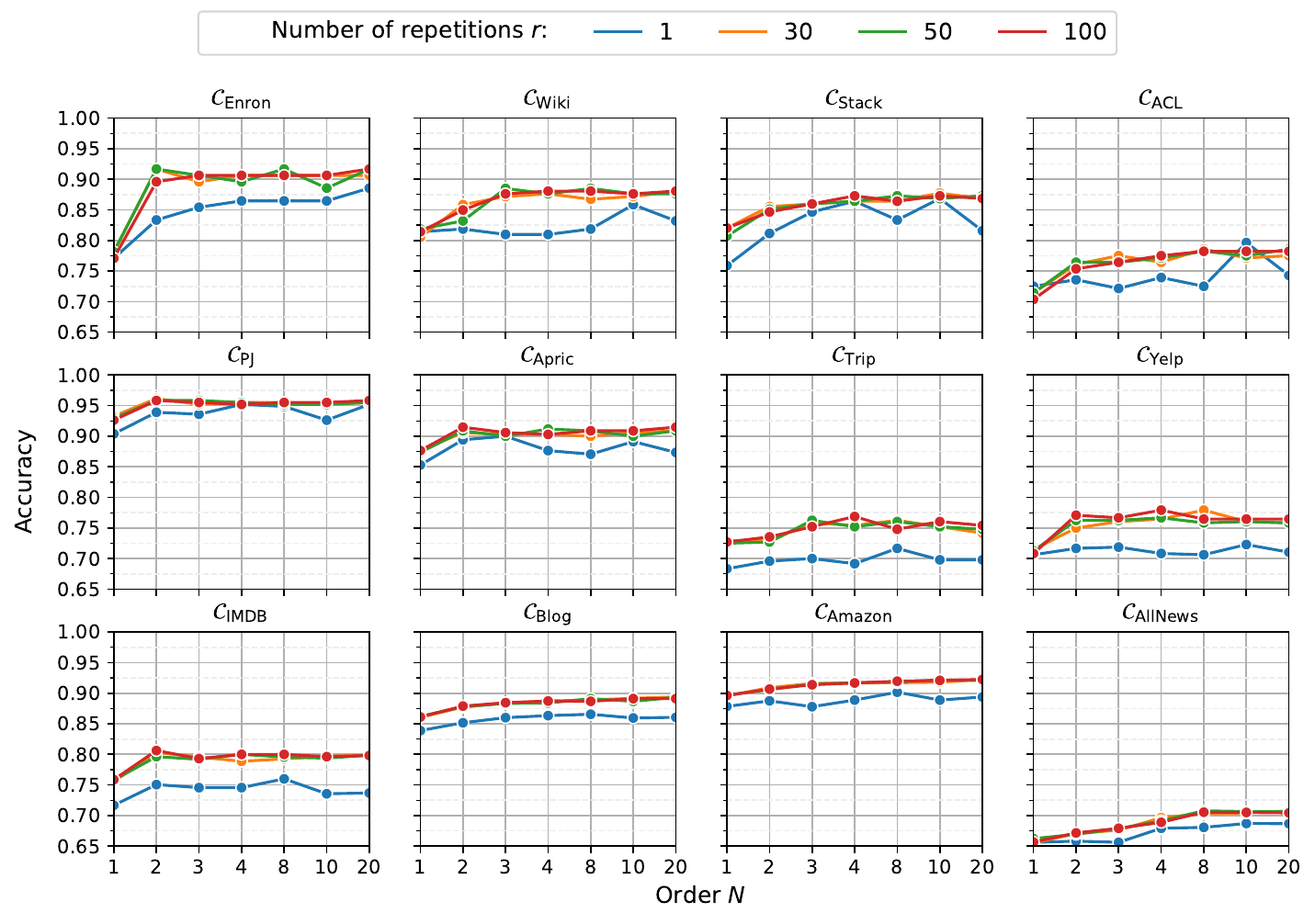}
    \caption{Variation in Accuracy depending on the number of repetitions, $\numberRepetitions$, and orders of the \ngram model, $\modelOrder$.}
    \label{fig:hyperparameters}
\end{figure}

All results shown in Table~\ref{tab:EvalResults} and Table~\ref{tab:calib} were obtained with \ourApproach using the hyperparameters $\numberRepetitions=100$ and $\modelOrder=10$. Figure~\ref{fig:hyperparameters} shows how changes in these hyperparameters affects Accuracy, revealing that the method is robust in the choice of these hyperparameters. It is clear from the graph that $\numberRepetitions = 1$, meaning that only one reference Grammar Model $G_{\text{ref}}$ is extracted from the reference corpus, is not enough to achieve optimal results. 
%------------------------------------------ 
However, in most corpora, setting $\numberRepetitions=100$ is unnecessary, as performance stabilizes at $\numberRepetitions = 30$. Regarding the other hyperparameter $\modelOrder$, the order of the \ngram model, the value $\modelOrder=10$ was chosen to ensure that grammatical dependencies are captured even for long sentences and Figure~\ref{fig:hyperparameters} does suggest that this assumption is correct. For example, for a corpus such as $\CorpusPJ$ consisting of chat messages or $\CorpusEnron$ consisting of emails, $\modelOrder=2$ is enough. In contrast, for the academic papers in $\CorpusACL$ and the news in $\CorpusAllTheNews$ a larger order of the model clearly has an advantage. In all corpora, however, there is no gain in increasing $\modelOrder$ to 20.  

Another important robustness test that we performed is the selection of the reference data. Reliance on external data is a potential limitation for all binary extrinsic \AV methods such as \imOrg. As explained in Section~\ref{sec-introduction}, \imOrg, suffers a substantial loss in performance if the impostors are selected from another genre. \ourApproach has the same crucial requirement that a reference corpus $\Dref$ is collected and it is presumably important that $\Dref$ is as compatible as possible, especially in genre, with $\DA$ and $\Dunk$. To test the robustness of \ourApproach to variations in $\Dref$ type we ran 144 tests by pairing each of the twelve corpora to every other corpora. For example, we performed the analysis for $\CorpusEnron$ but used $\CorpusACL$ as $\Dref$. Then, we calculated Accuracy and \cllr for all combinations and created a plot for each metric displaying the resulting drop in performances when using a different $\Dref$ (Figure~\ref{fig:crossgenre}). 
%-----------------------------------------
The diagonal of Figure~\ref{fig:crossgenre} (top) shows the same results reported in Table~\ref{tab:EvalResults}, which function as the benchmark of performance for each row of the plot, corresponding to the corpus being tested. 

These results show that, as expected, the performance of \ourApproach can be strongly affected by genre variations in the reference corpus. The overall performance loss, however, is not as substantial as might be expected for a method such as \imOrg \parencite{HalvaniPhD:2021}. Firstly, for reasonable cross-corpus comparisons such as using $\CorpusYelp$ as reference for problems in $\CorpusAmazon$, where both corpora are reviews, the loss in Accuracy is small. Secondly and more surprisingly, ‘unreasonable’ comparisons such using academic papers from $\CorpusACL$ as reference data for problems in a corpus of chat logs such as $\CorpusPJ$ leads to a reduction of, relatively speaking, only 0.27 Accuracy. Such a loss is clearly substantial but the expectation given the literature is that the method should perform at random in such a case. This is a comparison that a human analyst would be unlikely to perform and therefore it is a real test of the lower limits in terms of robustness to reference corpora.

Examining Figure~\ref{fig:crossgenre} (top) row-wise, the most notable pattern is that the two corpora that suffer the most performance loss overall are $\CorpusPJ$ and $\CorpusStack$. For $\CorpusPJ$ the explanation is likely to be that it is the only corpus with sentences corresponding to chat messages, and therefore any match of this corpus with other corpora suffers from this important difference. The chat logs in $\CorpusPJ$ are also the texts with the lowest level of formality and highest degree of linguistic freedom among all corpora. For $\CorpusStack$ this result could be due to its hybrid genre character (forum posts on academic topics) or it could simply be due the use of several tags to mask mathematical expressions or computer code. When looking at Figure~\ref{fig:crossgenre} column-wise, instead, the most notable pattern is that using $\CorpusACL$ or $\CorpusAllTheNews$ as reference corpora tends to affect performance more drastically for the majority of corpora, which again could be explained by their relatively higher degree of formality compared to the other corpora. All in all, this analysis suggests that, for $\CorpusPJ$ and $\CorpusStack$ the reference corpus has a moderate impact on performance, while for all other corpora it is only the use of $\CorpusACL$ or $\CorpusAllTheNews$ as a reference where this modest impact can be observed. For most other cross-corpus comparisons, the effect of using a different reference corpus is tolerable.

We also present the results for the \cllr metric in Figure~\ref{fig:crossgenre} (bottom). As explained in Section~\ref{sec-the-likelihood-ratio-framework}, a \cllr value close to or equal to one means that the system is not providing any useful information, while a \cllr value higher than 1 means that results tend to be misleading (\eg negative $\Lambda_G$ for a \classYdash{problem}). None of the cross-genre comparisons produce \cllr higher than one, meaning that in all cases, on average, the values of $\Lambda_G$ are not misleading the analyst but can be, at worst, inconclusive with $\Lambda_G \approx 0$. These results support the principle that an analyst should always compile a reference data set that is as close as possible to the case data \parencite{ishihara_validation_2024}. However, the results do also suggest that \ourApproach is certainly stable to small variations in genre or reference corpus composition.

The final tests we carried out relate to the contribution of \posNoise to the algorithm. We firstly performed a comparison of \ourApproach as implemented in Algorithm~\ref{AlgoGMComp} vs. \ourApproach without applying the \posNoise function. The rationale for this test is to understand whether more authorial information can be extracted and exploited using this method beyond just grammatical structure. The results in Table~\ref{tab:EvalResultsPosnoise} show that, regarding Accuracy and \auc, in all cases the proposed version of Algorithm~\ref{AlgoGMComp} with the \posNoise preprocessing is better performing than a version of the algorithm without this preprocessing. The difference in Accuracy is modest but consistent across all corpora, indicating that the method is clearly superior when the focus is entirely on grammar. 

%--------------------------------------------

%-------------------------------------
\definecolor{lgray}{gray}{0.6}
\definecolor{lyellow}{HTML}{fff7bc}
\definecolor{lightgray}{rgb}{0.85, 0.85, 0.85}
%-------------------------------------
\setlength{\aboverulesep}{0pt}
\setlength{\belowrulesep}{0pt}
\setlength{\extrarowheight}{.15ex}
%-------------------------------------
\newcolumntype{g}{>{\columncolor{lightgray}}r}
%-------------------------------------
\begin{table}[h!]
    \centering
	\begin{tabular}{p{01.3cm}lrrrrrrrrr} 
    \toprule
    %========================================================
    &  \textbf{$\Delta$Acc.} & \textbf{$\Delta$AUC} & \textbf{$\Delta$F1} & \textbf{$\Delta$Prec.} & \textbf{$\Delta$Rec.} &\color{lgray}\textbf{$\Delta$TP} & \color{lgray}\textbf{$\Delta$FN} & \color{lgray}\textbf{$\Delta$FP} & \color{lgray}\textbf{$\Delta$TN} \\\midrule	
    
    $\CorpusEnron$ & {\cellcolor[HTML]{F6F9FD}} \color[HTML]{000000} 0.031 & {\cellcolor[HTML]{F2F7FC}} \color[HTML]{000000} 0.039 & {\cellcolor[HTML]{F2F7FC}} \color[HTML]{000000} 0.039 & {\cellcolor[HTML]{FFF9F7}} \color[HTML]{000000} -0.017 & {\cellcolor[HTML]{E5EFF8}} \color[HTML]{000000} 0.083 & \color{lgray} 4 & \color{lgray} -4 & \color{lgray} 1 & \color{lgray} -1 \\
    $\CorpusWiki$ & {\cellcolor[HTML]{F7FAFD}} \color[HTML]{000000} 0.026 & {\cellcolor[HTML]{FEFEFF}} \color[HTML]{000000} 0.003 & {\cellcolor[HTML]{F6F9FD}} \color[HTML]{000000} 0.029 & {\cellcolor[HTML]{F7FAFD}} \color[HTML]{000000} 0.022 & {\cellcolor[HTML]{F4F8FC}} \color[HTML]{000000} 0.036 & \color{lgray} 4 & \color{lgray} -4 & \color{lgray} -2 & \color{lgray} 2 \\
    $\CorpusStack$ & {\cellcolor[HTML]{FEFEFF}} \color[HTML]{000000} 0.005 & {\cellcolor[HTML]{FEFEFF}} \color[HTML]{000000} 0.001 & {\cellcolor[HTML]{FFFEFE}} \color[HTML]{000000} -0.002 & {\cellcolor[HTML]{F1F6FB}} \color[HTML]{000000} 0.047 & {\cellcolor[HTML]{FEF1EA}} \color[HTML]{000000} -0.044 & \color{lgray} -5 & \color{lgray} 5 & \color{lgray} -6 & \color{lgray} 6 \\
    $\CorpusACL$ & {\cellcolor[HTML]{EDF4FB}} \color[HTML]{000000} 0.057 & {\cellcolor[HTML]{EFF5FB}} \color[HTML]{000000} 0.050 & {\cellcolor[HTML]{F6F9FD}} \color[HTML]{000000} 0.031 & {\cellcolor[HTML]{DEEBF7}} \color[HTML]{000000} 0.106 & {\cellcolor[HTML]{FEEFE8}} \color[HTML]{000000} -0.050 & \color{lgray} -7 & \color{lgray} 7 & \color{lgray} -23 & \color{lgray} 23 \\
    $\CorpusPJ$ & {\cellcolor[HTML]{FCFDFE}} \color[HTML]{000000} 0.009 & {\cellcolor[HTML]{FEFEFF}} \color[HTML]{000000} 0.005 & {\cellcolor[HTML]{FCFDFE}} \color[HTML]{000000} 0.010 & {\cellcolor[HTML]{FEFEFF}} \color[HTML]{000000} 0.001 & {\cellcolor[HTML]{F9FBFE}} \color[HTML]{000000} 0.019 & \color{lgray} 3 & \color{lgray} -3 & \color{lgray} 0 & \color{lgray} 0 \\
    $\CorpusApricity$ & {\cellcolor[HTML]{E6F0F9}} \color[HTML]{000000} 0.077 & {\cellcolor[HTML]{F2F7FC}} \color[HTML]{000000} 0.042 & {\cellcolor[HTML]{E5EFF8}} \color[HTML]{000000} 0.084 & {\cellcolor[HTML]{F4F8FC}} \color[HTML]{000000} 0.038 & {\cellcolor[HTML]{D7E7F5}} \color[HTML]{000000} 0.129 & \color{lgray} 22 & \color{lgray} -22 & \color{lgray} -4 & \color{lgray} 4 \\
    $\CorpusTripAdvisor$ & {\cellcolor[HTML]{F4F8FC}} \color[HTML]{000000} 0.037 & {\cellcolor[HTML]{F1F6FB}} \color[HTML]{000000} 0.049 & {\cellcolor[HTML]{F1F6FB}} \color[HTML]{000000} 0.048 & {\cellcolor[HTML]{F7FAFD}} \color[HTML]{000000} 0.022 & {\cellcolor[HTML]{EAF2FA}} \color[HTML]{000000} 0.071 & \color{lgray} 17 & \color{lgray} -17 & \color{lgray} -1 & \color{lgray} 1 \\
    $\CorpusYelp$ & {\cellcolor[HTML]{EDF4FB}} \color[HTML]{000000} 0.057 & {\cellcolor[HTML]{F1F6FB}} \color[HTML]{000000} 0.046 & {\cellcolor[HTML]{EBF3FA}} \color[HTML]{000000} 0.063 & {\cellcolor[HTML]{F2F7FC}} \color[HTML]{000000} 0.043 & {\cellcolor[HTML]{E5EFF8}} \color[HTML]{000000} 0.083 & \color{lgray} 20 & \color{lgray} -20 & \color{lgray} -7 & \color{lgray} 7 \\
    $\CorpusIMDB$ & {\cellcolor[HTML]{E6F0F9}} \color[HTML]{000000} 0.080 & {\cellcolor[HTML]{E0ECF7}} \color[HTML]{000000} 0.101 & {\cellcolor[HTML]{E3EEF8}} \color[HTML]{000000} 0.091 & {\cellcolor[HTML]{EAF2FA}} \color[HTML]{000000} 0.070 & {\cellcolor[HTML]{DEEBF7}} \color[HTML]{000000} 0.108 & \color{lgray} 87 & \color{lgray} -87 & \color{lgray} -42 & \color{lgray} 42 \\
    $\CorpusBlogs$ & {\cellcolor[HTML]{EBF3FA}} \color[HTML]{000000} 0.063 & {\cellcolor[HTML]{F4F8FC}} \color[HTML]{000000} 0.038 & {\cellcolor[HTML]{EAF2FA}} \color[HTML]{000000} 0.066 & {\cellcolor[HTML]{F1F6FB}} \color[HTML]{000000} 0.048 & {\cellcolor[HTML]{E5EFF8}} \color[HTML]{000000} 0.083 & \color{lgray} 75 & \color{lgray} -75 & \color{lgray} -38 & \color{lgray} 38 \\
    $\CorpusAmazon$ & {\cellcolor[HTML]{F1F6FB}} \color[HTML]{000000} 0.046 & {\cellcolor[HTML]{F6F9FD}} \color[HTML]{000000} 0.030 & {\cellcolor[HTML]{F1F6FB}} \color[HTML]{000000} 0.046 & {\cellcolor[HTML]{F2F7FC}} \color[HTML]{000000} 0.041 & {\cellcolor[HTML]{EFF5FB}} \color[HTML]{000000} 0.050 & \color{lgray} 60 & \color{lgray} -60 & \color{lgray} -49 & \color{lgray} 49 \\
    $\CorpusAllTheNews$ & {\cellcolor[HTML]{F2F7FC}} \color[HTML]{000000} 0.043 & {\cellcolor[HTML]{F1F6FB}} \color[HTML]{000000} 0.049 & {\cellcolor[HTML]{EDF4FB}} \color[HTML]{000000} 0.059 & {\cellcolor[HTML]{F4F8FC}} \color[HTML]{000000} 0.034 & {\cellcolor[HTML]{E8F1F9}} \color[HTML]{000000} 0.076 & \color{lgray} 101 & \color{lgray} -101 & \color{lgray} -15 & \color{lgray} 15 \\
    %========================================================
	\bottomrule	
         \caption{Differences in evaluation results when applying \posNoise in \ourApproach compared to omitting it. Positive values (shaded in blue) indicate results improvement, while negative values (shaded in red) indicate results degradation by using \posNoise.\label{tab:EvalResultsPosnoise}} 
	\end{tabular} 
\end{table}

\begin{table}[h!]
    \centering
	\begin{tabular}{p{01.3cm}lrrrrrrrrr} 
    \toprule
    %========================================================
    &  \textbf{$\Delta$Acc.} & \textbf{$\Delta$AUC} & \textbf{$\Delta$F1} & \textbf{$\Delta$Prec.} & \textbf{$\Delta$Rec.} &\color{lgray}\textbf{$\Delta$TP} & \color{lgray}\textbf{$\Delta$FN} & \color{lgray}\textbf{$\Delta$FP} & \color{lgray}\textbf{$\Delta$TN} \\\midrule	
    %========================================================	
$\CorpusEnron$ & {\cellcolor[HTML]{FCFDFE}} \color[HTML]{000000} 0.010 & {\cellcolor[HTML]{FCFDFE}} \color[HTML]{000000} 0.010 & {\cellcolor[HTML]{FBFCFE}} \color[HTML]{000000} 0.014 & {\cellcolor[HTML]{FFF9F7}} \color[HTML]{000000} -0.019 & {\cellcolor[HTML]{F2F7FC}} \color[HTML]{000000} 0.042 & \color{lgray} 2 & \color{lgray} -2 & \color{lgray} 1 & \color{lgray} -1 \\
$\CorpusWiki$ & {\cellcolor[HTML]{F7FAFD}} \color[HTML]{000000} 0.026 & {\cellcolor[HTML]{FBFCFE}} \color[HTML]{000000} 0.014 & {\cellcolor[HTML]{F7FAFD}} \color[HTML]{000000} 0.027 & {\cellcolor[HTML]{F4F8FC}} \color[HTML]{000000} 0.036 & {\cellcolor[HTML]{F9FBFE}} \color[HTML]{000000} 0.018 & \color{lgray} 2 & \color{lgray} -2 & \color{lgray} -4 & \color{lgray} 4 \\
$\CorpusStack$ & {\cellcolor[HTML]{FEFEFF}} \color[HTML]{000000} 0.000 & {\cellcolor[HTML]{FFFEFE}} \color[HTML]{000000} -0.004 & {\cellcolor[HTML]{FFFEFE}} \color[HTML]{000000} -0.001 & {\cellcolor[HTML]{FCFDFE}} \color[HTML]{000000} 0.008 & {\cellcolor[HTML]{FFFCFB}} \color[HTML]{000000} -0.009 & \color{lgray} -1 & \color{lgray} 1 & \color{lgray} -1 & \color{lgray} 1 \\
$\CorpusACL$ & {\cellcolor[HTML]{FFF7F4}} \color[HTML]{000000} -0.022 & {\cellcolor[HTML]{FFFBF9}} \color[HTML]{000000} -0.014 & {\cellcolor[HTML]{FFF9F7}} \color[HTML]{000000} -0.019 & {\cellcolor[HTML]{FFF4EF}} \color[HTML]{000000} -0.033 & {\cellcolor[HTML]{FFFCFB}} \color[HTML]{000000} -0.007 & \color{lgray} -1 & \color{lgray} 1 & \color{lgray} 5 & \color{lgray} -5 \\
$\CorpusPJ$ & {\cellcolor[HTML]{FBFCFE}} \color[HTML]{000000} 0.016 & {\cellcolor[HTML]{FEFEFF}} \color[HTML]{000000} 0.000 & {\cellcolor[HTML]{F9FBFE}} \color[HTML]{000000} 0.017 & {\cellcolor[HTML]{FEFEFF}} \color[HTML]{000000} 0.001 & {\cellcolor[HTML]{F6F9FD}} \color[HTML]{000000} 0.032 & \color{lgray} 5 & \color{lgray} -5 & \color{lgray} 0 & \color{lgray} 0 \\
$\CorpusApricity$ & {\cellcolor[HTML]{FFFEFE}} \color[HTML]{000000} -0.003 & {\cellcolor[HTML]{FFFEFE}} \color[HTML]{000000} -0.004 & {\cellcolor[HTML]{FFFEFE}} \color[HTML]{000000} -0.003 & {\cellcolor[HTML]{FEFEFF}} \color[HTML]{000000} 0.000 & {\cellcolor[HTML]{FFFCFB}} \color[HTML]{000000} -0.006 & \color{lgray} -1 & \color{lgray} 1 & \color{lgray} 0 & \color{lgray} 0 \\
$\CorpusTripAdvisor$ & {\cellcolor[HTML]{FBFCFE}} \color[HTML]{000000} 0.014 & {\cellcolor[HTML]{FCFDFE}} \color[HTML]{000000} 0.010 & {\cellcolor[HTML]{F9FBFE}} \color[HTML]{000000} 0.019 & {\cellcolor[HTML]{FCFDFE}} \color[HTML]{000000} 0.007 & {\cellcolor[HTML]{F6F9FD}} \color[HTML]{000000} 0.029 & \color{lgray} 7 & \color{lgray} -7 & \color{lgray} 0 & \color{lgray} 0 \\
$\CorpusYelp$ & {\cellcolor[HTML]{FFFBF9}} \color[HTML]{000000} -0.014 & {\cellcolor[HTML]{FFFCFB}} \color[HTML]{000000} -0.010 & {\cellcolor[HTML]{FFF9F7}} \color[HTML]{000000} -0.019 & {\cellcolor[HTML]{FFFEFE}} \color[HTML]{000000} -0.003 & {\cellcolor[HTML]{FFF4EF}} \color[HTML]{000000} -0.038 & \color{lgray} -9 & \color{lgray} 9 & \color{lgray} -2 & \color{lgray} 2 \\
$\CorpusIMDB$ & {\cellcolor[HTML]{FBFCFE}} \color[HTML]{000000} 0.015 & {\cellcolor[HTML]{FBFCFE}} \color[HTML]{000000} 0.013 & {\cellcolor[HTML]{FBFCFE}} \color[HTML]{000000} 0.014 & {\cellcolor[HTML]{F9FBFE}} \color[HTML]{000000} 0.019 & {\cellcolor[HTML]{FCFDFE}} \color[HTML]{000000} 0.010 & \color{lgray} 8 & \color{lgray} -8 & \color{lgray} -16 & \color{lgray} 16 \\
$\CorpusBlogs$ & {\cellcolor[HTML]{FCFDFE}} \color[HTML]{000000} 0.006 & {\cellcolor[HTML]{FEFEFF}} \color[HTML]{000000} 0.002 & {\cellcolor[HTML]{FCFDFE}} \color[HTML]{000000} 0.006 & {\cellcolor[HTML]{FEFEFF}} \color[HTML]{000000} 0.005 & {\cellcolor[HTML]{FCFDFE}} \color[HTML]{000000} 0.006 & \color{lgray} 6 & \color{lgray} -6 & \color{lgray} -5 & \color{lgray} 5 \\
$\CorpusAmazon$ & {\cellcolor[HTML]{FFFEFE}} \color[HTML]{000000} -0.002 & {\cellcolor[HTML]{FFFEFE}} \color[HTML]{000000} -0.001 & {\cellcolor[HTML]{FFFEFE}} \color[HTML]{000000} -0.003 & {\cellcolor[HTML]{FEFEFF}} \color[HTML]{000000} 0.003 & {\cellcolor[HTML]{FFFCFB}} \color[HTML]{000000} -0.009 & \color{lgray} -10 & \color{lgray} 10 & \color{lgray} -5 & \color{lgray} 5 \\
$\CorpusAllTheNews$ & {\cellcolor[HTML]{F7FAFD}} \color[HTML]{000000} 0.027 & {\cellcolor[HTML]{F6F9FD}} \color[HTML]{000000} 0.031 & {\cellcolor[HTML]{F2F7FC}} \color[HTML]{000000} 0.040 & {\cellcolor[HTML]{FBFCFE}} \color[HTML]{000000} 0.015 & {\cellcolor[HTML]{EDF4FB}} \color[HTML]{000000} 0.057 & \color{lgray} 76 & \color{lgray} -76 & \color{lgray} 5 & \color{lgray} -5 \\
    %========================================================
	\bottomrule	
         \caption{Differences in evaluation results when applying \posNoise in \ourApproach compared to applying the \posNoise without POS-Tagging (\ie using a simple star). Positive values (shaded in blue) indicate results improvement, while negative values (shaded in red) indicate results degradation by using full \posNoise.\label{tab:EvalResultsDistnoise}} 
	\end{tabular} 
\end{table}

%-------------------------------------------- 

Secondly, we evaluated \ourApproach using a variant of \posNoise similar to \textDistortion, but instead of utilizing \posTags, we replaced all non-functional items with asterisks. The results, presented in Table~\ref{tab:EvalResultsDistnoise}, indicate that the \posTag labels contribute very little information overall, implying that a POS tagger might not be necessary.

As explained in Section~\ref{sec-ngram-modelling}, the advantage of \ourApproach is that it can be better interpreted than other methods. The analyst can produce a ranking of all sentences in $\Dunk$ and each token can be color-coded according to its value of $\lambda_G$, similarly to \coav. Some of these visualizations explaining the features discovered are shown in Section~\ref{sec-appendix-a}. 

\section{Discussion}\label{sec-discussion}
We believe that our proposed \ourApproach approach matches many desiderata of an \AV method. The evaluation on our set of \numberOfEvalCorpora datasets demonstrates that \ourApproach works consistently well for relatively short texts in a variety of different genres. In addition, because the method only considers information in functional items, \ourApproach is also minimally affected by potential content biases in the data, such as named entities or topic. In contrast to the other very successful binary-extrinsic \AV method, \imOrg, \ourApproach tends to be much more robust to variations in the reference corpus. Moreover, unlike \imOrg, which relies on the similarity between vectors of thousands of \charNgrams, \ourApproach generates text heat maps that aid the analyst's understanding of the output of the method. Finally, \ourApproach is unique among other \AV methods in being motivated by Cognitive Linguistics and in having a theoretical explanation for its mechanisms.

We believe that its greater scientific plausibility justifies \ourApproach over other \sota methods. Since the pioneering study of  \textcite{mosteller1963}, it is well known that the frequency of functional items distinguish authors and this conclusion is also evident in our results. The reason why this is the case, however, has always been seen as a mystery in the field of authorship analysis and stylometry \parencite{KestemontFunctionWordsAA:2014}. A connection between the unreasonable effectiveness of these functional items and Cognitive Linguistics has been proposed by \textcite{NiniTheoryLingAA:2023} and we believe that the success of \ourApproach is additional evidence for these explanations.

As outlined in Section~\ref{sec-cognitive}, the mental representation of the grammar for an individual can be seen as a a probability model over lexicogrammatical chunks quite like the Grammar Models used in \ourApproach. Although language modeling is seen in computational linguistics as a tool and not necessarily a realistic representation of language, evidence from Cognitive Linguistics suggests that, in reality, a language model, while not being realistic as a model of linguistic production, is instead a plausible representation of how knowledge of grammar is stored in the mind, as well as an \textit{in silico} model of the brain language system \parencite{tuckute_language_2024}. The mechanisms of such a sequential processing can be efficiently represented through the short-term correlations captured by \ngram models as done in this work.

A deeper connection between \ourApproach and approaches to \AV based on language modeling or, as for \coav, on compression algorithms can also be drawn, thus offering a plausible explanation for their success. Through the lenses of \e{Information Theory}, the average $P(S; G)$ for each sentence $S$ in $\D$ can be also seen as an estimate of the {cross-entropy} rate of the Grammar Model $G$ against the probability distribution of grammatical tokens that generates $\D$. In language modeling literature, the cross-entropy is often exponentiated to calculate the \e{perplexity} of a document given a model. Because for \ourApproach the two alternative Grammar Models are compared in relation to the same $\D$, for the purposes of this \AV application, $\lambda_G$ is equivalent to comparing the perplexities of the candidate author's model vs. the reference model given the same $\D$ and assign the document to the candidate author if this perplexity is lower. The cross-entropy rate can also be interpreted as the average length in bits of a binary encoding of grammatical tokens for a data-compression scheme that is optimized for the probability distribution \(P(\cdot;G)\). For this reason, a language model can be seen as a text-compression scheme. This connection thus links compression and language model \AV methods like \ourApproach and \coav to the Cognitive Linguistics conception of an individual's grammar as a compressed representation of real language processed by said individual.

%---------------------------------------------- 
In addition to the successful results of \ourApproach, we note that the qualitative explorations reported in Section~\ref{sec-appendix-a} also constitute evidence that \ourApproach captures linguistic behavior compatible with these Cognitive Linguistic explanations. That individuals vary greatly in their mental grammar, as theorized by Langacker, can be seen at play in these text heat maps. The theoretical interpretation of these results is that a stronger shade of red effectively means a greater likelihood that the token is entrenched for that individual's unique grammar in contrast to the population. Although it is impossible to know for certain the effective mental status of the sequences found in the examples in Section~\ref{sec-appendix-a}, we propose that it is plausible to believe that the authors of these texts do not know that they use the sequences of tokens highlighted in red or that these sequences identify them in the general population. These sequences, which are often seemingly unremarkable (\eg \quotetxt{enough to be ADVERB VERB} or \quotetxt{so PRONOUN cant}[sic]), can be extremely idiosyncratic and uniquely used only by the candidate author. Given the fact that the problems analyzed in this paper involve relatively short texts (the longest texts are just around 2,000 tokens), we must conclude that these idiosyncratic units are easy to find and this conclusion therefore strongly supports Langacker's proposal that most units in a language are not shared by many individuals.

We believe that additional evidence towards this assertion is given by the results of the cross-reference experiments. Controlling for genre compatibility is probably the hardest problem in \AV and \AA. \ourApproach is very robust to this variation in many cases, much more than expected given the results of similar tests carried out on \imOrg \parencite{HalvaniPhD:2021}. This is despite the incredible challenges in some of the comparisons, such as in the analysis of the $\CorpusPJ$ chat logs using as reference the academic papers in $\CorpusACL$. The genre differences between these two corpora cannot be greater and therefore it is hard to believe that the method is not performing at chance level. A possible explanation for the success of \ourApproach in these circumstances is that the reference corpus effectively plays a much smaller role than initially thought. If individual grammars are so different from each other, then the role of the reference corpus could be to just weight down the those units that are well-entrenched for most individuals (\eg \quotetxt{of the}) and that therefore carry almost no authorial information. If this explanation is correct, then this means that once these common units are taken into account, most of the rest is idiosyncratic enough to be identifying.

Finally, due to the fact that these idiosyncratic units are easy to find within as little as 1,000 to 2,000 tokens, we add that a consequence of the present findings and of the underlying Cognitive Linguistic theory is that the grammar of a person can be seen as a behavioral biometric \parencite{NiniTheoryLingAA:2023}. Similarly to handwriting, gait, typing patterns, computer mouse usage, and most features of human voice, grammar is also a cognitive process relying on procedural memory that strengthens with habit and repetition. We propose that this is the explanation underlying most if not all \AV and \AA methods that, by measuring the frequency of functional items, are indirectly analyzing the idiosyncratic grammar of an individual. It is possible that \ourApproach is more successful because it is measuring this same phenomenon more directly.

\subsection{Limitations and Future Research}
We argue that future research should therefore continue in this direction of extracting better representations of the grammar of an individual. The \posNoise algorithm used here is language dependent and suffers from the use of general part of speech categories that are actually not fully compatible with Cognitive Linguistics, which instead predicts the existence of categories that are more semantic in nature and at lower degrees of abstraction. New future methods should embrace the close relationship between Cognitive Linguistics and Information Theory to gain more insight into grammar, which could consequently lead to further breakthroughs in \AV.

%---------------------------------------------- 
In addition, \ourApproach is still affected by genre differences and performs less well in registers where there is less variation, such as academic prose. These two last limitations affect more or less all existing algorithms which are outperformed by \ourApproach, which could mean that these limitations are intrinsic to \AV and, indeed, there would be good linguistic reasons for why this would be the case. Despite these limitations, future methods based on a better understanding of grammar could still lead to more progress. In this respect, the reliance on a reference corpus could also be seen as a limitation of any binary-extrinsic \AV method. The tendency for future \AV methods should be to increase the power of unary methods that consider only the documents within a verification case.

\ourApproach also suffers from two limitations that also apply to \imOrg because of their bootstrapping component: both algorithms are stochastic and require increased run time due to the number of repetitions. In our experiment we noted that \ourApproach is faster than \imOrg, although this depends on the specific implementation. Our test of the $\numberRepetitions$ hyperparameter also shows that 30 repetitions are enough for best performance, in contrast to the 100 iterations in \imOrg. Although the computational complexity of \ourApproach is still very limited compared to any method using neural networks and running on GPUs, such as \luarAV, a deterministic algorithm would definitely be preferable.

Finally, another limitation of the present paper is that the research was carried out entirely on English. The theory underlying \ourApproach generalizes to other languages and for this reason the prediction is that it would also outperform the same baseline methods in other languages. The way Grammar Models are trained in languages with a rich morphology is likely to be different, however. Different tokenizers might have to be employed as well as different algorithms such as \posNoise to focus the model on the grammatical aspects of the text. Our test showing that \posTags are not necessary is also encouraging for the replicability of this algorithm for languages with less computational resources. In general, though, another goal for the future would therefore be to find a language-independent variant of \ourApproach.

\subsection{Final Remarks}
The advances in \AV made in this paper were possible because of the collaboration between Linguistics and Natural Language Processing. In the 2022 edition of the PAN competition on \AV, \textcite{PANOverviewAV:2022} concluded that \quote{\e{all submissions, despite their increased level of sophistication in most of the cases, were outperformed by a naive baseline based on \charNgrams and cosine similarity}} thus noticing a situation of stall in the progress on \AV. Knowledge from Linguistics could resolve these stalls, as demonstrated in this paper, by shifting the point of view rather than adding more computational sophistication or making algorithms more complex. The method and philosophy introduced in the present work offer significant advancements in integrating Linguistics theory and Computer Science methods towards a new paradigm in \AA and \AV where these two fields are better harmonized. The task of identifying a writer from the language used in a text is really an application of Linguistics and there are various reasons for computational methods to maintain a close relationship with Linguistics. Firstly, these methods directly or indirectly exploit linguistic patterns that require knowledge of Linguistics to be described and understood. Secondly, knowledge of Linguistics is needed to understand and potentially explain in a court of law the reasons why these patterns identify or exclude a certain author. We believe that the work in this paper has made some significant advances in \AV while also proposing and stimulating a paradigm change to move on from the impasse described by \textcite{PANOverviewAV:2022} towards new \AV methods that are not only well-performing but also scientifically plausible.
\section{Methods} \label{sec-methods}
%--------------------------------------------------------------------
In this section, we present the more technical details of our work, first with an introduction to the corpora we have compiled to evaluate \ourApproach, then by providing more details about the \posNoise content-masking algorithm as well as the details of the Kneser-Ney smoothing algorithm, the likelihood ratio framework for forensic sciences.
%--------------------------------------------------------------------

\subsection{Corpora} \label{sec-data-sets}
%------------------------------------------------------------------------------------------------------ 
To evaluate \ourApproach, we compiled 12 corpora reflecting a range of challenges, including very short texts ($\approx5$ sentences), incoherent texts, texts with extensive slang, texts from the same authors written at different times, topic-opposed texts, post-edited texts with distorted style, texts with truncated sentences as well as texts with non-standard spellings or missing spaces. 
%------------------------------------------------------------------------------------------------------ 
In total, the corpora comprise 17,306 verification cases, were each corpus $\Corpus = \{ \Problem_1, \Problem_2, \ldots \}$ is split into author-disjunct training and test partitions based on a 40/60 ratio, resulting in 6,402 training and 10,904 test verification cases. Each $\Problem \in \Corpus$ denotes a verification case $(\Dunk, \Arefset)$, where $\Dunk$ is an unknown document and $\Arefset$ is a set of example documents of the known author $\A$. 
%------------------------------------------------------------------------------------------------------ 
To counteract \e{population homogeneity}, a known bias in \AV described by \textcite{BevendorffBiasAV:2019}, we ensured that for each author $\A$ there are exactly two verification cases \ie one with \textbf{same-authorship} (\classY) and one with \textbf{different-authorship} (\classN). Furthermore, we constructed all corpora so that the number of \classYdash{} and \classNdash{cases} is equal, in other words, all training and test corpora are (strictly) \textbf{balanced}. 
%------------------------------------------------------------------------------------------------------ 
In what follows, we describe all corpora in detail, underline their respective challenges and explain the respective preprocessing steps (using \e{TextUnitLib}, \cite{HalvaniTextUnitLib:2024}) we have performed for each corpus. A compact overview of the 12 corpora, including their partitioning into training and test corpora, as well as further information can be found in Table~\ref{tab:CorpusStatistics}. 
%-------------------------------------------------------
\definecolor{chargray}{RGB}{150,150,150}
\begin{table*} %[h!]
	\centering %\small 
	\begin{tabularx}{15.8cm}{llllrrr}
		\toprule		
		\bfseries\boldmath Corpus         & \textbf{Genre} & \textbf{Topic} & \textbf{Partition} & \boldmath$|\Corpus|$ &  \bfseries\boldmath avg$(\Arefset)$ & \bfseries\boldmath avg$(\Dunk)$ \\\midrule
        %--------------------------------------------------------------------------------------	
		\multirow{2}{*}{$\CorpusEnron$}   & \multirow{2}{*}{E-mails}   & \multirow{2}{*}{Related} & $\CorpusTrain$ & 64 & 783 \textcolor{chargray}{(3,866)} & 906 \textcolor{chargray}{(4,432)} \\
		&                                 &                                                              & $\CorpusTest$  & 96 & 763 \textcolor{chargray}{(3,808)} & 882 \textcolor{chargray}{(4,391)}  \\\midrule
        %--------------------------------------------------------------------------------------	
        \multirow{2}{*}{$\CorpusWiki$}    & \multirow{2}{*}{Wikipedia talk pages} & \multirow{2}{*}{Related} & $\CorpusTrain$ & 150 & 467 \textcolor{chargray}{(1,266)} & 614 \textcolor{chargray}{(1,654)} \\
		&                                 &                                                                         & $\CorpusTest$  & 226 & 492 \textcolor{chargray}{(1,328)} & 626 \textcolor{chargray}{(1,699)} \\\midrule
        %--------------------------------------------------------------------------------------	
  	\multirow{2}{*}{$\CorpusStack$} & \multirow{2}{*}{Q$\,$\&$\,$A posts} & \multirow{2}{*}{Cross} & $\CorpusTrain$ & 150 & 2,204 \textcolor{chargray}{(11,247)} & 1,972 \textcolor{chargray}{(9,956) } \\
		&                               &                                                                     & $\CorpusTest$  & 228 & 2,306 \textcolor{chargray}{(11,803)} & 2114 \textcolor{chargray}{(10,700)} \\\midrule
        %--------------------------------------------------------------------------------------	
        \multirow{2}{*}{$\CorpusACL$}     & \multirow{2}{*}{Scientific papers} & \multirow{2}{*}{Related} & $\CorpusTrain$ & 186 & 1,707 \textcolor{chargray}{(9,497) } & 2,553 \textcolor{chargray}{(14,196)} \\
		&                                 &                                                                      & $\CorpusTest$  & 280 & 1,479 \textcolor{chargray}{(8,250) } & 2,492 \textcolor{chargray}{(13,913)} \\\midrule
		%--------------------------------------------------------------------------------------
		\multirow{2}{*}{$\CorpusPJ$}      & \multirow{2}{*}{Chat logs}   & \multirow{2}{*}{Related} & $\CorpusTrain$ & 208 & 594 \textcolor{chargray}{(2,329)} & 752 \textcolor{chargray}{(2,946)} \\
		&                                 &                                                                & $\CorpusTest$  & 312 & 592 \textcolor{chargray}{(2,324)} & 754 \textcolor{chargray}{(2,967)} \\\midrule
        %--------------------------------------------------------------------------------------
		\multirow{2}{*}{$\CorpusApricity$}& \multirow{2}{*}{Forum posts}   & \multirow{2}{*}{Related} & $\CorpusTrain$ & 228 & 796 \textcolor{chargray}{(3,900)} & 825 \textcolor{chargray}{(4,023)} \\
		&                                 &                                                                  & $\CorpusTest$  & 340 & 799 \textcolor{chargray}{(3,921)} & 817 \textcolor{chargray}{(4,020)} \\\midrule
        %--------------------------------------------------------------------------------------
		\multirow{2}{*}{$\CorpusTripAdvisor$}  & \multirow{2}{*}{Travel reviews} & \multirow{2}{*}{Related} & $\CorpusTrain$ & 120 & 658 \textcolor{chargray}{(3,169)} & 249 \textcolor{chargray}{(1,186)} \\
		&                                 &                                                                        & $\CorpusTest$  & 480 & 698 \textcolor{chargray}{(3,344)} & 252 \textcolor{chargray}{(1,202)} \\\midrule
		%--------------------------------------------------------------------------------------
         \multirow{2}{*}{$\CorpusYelp$}   & \multirow{2}{*}{Hotel/restaurant reviews} & \multirow{2}{*}{Related} & $\CorpusTrain$ & 320  & 139 \textcolor{chargray}{(637) } & 167 \textcolor{chargray}{(767) } \\
		&                                 &                                                                             & $\CorpusTest$  & 480  & 139 \textcolor{chargray}{(640) } & 167 \textcolor{chargray}{(768) } \\\midrule
		%--------------------------------------------------------------------------------------
  	\multirow{2}{*}{$\CorpusIMDB$}    & \multirow{2}{*}{Movie reviews} & \multirow{2}{*}{Mixed} & $\CorpusTrain$ & 400 & 658 \textcolor{chargray}{(3,151)} & 374 \textcolor{chargray}{(1,796)}  \\
		&                                 &                                                                & $\CorpusTest$ & 1,600 & 690 \textcolor{chargray}{(3,291)} & 366 \textcolor{chargray}{(1,753)}  \\\midrule
        %--------------------------------------------------------------------------------------
		\multirow{2}{*}{$\CorpusBlogs$}   & \multirow{2}{*}{Blogs posts} & \multirow{2}{*}{Mixed} & $\CorpusTrain$ & 1,200 & 742 \textcolor{chargray}{(3,416)} & 881 \textcolor{chargray}{(4,086)} \\
		&                                 &                                                              & $\CorpusTest$  & 1,800 & 740 \textcolor{chargray}{(3,385)} & 882 \textcolor{chargray}{(4,061)}  \\\midrule
        %--------------------------------------------------------------------------------------
		\multirow{2}{*}{$\CorpusAmazon$}  & \multirow{2}{*}{Product reviews} & \multirow{2}{*}{Mixed} & $\CorpusTrain$ & 1,600 & 861 \textcolor{chargray}{(4,010)} & 869 \textcolor{chargray}{(4,041)} \\
		&                                 &                                                                  & $\CorpusTest$  & 2,400  & 860 \textcolor{chargray}{(4,009)} & 868 \textcolor{chargray}{(4,041)} \\\midrule
        %--------------------------------------------------------------------------------------
		\multirow{2}{*}{$\CorpusAllTheNews$}  & \multirow{2}{*}{News articles} & \multirow{2}{*}{Mixed} & $\CorpusTrain$ & 1,776 & 675 \textcolor{chargray}{(3,629)} & 941 \textcolor{chargray}{(5,064)} \\
		&                                 &                                                                    & $\CorpusTest$  & 2,662  & 680 \textcolor{chargray}{(3,665)} & 942 \textcolor{chargray}{(5,071)} \\\bottomrule  
		%--------------------------------------------------------------------------------------
	\end{tabularx}	
	\caption{All corpora used and their key statistics. Notation: $|\Corpus|$ denotes the number of verification cases in each corpus $\Corpus$, while avg$(\Arefset)$ and avg$(\Dunk)$ denotes the average token count (and character count in parentheses) of all $\DA \in \Arefset$ and $\Dunk$, respectively. \label{tab:CorpusStatistics}}
\end{table*}

\subsubsection{\boldmath$\CorpusEnron$}
The $\CorpusEnron$ corpus was derived from the well-known \e{Enron Email Dataset} \parencite{TheEnronCorpus:2004}, which has been used in a variety of \AV and \AA studies, including \parencite{BrocardoStylometryAV:2013,BrocardoContinuousAuthenticationAV:2015,BrocardoDeepBeliefAV:2017,ChenAuthorshipSimilarityDetection:2011,DingVisualizableEvidenceApproachAA:2015,LitvakAVwithCNNs:2019,NovinoSingleClassAA:2015,WrightAAviaWordNgrams:2017,ApoorvaSangeethaDeepAA:2021}, as well as in many other research fields. The \e{Enron email dataset} has gained considerable popularity in the authorship analysis community, likely due to the fact that it is (to the best of our knowledge) the only publicly available corpus of \textbf{real-world emails} with a guaranteed ground truth. Similar to previous \AV studies, we also reduced the initial set to a smaller number of authors (in our case from 150 to 80), as only a few suitable emails were available for some of the authors. In total, $\CorpusEnron$ consists of 362 documents, with each document representing a concatenation of several emails from the corresponding author in order to obtain a sufficient length. Since the emails in the \e{Enron Email Dataset} contain a wide range of noise, extensive preprocessing was necessary. To this end, we decided to manually preprocess all considered texts and instead create individual replacement rules for various types of noise. First, we removed URLs, email headers, greeting/closing formulas, email signatures, (phone) numbers, quotes as well as various non-letter repetitions. Next, we normalized UTF-8 symbols and, in a final step, replaced multiple consecutive spaces, line breaks and tabs with a single space. A similar preprocessing procedure was also described by \textcite{BrocardoStylometryAV:2013,BrocardoContinuousAuthenticationAV:2015}.

\subsubsection{\boldmath$\CorpusWiki$}
The $\CorpusWiki$ corpus consists of 752 excerpts from 288 Wikipedia talk page editors taken from the Wikipedia sockpuppets dataset, released by \textcite{SolorioSockpuppet:2014}. The original dataset contains two partitions comprising sockpuppets and non-sockpuppets cases, where for $\CorpusWiki$ we only considered the latter subset. In addition, we did not use the full range of authors within the non-sockpuppets cases, as there were not sufficiently long texts available for each of the authors. Regarding the selected texts, we removed Wiki markup, timestamps, URLs and other types of noise. Furthermore, we removed sentences containing many numbers, proper names and near-duplicate string fragments as well as truncated sentences. 

\subsubsection{\boldmath$\CorpusStack$}
The $\CorpusStack$ corpus consists of 567 posts from 189 users crawled from the question-and-answer (Q$\,$\&$\,$A) network \e{Stack Exchange} \parencite{StackExchangeDataset:2019} in 2019. The network comprises 173 Q$\,$\&$\,$A communities, with each community focusing on a specific topic (\eg linguistics, computer science, philosophy, etc.). For the compilation of $\CorpusStack$, which represents a \textbf{cross-topic} corpus, we collected questions and answers from users who were simultaneously active in the two thematically different communities \e{Cross Validated} ($\mathcal{S}_1$) and \e{Academia} ($\mathcal{S}_2$). The verification cases were constructed as follows. Each verification case consists of exactly two documents, a known and an unknown document $\DA$ and $\Dunk$. Within the \classYdash{cases}, $\DA$ and $\Dunk$ were taken from $\mathcal{S}_1$ and $\mathcal{S}_2$ respectively, while within the \classNdash{cases} $\DA$ and $\Dunk$ were taken from the same community $\mathcal{S}_1$. Due to the cross-topic nature of this corpus, \AV methods that make use of topic-affected features are more likely to provide inverse predictions (\ie \classNdash{cases} are more likely to be classified as \classY and vice versa) while \AV methods that use topic-agnostic features behave more robustly. Another challenge of $\CorpusStack$ is that the corpus \textbf{is highly unbalanced in terms of document length}, with lengths between 3,014 and 20,476 characters. To avoid stylistic distortions from the outset, we only used the original posts of the respective authors for all texts in $\CorpusStack$, while subsequent edits by third parties (for the purpose of spelling or stylistic improvement) were excluded (\e{Stack Exchange} provides metadata to show exactly which parts of the text have been edited, and also allows the original unedited post to be retrieved). In addition, we replaced equations, numerical values (\eg percentages and currency amounts), dates, references and URLs with appropriate placeholders. Rather than discarding the affected sentences, we applied this approach to preserve as much text per author as possible and thereby ensure a sufficient number of authors.

\subsubsection{\boldmath$\CorpusACL$}
The $\CorpusACL$ corpus comprises 466 excerpts from scientific papers by a total of 233 researchers, taken from the computational linguistics archive \e{ACL Anthology}\footnote{\url{https://aclanthology.org}}. The corpus was constructed in such a way that there are exactly two papers for each author\footnote{Using the metadata of the papers (more precisely, the author fields in the corresponding BibTex entries), we ensured that each paper was written by exactly one author.}, dating from different time periods. The average time span between the two documents of each author is approximately 15.6 years, while the minimum and maximum time spans are 8 and 31 years, respectively. In addition to the temporal aspect of the documents in $\CorpusACL$, another challenge of this corpus is the \textbf{formal language}, in which the use of stylistic devices (\eg repetitions, metaphors or rhetorical questions) is more restricted in contrast to other text types such as chat logs, forum posts or product reviews. For the original papers, we have tried to limit the content of each text as much as possible to those sections that consist mainly of natural language text (\eg \e{abstract}, \e{introduction}, \e{discussion}, \e{conclusion} or \e{future work}). To ensure that the extracted fragments are consistent with the remarks of \textcite{BevendorffBiasAV:2019}, we decided to preprocess each paper excerpt in $\CorpusACL$ (semi-)manually. Among other things, we removed tables, quotations and sentences that contained a large amount of non-linguistic content such as mathematical constructs as well as specific names of researchers, systems and algorithms. We also replaced (inline) mathematical expressions, references (\eg~\quotetxt{[12]}, \quotetxt{[SHL13]}), endophoric markers, numeric values and URLs with appropriate placeholders, as we did with the texts in $\CorpusStack$.

\subsubsection{\boldmath$\CorpusPJ$} 
The $\CorpusPJ$ corpus comprises a total of 738 chat logs from 260 sex offenders collected from the \e{Perverted-Justice}\footnote{\url{http://www.perverted-justice.com}} portal. The chat logs were obtained from older instant messaging clients (\eg MSN, AOL and Yahoo), whereby for each conversation we ensured that only the offender's chat lines were extracted based on the annotations provided by the portal. To maximize variability in terms of conversation content, we selected chat lines from different messaging clients as well as from different time spans (where possible). A major challenge of $\CorpusPJ$ is that it contains texts characterized by \textbf{excessive use of non-standard English} and also contains many chat-specific abbreviations and other forms of noise. This in turn makes it difficult to extract syntax-based features such as \posTags from the texts, at least when using NLP models trained on texts written in standard English (\eg news articles or scientific documents). Another challenge of $\CorpusPJ$ is that the text fragments it contains are \textbf{very short}, which in turn limits stylistic variability. In contrast to $\CorpusEnron$, we have automatically preprocessed the texts in this corpus. As part of the preprocessing, we excluded chat lines containing fewer than five tokens or ten characters, along with those including usernames, timestamps, URLs or words exhibiting excessive character repetitions (\eg~\quotetxt{mmmhhhmm}, \quotetxt{heeheehee} and \quotetxt{:-*:-*:-*:-*:-*:-*:-*:-*}). To further increase the variability in the content of each chat log $X$, we made sure that all contained lines $(x_1, x_2, \ldots, x_m)$ are different from each other. For this purpose, we applied a string similarity function to each pair $(x_i, x_j)$ with $i \neq j$. As a metric for string similarity, we chose the \e{Levenshtein distance} (using \e{TheFuzz} library \parencite{TheFuzz:2021}) based on a threshold of 0.25, which we considered appropriate for our purposes. All pairs of lines that exceeded this threshold were discarded with respect to the text fragments. 

\subsubsection{\boldmath$\CorpusApricity$} 
The $\CorpusApricity$ corpus comprises 1,395 forum posts from 284 users crawled from the portal \e{The Apricity - A European Cultural Community} \parencite{TheApricityDataset:2018}. To compile $\CorpusApricity$, we ensured that all posts within each verification case were collected from different sub-forums covering a variety of topics, including \e{anthropology}, \e{genetics}, \e{race and society} and \e{ethno-cultural discussion}. We then preprocessed each post and removed certain markup tags, URLs, forum signatures, quotations (as well as nested quotations that contained the original author's content) and other types of noise, similar to what we did with the texts within the $\CorpusEnron$ corpus. However, despite the preprocessing, some \textbf{textual peculiarities} remained in the texts, such as truncated sentences, spelling errors and missing spaces between adjacent words or between a punctuation mark and the following word. These can be seen as a challenge of $\CorpusApricity$, especially with regard to \AV methods that use features based on word lists.

\subsubsection{\boldmath$\CorpusTripAdvisor$} 
The $\CorpusTripAdvisor$ corpus consists of reviews from 300 bloggers, which were extracted from the \e{Webis TripAdvisor Corpus 2014}, compiled by \textcite{TrenkmannSpielWebisTripad14:2015}. Given that the original corpus was not intended for \AV purposes, we reformatted the data accordingly. While inspecting the data, we found that most of the included reviews were already of reasonable quality, so we only carried out moderate preprocessing. Among other things, we adjusted spaces next to punctuation marks (\eg~\quotetxt{( alight} $\rightarrow$ \quotetxt{(alight}) and normalized many character repetitions to a maximum of three characters (\eg~\quotetxt{(!!!!!} $\rightarrow$ \quotetxt{!!!}). Besides, we excluded reviews that contained too many named entities and we discarded authors for which only one review was available.

\subsubsection{\boldmath$\CorpusYelp$} 
The $\CorpusYelp$ corpus comprises 2,400 reviews written by 400 so-called \quote{Yelplers} extracted from the well-known \e{Yelp dataset} \parencite{YelpOpenDataset:2018}. The corpus was developed with several challenges in mind. First, compared to all other corpora we have compiled, it contains the \textbf{shortest texts}, with each $\Dunk$ consisting of at most ten sentences. For this reason, the stylistic variability within these texts is correspondingly limited. Secondly, many reviews were written at \textbf{different times} (in several cases often about four years apart), so linguistic changes are to be expected in reviews written by the same person. Thirdly, most of the texts in the $\CorpusYelp$ are restaurant reviews, which focus heavily on similar topics such as food, atmosphere and taste experiences. To a lesser extent, however, other forms of reviews are also included in the corpus (\eg~about wedding or clothing stores), which are accompanied by \textbf{linguistic and also stylistic diversity} with respect to the authors. Fourth, some of the reviews contain a mixture of moderate and colloquial writing styles. Regarding the latter, we observed certain spelling variations in the texts (\eg missing apostrophes within contractions), which may result in \textbf{noticeable loss of features} for \AV methods that rely on word lists. For the review texts in $\CorpusYelp$, we performed the following preprocessing steps. Sentences containing capitalized words throughout were removed (\eg~\quotetxt{I WAITED 4 HOURS IN THE BEND AND COULDN'T GET IN!!!!!!!!}), as were sentences containing quotes and many repetitive characters (\eg~\quotetxt{wait for a heart attack!!!!! Incompetent staff!!!! will never come here again!!!!}). Moreover, we removed sentences that only consist of single words (\eg~\quotetxt{fraud. Fraud. Lies.}) as well as review-type summaries such as \quotetxt{Quality: 10/10; Service: 9/10;} \quotetxt{Ambience/Location: 10/10; Overall: 9.5/10}.

\subsubsection{\boldmath$\CorpusIMDB$} 
The $\CorpusIMDB$ corpus consists of reviews from a total of 1,000 IMDB users taken from the \e{IMDB-1M Dataset} published by \textcite{SeroussiAATopicModels:2014}. From the original dataset, only the review texts and corresponding author names (more precisely, user IDs) were selected, while other metadata were discarded. Analogous to $\CorpusTripAdvisor$, here we also performed a slight preprocessing since the quality of the texts was already suitable for the \AV task. Among others, we normalized spaces adjacent to punctuation marks (\eg~\quotetxt{but . . .	I} $\rightarrow$ \quotetxt{but... I}) and replaced year numbers with a placeholder (\eg~\quotetxt{1982} $\rightarrow$ \quotetxt{YEAR}). Furthermore, we also excluded authors for whom only one review was available as we did for $\CorpusTripAdvisor$. 

\subsubsection{\boldmath$\CorpusBlogs$} 
The $\CorpusBlogs$ corpus comprises 6,000 documents written by 1,500 bloggers, taken from the well-known \e{Blog Authorship Corpus}, released by \textcite{SchlerKoppelAPGenderAge:2006}. The original dataset comprises 681,288 posts of 19,320 bloggers and was collected from \emph{blogger.com} in 2004. However, a large proportion of the posts were not suitable for our purposes for various reasons. For example, many posts consisted of non-English content or contained too much noise (\eg superfluous strings, duplicate phrases, quotes, etc.) that was difficult to preprocess without losing too much text for the respective authors. In addition, the amount of text was insufficient in itself for many authors. Therefore, we only considered a smaller fraction of the original corpus. To counteract various noise types, we first sampled sentences from the original posts and ensured that each sentence is solely restricted to (case-sensitive) English letters, spaces and standard punctuation marks. From the resulting texts we then removed sentences containing repeated characters (\eg~\quotetxt{?!??!}, \quotetxt{---} and \quotetxt{-.-;;}) and words (\eg~\quotetxt{FUN!! FUN!! FUN!!} and \quotetxt{kekeke}), as well as sentences with many capitalized words (\eg~\quotetxt{i am saying that WHOEVER HAS OPTUS PLEASE TELL ME}). Besides, we discarded sentences containing URLs, gibberish strings (\eg~\quotetxt{rak;db'dfth6yortyi95}, \quotetxt{aDr0} and \quotetxt{w00t}) as well as truncated sentences. Finally, we concatenated the preprocessed sentences into a single document, which in turn led to a \textbf{disruption of the coherence} of the resulting texts. Similar to $\CorpusPJ$, the texts in $\CorpusBlogs$ also exhibit an excessive amount of \textbf{non-standard English expressions} such as \quotetxt{w8 4} (\e{wait for}), \quotetxt{b4} (\e{before}), \quotetxt{2day} (\e{today}), \quotetxt{sum1's} (\e{someone's}) and \quotetxt{jnr} (\e{junior}). This can be seen as another challenge of $\CorpusBlogs$, since the spelling of such words is not always consistently adhered to with respect to the authors.

\subsubsection{\boldmath$\CorpusAmazon$} 
The $\CorpusAmazon$ corpus consists of 10,000 reviews written by 2,000 users, which were taken from the \emph{Amazon Product Data} corpus, published by \textcite{AmazonReviewCorpus:2015}. The original dataset contains approximately 143 million product reviews of the online marketplace \e{Amazon} collected between 1996 and 2014. While this dataset possesses a comprehensive structure that includes reviews (\e{ratings}, \e{text}, \e{helpfulness votes}), product metadata (\e{descriptions}, \e{category information}, \e{price}, \e{brand}, and \e{image features}) and links (\e{also viewed}/\e{bought graphs}), we only used the reviewer IDs, their corresponding texts as well as the category information to compile our $\CorpusAmazon$ corpus. Based on the latter, we selected for each author review texts from distinct product categories (in total there are 17 categories including \e{electronics}, \e{movies and TV} and \e{office products}), so that $\CorpusAmazon$ represents a \textbf{mixed-topic} corpus. Regarding the texts that came from the \emph{Amazon Product Data} corpus, we performed several preprocessing steps before selecting them for $\CorpusAmazon$. Among others, we normalized HTML-encoded punctuation marks with respect to their mapped characters (for example, \texttt{\&\#34;} $\rightarrow$ \texttt{"} or \texttt{\&\#8217;} $\rightarrow$ \texttt{'}). In addition, we have added spaces between punctuation marks and adjacent words (\eg~\quotetxt{\quotetxt{daily use\textbf{.}I also}} $\rightarrow$ \quotetxt{daily use I also}) to allow for better separation of sentences. Furthermore, we removed sentences containing various forms of noise such as words with repetitive characters (\eg~\quotetxt{****update*****} or \quotetxt{mix.....it}), many capitalized words \quotetxt{VERY NICELY MADE AND THE COLOR IS BEAUTIFUL}, non-natural strings (\eg~\quotetxt{A++++}, \quotetxt{(48x18x21)} or \quotetxt{(27 \& 23 Lbs)}) as well as quotes. 

\subsubsection{\boldmath$\CorpusAllTheNews$} 
The $\CorpusAllTheNews$ corpus consists of 4,438 news articles written by 2,219 authors. The articles stem from the large well-known portal \e{All the News 2.0} \parencite{AllTheNewsDataset:2017}, which consists of 2.7 million news articles and essays from 27 American news agencies and podcasting companies. From these, we limit ourselves to a single source, namely Reuters, which contains the largest number of news articles (840,094 in total). However, since not all articles were suitable for the creation of $\CorpusAllTheNews$, it was necessary to preprocess the initial Reuters subset accordingly. First, we discarded all authors who wrote only a single article, as otherwise there is no suitable possibility to construct two verification cases for each author. Next, we sorted all articles of each author by their publication year and kept only four articles (the two oldest and the two newest). The main reasons for this were, firstly, to maximize the stylistic diversity of each author and, secondly, to ensure a sufficient length of the resulting documents, which in some cases were heavily shortened due to the preprocessing. Based on the four documents per author, the preprocessing was carried out as follows. As a first step, we removed all quotations (which are particularly common in news articles) to mitigate stylistic distortion. Next, we removed headlines, sentences that consisted of more than 40 percent named entities and/or numeric tokens (\eg dates, amounts of money, percentages, etc.) as well as article signatures that contained names of people who reported, wrote and edited the articles. Finally, we normalized apostrophes and replaced multiple consecutive spaces with a single space. The preprocessing procedure severely impaired the coherence of the resulting documents, as the natural order of many sentences was no longer given and some sentences also resulted in a truncated form. The \textbf{impaired coherence} can be seen as a challenge of this corpus, as \AV methods that use implicit features (\eg character/word \ngrams) may capture \quote{artificial features} that occur across sentence boundaries. Another challenge of $\CorpusAllTheNews$ is that the documents often have been \textbf{edited by other authors} (\eg the editor-in-chief) before online publication, so the original writing style of the articles may be distorted to some extent. Unlike the \e{Stack Exchange} network, \e{All the News 2.0} does not provide metadata indicating which specific parts of the original article have been edited. Therefore, it was not possible for us to remove these retrospectively.

\subsection{Topic-Masking via \posNoiseBold} \label{sec-the-posnoise-algorithm}
\begin{figure}[h!]
    \centering
    \includegraphics[width=1\linewidth]{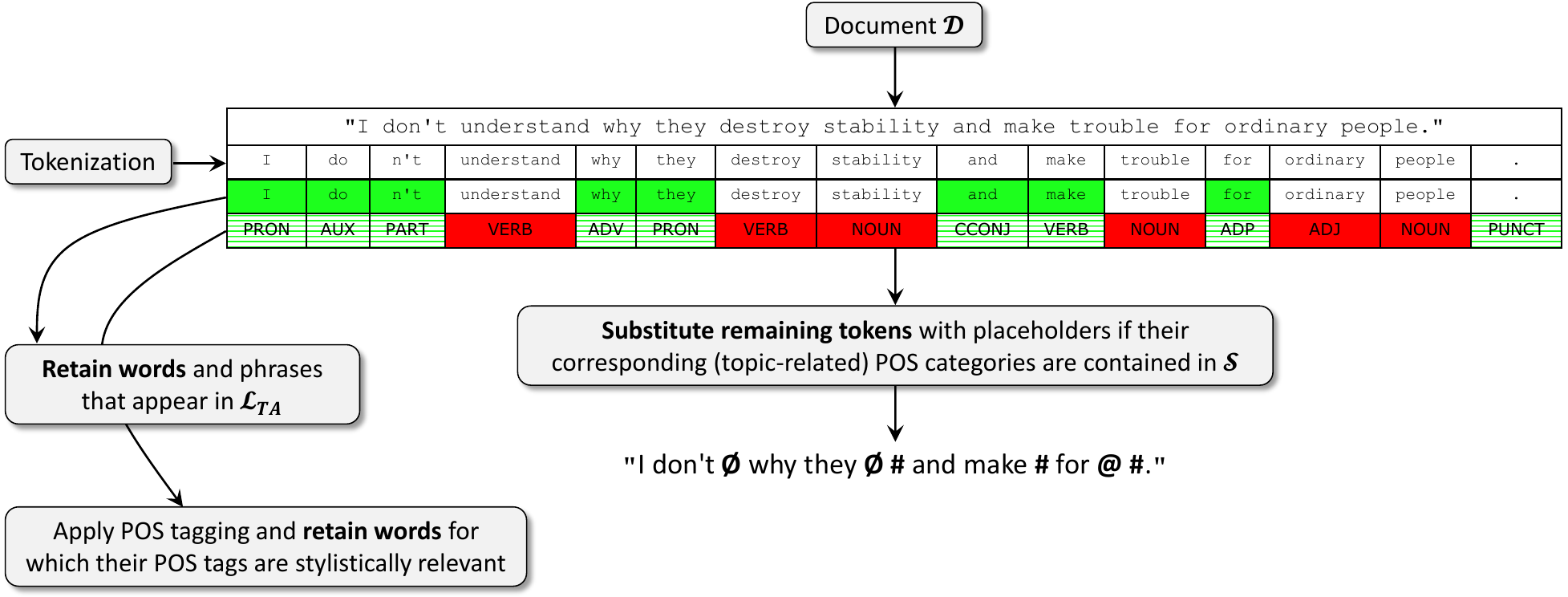}
    \caption{Illustration of the \posNoise algorithm. The notation is explained in detail by \textcite{HalvaniPOSNoise:2021}.}
    \label{fig:posnoise}
\end{figure}

\AV constitutes a similarity detection problem in which the subject of the similarity determination is the language of the author rather than other document aspects such as the topic \parencite{HalvaniPOSNoise:2021}. However, a large number of existing \AV methods including  \parencite{AgbeyangiAVYorubaBlogPostsCharNgrams:2020,BrocardoStylometryAV:2013,BrocardoDeepBeliefAV:2017,CastroAVAverageSimilarity:2015,KoppelSeidmanEMNLP:2013,KoppelWinter2DocsBy1:2014,LitvakAVwithCNNs:2019,NealAVviaIsolationForests:2018,PothaStamatatosExtrinsicAV:2019} use features that are directly influenced by the topic and thus run the risk of biasing their predictions accordingly. To counteract the problem of biased predictions regarding \ourApproach, which also employs topic-affected features, we apply topic masking to all involved documents (see Algorithm~\ref{AlgoGMComp}). Topic masking is an approach to preprocess texts in such a way that topic-dependent text units such as words or phrases are masked by appropriate placeholders, while grammatical text units (\eg determiners, conjunctions, contractions, prepositions, pronouns, auxiliary verbs, quantifiers, transitional phrases, etc.) are retained within the texts. Of the few topic masking approaches published in the literature, we chose \posNoise\footnote{An implementation of \posNoise can be found at \url{https://github.com/Halvani/POSNoise}} presented by \textcite{HalvaniPOSNoise:2021}, which proved to outperform the previously published topic masking approach called \textDistortion \parencite{StamatatosTextDistortion:2017} in 34 out of 42 cases, with an increase in Accuracy of up to 10\%. 

\begin{table} 
	\centering
    \small
	\begin{tabular}{ll} 
		\toprule 		%============================================================================================================
		\textbf{Category} & \textbf{Examples} \\ \midrule
		%============================================================================================================
		Conjunctions         & $\{\texttt{and, as, but, either, for, hence, however, if, neither, nor, once,}\,\e{...}\,\}$  \\
		%------------------------------------------------------------------------------------------------------------ 		
		Determiners          & $\{\texttt{a, an, both, each, either, every, no, other, our, some,}\,\e{...}\,\}$  \\
		%------------------------------------------------------------------------------------------------------------ 		
		Prepositions         & $\{\texttt{above, across, after, among, below, beside, between, beyond, inside,}\,\e{...}\,\}$  \\
		%------------------------------------------------------------------------------------------------------------		
		Pronouns             & $\{\texttt{all, another, any, anyone, anything, everything, few, he, her, hers,}\,\e{...}\,\}$  \\
		%------------------------------------------------------------------------------------------------------------		
		Quantifiers          & $\{\texttt{any, certain, each, either, few, less, lots, many, more, most, much,}\,\e{...}\,\}$  \\	\midrule	
		%============================================================================================================
		Auxiliary verbs      & $\{\texttt{can, could, might, must, ought, shall, will,}\,\e{...}\,\}$  \\
		%------------------------------------------------------------------------------------------------------------ 
		Delexicalised verbs  & $\{\texttt{get, go, take, make, do, have, give, set,}\,\e{...}\,\}$  \\
		%------------------------------------------------------------------------------------------------------------ 
		Empty verbs          & $\{\texttt{do, did, does, got, getting, have, had, had, gives, gave, give, gets,}\,\e{...}\,\}$  \\
		%------------------------------------------------------------------------------------------------------------ 
		Helping verbs        & $\{\texttt{am, is, are, was, were, be, been, being, will, should, would, could,}\,\e{...}\,\}$  \\ \midrule
		%============================================================================================================ 
		Contractions         & $\{\texttt{i'm, i'd, i'll, i've, he's, it's, we'd, she's, it'll, we're,  you're,}\,\e{...}\,\}$  \\ \midrule
		%============================================================================================================	
		Adverbs of degree    & $\{\texttt{almost, enough, hardly, just, nearly, quite, simply, so, too,}\,\e{...}\,\}$  \\	
		Adverbs of frequency & $\{\texttt{again, always, never, normally, rarely, seldom, sometimes, usually,}\,\e{...}\,\}$  \\		
		Adverbs of place     & $\{\texttt{above, below, everywhere, here, in, into, nowhere, out, there,}\,\e{...}\,\}$  \\
		Adverbs of time      & $\{\texttt{already, during, now, late, recently, still, then, sometimes, yet,}\,\e{...}\,\}$  \\		
		Pronominal adverbs   & $\{\texttt{hereafter, hereby, thereafter, thereby, therefore, therein, whereas,}\,\e{...}\,\}$  \\
		Focusing adverbs     & $\{\texttt{especially, mainly, generally, only, simply, exactly, merely, solely,}\,\e{...}\,\}$  \\
		Conjunctive adverbs  & $\{\texttt{likewise, meanwhile, moreover, namely, nonetheless, otherwise,}\,\e{...}\,\}$  \\ \midrule
		%============================================================================================================
		Transition words     & $\{\texttt{besides, furthermore, generally, hence, thus, however, subsequently,}\,\e{...}\,\}$  \\
		Transitional phrases & $\{\texttt{of course, as a result, because of, in contrast, on the other hand,}\,\e{...}\,\}$  \\
		Phrasal prepositions & $\{\texttt{as opposed to, in regard to, inspite of, out of, with regard to,}\,\e{...}\,\}$  \\
		%------------------------------------------------------------------------------------------------------------				
		\bottomrule		
	\end{tabular}
	\caption{All categories of topic agnostic words and phrases whitelisted by \posNoise with some examples.\label{table:POSNoiseFeatures}}
\end{table}

\begin{table} [h!]
	\centering % \small 
	\begin{tabularx}{16cm}{ll} \toprule  
		\textbf{Corpus} &     \textbf{Original / topic-masked sentences} \\ \midrule
  %----------------------------------------------------------------------------------------------------------
		\multirow{2}{*}{$\CorpusWiki$}  & \texttt{If they actually censor anything is another question.} \\
		                                & \texttt{If they © Ø anything is another \#.} \\\midrule	
        %----------------------------------------------------------------------------------------------------------  
        \multirow{2}{*}{$\CorpusStack$} & \texttt{When I took the course, I ended up not being able to learn from the professor} \\		
		                                  & \texttt{When I took the \#, I Ø up not being @ to Ø from the \#} \\\midrule	
		%----------------------------------------------------------------------------------------------------------		
		\multirow{2}{*}{$\CorpusPJ$}    & \texttt{going to wash clothes ,call me,tim} \\
                                        & \texttt{going to Ø \# ,call me,§} \\\midrule	
		%----------------------------------------------------------------------------------------------------------		
		\multirow{2}{*}{$\CorpusYelp$}  & \texttt{Our meal started with a basket of freshly baked baguette and focaccia.} \\
		                                  & \texttt{Our \# started with a \# of © Ø \# and \#.} \\\midrule	
	    %----------------------------------------------------------------------------------------------------------
		\multirow{2}{*}{$\CorpusBlogs$} & \texttt{Then, Anna, Peter and I all went to baker's Square.} \\		
		                                & \texttt{Then, §, § and I all went to \#'s §.} \\\midrule                                    
		%---------------------------------------------------------------------------------------------------------- 
	    \multirow{2}{*}{$\CorpusAmazon$}& \texttt{I got a new mascara that was impervious to makeup remover } \\
		                                  & \texttt{I got a @ \# that was @ to \# \#} \\	
		%----------------------------------------------------------------------------------------------------------
		\bottomrule	
	\end{tabularx}
	\caption{Comparison between the original texts and their topic-masked representations generated by \posNoise \parencite[Algorithm:~1]{HalvaniPOSNoise:2021}. 
    The semantics of the placeholders \texttt{©}, \texttt{Ø}, \texttt{@}, \texttt{\#} and \texttt{§} are provided in \parencite[Table:~2]{HalvaniPOSNoise:2021}. \label{table:ComparisonTopicMasking}}
\end{table} 

\subsection{N-Gram Language Modeling with Kneser-Ney Smoothing}\label{sec-ngram-modelling}
This section provides a self-contained introduction to (modified) Kneser-Ney smoothing \parencite{kneser1995, ChenGoodmanLMSmoothing:1996}. Our main purpose is to provide the reader with all necessary details required to reproduce our \AV method. For a more in-depth discussion of probabilistic and language modeling aspects, as well as for comparisons with other smoothing algorithms, we refer the reader to the original references \parencite{kneser1995, ChenGoodmanLMSmoothing:1996}.

\subsubsection{Basic Definitions and Notation}
We assume we are given a dictionary $\vocabulary$ and a corpus $\mathcal {C}$, the latter consisting of (pre-tokenized) sentences built out of tokens from $\vocabulary$. Sequences of tokens of length $n$, or \ngrams, will be indicated by juxtaposition, such as in $\token_1 \token_2 \cdots \token_n$, and their observed frequencies will be denoted by:
\begin{equation}\label{eq:ngramCounts}
c(\token_1 \token_2\cdots \token_n) = \text{Number of occurrences of $\token_1 \token_2 \dots \token_n$ in $\mathcal C$.} 
\end{equation}
In discussions of \ngram models, one usually encounters three special tokens: $\langle \text{UNK}\rangle$, $\langle \text{BOS}\rangle$ and $\langle \text{EOS}\rangle$ - the "UNKnown token", "Begin Of Sentence" and "End Of Sentence" tokens. In order to avoid notational ambiguities we make here the convention that $\langle \text{UNK}\rangle \in \vocabulary$, but $\langle \text{BOS}\rangle,\,\langle \text{EOS}\rangle\notin \vocabulary$. In particular, sequences in $\mathcal C$ can feature the $\langle \text{UNK}\rangle$ token, but not the $\langle \text{BOS}\rangle$ or $\langle \text{EOS}\rangle$ tokens, which only play the role of book-keeping devices for \ngram functions such as \eqref{eq:ngramCounts}, as explained below.

The definition \eqref{eq:ngramCounts} is also valid for \ngrams featuring the $\langle \text{UNK}\rangle$ token, with the understanding that out-of-vocabulary tokens in $\mathcal C$ should be replaced by $\langle \text{UNK}\rangle$ before counting. The case of \ngrams featuring the pseudo-tokens $\langle \text{BOS}\rangle$ and $\langle \text{EOS}\rangle$ is treated as follows: when counting the \ngrams in $\mathcal C$, we pad all sentences with exactly $n$ $\langle \text{BOS}\rangle$ tokens on the left, and a single $\langle \text{EOS}\rangle$ token on the right. Our definition implies, in particular, that: $$c(\langle \text{BOS}\rangle \langle \text{BOS}\rangle \cdots \langle \text{BOS}\rangle ) = c(\langle \text{EOS}\rangle) = \text{Number of sentences in }\mathcal{C}$$
for any number of $\langle \text{BOS}\rangle$ tokens in the left hand side of this equation. Furthermore,
\[ 
   c(\langle \text{BOS}\rangle \langle \text{BOS}\rangle  \cdots \langle \text{BOS}\rangle \token_1 \token_2 \cdots \token_n ) \text{ and } c(\token_1 \token_2 \cdots \token_n \langle \text{EOS}\rangle)
\]
are given by the number of sentences that started and ended with the string $\token_1 \token_2 \dots \token_n$, respectively. Finally, for notational convenience, we define the left trimming operator $L$ that maps \ngrams into $(n-1)$-grams:
\begin{align}\label{eq:LeftTrimOp}
    L(\token_{1} \token_{2}\cdots \token_{n}) & =\token_{2}\token_{3}\cdots \token_{n},\\
    L(\token) & = \emptyset
\end{align}
Here, $\emptyset$ denotes the empty $0$-gram.

\subsubsection{Prefix and Suffix Counts}
Kneser-Ney smoothing relies on a few other \ngram functions besides the basic counts $c(\cdot)$ introduced in the previous subsection. Specifically, given an \ngram $g$ and a positive integer $\numberRepetitions = 1,\,2,\,\dots$, we define:
\begin{align} 
\label{eq:prefixCount}
N_{r}(\bullet g)        &= \left|\left\{ \token \in \vocabulary \cup \{\langle \text{BOS}\rangle\}\colon\,c(\token g)= \numberRepetitions\right\} \right|, \\
\label{eq:suffixCount}
N_{r}(g \bullet)            &= \left|\left\{ \token \in \vocabulary \cup \{\langle \text{EOS}\rangle\}\colon\,c(g \token)= \numberRepetitions\right\} \right|,
\end{align}
and, correspondingly:
\begin{align} 
\label{eq:prefixCountPlus}
N_{r+}(\bullet g)       &= \left|\left\{ \token \in \vocabulary \cup \{\langle \text{BOS}\rangle\}\colon\,c(\token g)\geq \numberRepetitions\right\} \right|, \\
\label{eq:suffixCountPlus}
N_{r+}(g \bullet)           &= \left|\left\{ \token \in \vocabulary \cup \{\langle \text{EOS}\rangle\}\colon\,c(g \token)\geq \numberRepetitions\right\} \right|,
\end{align}

where $|\cdot|$ specifies the cardinality of a set. In plain words, these functions provide the number of different prefixes or suffixes a $n$-gram has been observed with exactly, or at least, $\numberRepetitions$ times. Notice that all definitions automatically extend to the case where $g=\emptyset$, and we will use the shorthand notations $N_{r}(\bullet) \equiv N_{r}(\emptyset\bullet)$ and $N_{r}(\bullet\bullet) \equiv  N_{r}(\bullet\emptyset\bullet)$. The juxtaposition of the empty $0$-gram to other \ngrams is understood as \e{no operation}, \eg $\token \emptyset \token' \equiv \token \token'$.

\subsubsection{N-gram Language Models}
$N$-gram language models provide a probabilistic description of language production that relies on two essential simplifying assumptions:
\begin{itemize}
    \item Sentences are independent of each other.
    \item Tokens are conditionally independent of previous history, given a finite (fixed) number, say $\modelOrder$, of preceding tokens.
\end{itemize}
According to these assumptions, the probability of sampling a document $\mathcal D _m$, consisting of $m$ sentences $\sentence_1,\,\sentence_2,\,\dots ,\,\sentence_m$, from a stream of natural language would be given by:
\begin{equation} \label{eq:probDocument}
P(\mathcal D _m \vert m) = \prod _{i = 1} ^m P (\sentence_i) 
\end{equation}
where, in turn, the probability of a sentence $\sentence_i = \token_{i,1}\token_{i,2}\cdots \token_{i, z_i}$ is:

\begin{equation} \label{eq:probSentence}
P(\sentence_i) = \prod _{j = 1} ^{z_i + 1} P(\token_{i,j}\vert \token_{i,j - N +1} \token_{i,j - N + 2} \cdots \token_{i,j-1})
\end{equation}

and we make the conventions $\token_{i,z_i +1} = \langle \text {EOS} \rangle$, $\token_{i,-1} = \token_{i,-2} = \cdots = \token_{i,-N+1}=\langle \text {BOS} \rangle$. As a parenthetical remark, we observe that the probability in Eq. \eqref{eq:probDocument} is conditional on the number $m$ of sentences that compose the document $\mathcal D _m$, as reflected by our notation.

\subsubsection{Kneser-Ney Smoothing}
We are now ready to present the (modified) Kneser-Ney \ngram model of order $\modelOrder$. The algorithm follows a general recursion scheme:
\begin{align}\label{eq:knRecursion}
p _{\text {KN}} (\token \vert g) &= \alpha(\token \vert g)+\gamma (g) \cdot p _{\text {KN}} (\token \vert L(g)),\\
\label{eq:knBaseCase}
p _{\text {KN}} (\token \vert \emptyset) &= \alpha(\token \vert \emptyset)+\gamma (\emptyset) \cdot \dfrac{1}{|\vocabulary|+ 1}.
\end{align}
The functions $\alpha(\cdot)$ and $\gamma(\cdot)$ are defined by:
\begin{equation} \label{eq:knAlpha}
\alpha(\token \vert g) = \begin{cases}
  \dfrac {\max (c_{\text{KN}}(g \token)-D(c_{\text{KN}}(g\token)), \,0)}{\sum _{\token' \in \vocabulary^*}  c_{\text {KN}}(g \token')} & \text{if } c(g)>0, \\
  0 & \text{otherwise},
\end{cases}
\end{equation}
and:
\begin{equation} \label{eq:knGamma} 
\gamma(g) = \begin{cases}
  \dfrac{\sum _{k = 1} ^ {\infty} D(k) N_n(g \bullet)}{\sum _{\token' \in \vocabulary^*} c_{\text {KN}}(g \token')} & \text{if } c(g)>0, \\
  1 & \text{otherwise},
\end{cases}
\end{equation}
where $\vocabulary^* \equiv \vocabulary \cup \{\langle \text{EOS} \rangle \}$. In turn, the modified count functions $c_{\text{KN}}(\cdot)$ are given by:
\begin{equation} \label{eq:knCounts}    
c_{\text{KN}}(\token_1 \token_2 \cdots \token_n) = \begin{cases}
  c(\token_1 \token_2 \cdots \token_n) & \text{if } n=\modelOrder, \\
  N_{1+}(\bullet \token_1 \token_2 \cdots \token_n) & \text{otherwise}.
\end{cases}
\end{equation}
Finally, $D(1),\, D(2),\ldots$ are free parameters satisfying $0<D(k)<1$, whose optimal values can be estimated from data. 
%----------------------------------- 
The parameter space is usually made finite by assuming $D(k) = D(\numberRepetitions)$ for some $\numberRepetitions$ and $k\geq \numberRepetitions$, in which case we can rewrite the infinite sum in \eqref{eq:knGamma} as:
$$
\sum _{k = 1} ^ {\infty} D(k) N_n(g \bullet) = \sum _{k = 1} ^ {r-1} D(k) N_n(g \bullet) +  D(\numberRepetitions) N_{\numberRepetitions+}(g \bullet).
$$
In our work we use $\numberRepetitions = 3$, as suggested by  \textcite{ChenGoodmanLMSmoothing:1996}.
%----------------------------------- 
We conclude this Section with a few remarks:
\begin{itemize}
    \item As shown in Eq.~\eqref{eq:knBaseCase}, the $1$-gram distribution is interpolated with the uniform distribution over the extended dictionary $\vocabulary^*$ (see next point). This is relevant to our application, since documents featuring tokens not seen during training would otherwise be assigned zero probability, \emph{i.e.} infinite perplexity.

    \item The conditional probability $p_{\text{KN}}(\token\vert g)$ is defined for $\token \in \vocabulary^*$, which is the reason why the denominator in the second term of \eqref{eq:knBaseCase} is $|\vocabulary| + 1$. In particular $p_{\text{KN}}(\langle \text{EOS}\rangle \vert g)$ gives the probability that the $\modelOrder-1$ gram $g$ terminates the sentence, whereas $p_{\text{KN}}(\langle \text{BOS}\rangle \vert g)$ is not defined (and indeed has no meaningful interpretation).
    
    \item The modified count \eqref{eq:knCounts} treats differently \ngrams of top and lower orders, respectively. We can get some intuition about the usage of the prefix counts $N_{1+}(\bullet \token_1 \token_2\cdots \token_n)$ for lower orders \ngrams by noticing that these put more weights into tokens that are observed with many contexts, so that we may interpret $N_{1+}(\bullet \token_1 \token_2\cdots \token_n)$ as (being proportional) to a kind of \e{continuation probability} \parencite{ChenGoodmanLMSmoothing:1996}. If we want to insist with such a probabilistic interpretation, we may observe that \e{\eg} $\frac{N_{1+}(\bullet \token)}{N_{1+}(\bullet \bullet)}$ estimates the probability that, out of all distinct bigrams $\token_1\token_2$ found in a text corpus, the second token $\token_2=\token$.
    
    \item Kneser-Ney smoothing as originally formulated \parencite{kneser1995} uses a constant discount function $D(k) \equiv D$, corresponding to $\numberRepetitions=1$ in the present formulation. The proposal of a non-constant discount function $D(k)$ was advanced by \textcite{ChenGoodmanLMSmoothing:1996}. In the present paper we used a constant discount value of $D = 0.75$.
\end{itemize}

\subsection{The Likelihood Ratio Framework}\label{sec-the-likelihood-ratio-framework}
The Likelihood Ratio Framework for forensic sciences approaches the forensic analysis as an assessment of two competing hypotheses. Typically, in \AV, these hypotheses are: 
\begin{itemize}
    \item The \emph{Prosecution Hypothesis} ($H_p$): The disputed document ($\Dunk$) was written by the author of the known document(s) (the defendant, $\A$).
    
    \item The \emph{Defense Hypothesis} ($H_d$): The disputed document ($\Dunk$) was not written by the author of the known document(s) (the defendant, $\A$), but instead by someone else, $\notA$.
\end{itemize}

The police's or stakeholders' interest in the given evidence ($E$), comprising $\Dunk$ and $\DA$, is probabilistically expressed as the ratio of the following two Bayesian conditional probabilities:

\begin{equation}{
\text{odds}_{\text{post}}=\frac{P(H_p|E = \lbrace \Dunk, \DA \rbrace)}{P(H_d|E = \lbrace \Dunk, \DA \rbrace)}
}\label{eq-postodds}\end{equation}

The numerator of Equation~\ref{eq-postodds} states \emph{Given the evidence, what is the probability of} \(H_p\) \emph{being true?} The denominator of Equation~\ref{eq-postodds} states \emph{Given the same evidence, what is the probability of} \(H_d\) \emph{being true?} If the ratio is greater than one, the evidence supports \(H_p\) more than \(H_d\); the larger than one is the ratio, the stronger the support for \(H_p\). Similarly, if the ratio is smaller than one, the evidence supports \(H_d\) more than \(H_p\); the smaller than one is the ratio, the stronger the support for \(H_d\). The ratio provided in Equation~\ref{eq-postodds} is known as the posterior odds in Bayes' Theorem. These odds can be estimated using Bayes' Theorem, which, in its odds form is given in Equation~\ref{eq-bayes}.

\begin{equation}{
\underbrace{\frac{P(H_p\ |E)}{P(H_d\ |E)}}_{\text{posterior\ odds}}\ =\underbrace{\frac{P(H_p)}{P(H_d)}}_{\text{prior\ odds}} \times \underbrace{\frac{P(E|H_p)}{P(E|H_d)}}_{\text{likelihood\ ratio}} 
}\label{eq-bayes}\end{equation}

Equation~\ref{eq-bayes} states that posterior odds can be estimated as the product of the prior odds and the likelihood ratio. The prior odds represents the trier-of-fact's belief about the two competing hypotheses (\(H_p\) and \(H_d\)) before observing the evidence (\ie linguistic text evidence). The forensic scientist's task is to evaluate the strength of the evidence, quantified in the form of a likelihood ratio. In other words, Equation~\ref{eq-bayes} states that the belief of the trier-of-fact (prior odds), which was formed or influenced by other evidence presented prior to the linguistic text evidence, is updated through the evaluation of the linguistic text evidence (\e{LR}), resulting in an up-to-date belief (posterior odds).

It is important to note that the forensic expert cannot estimate and legally must not refer to the posterior odds. They cannot estimate it logically because they are not privy of the trier-of-fact's belief regarding the hypotheses and do not have access to other pieces of evidence presented prior to the linguistic text evidence. Legally, they must not refer to it because the posterior odds represents the ultimate issue of being guilty or innocence \parencite[p. 96]{lynch2003} and this determination is a specific responsibility reserved only for the trier-of-fact. 
%----------------------------------- 
The odds form of Bayes' Theorem explicitly outlines the role of the forensic scientist. In the Likelihood Ratio framework, their responsibility lies in estimating the strength of evidence in the form of a likelihood ratio. As indicated by the rightmost term of Equation~\ref{eq-bayes}, the likelihood ratio is the ratio between the probability of the evidence given \(H_p\) and the probability of the same evidence given \(H_d\). The Likelihood Ratio serves as a gradient, quantifying how much more likely the evidence is given one hypothesis over the other \parencite{evett2000}. When the \e{LR} is greater than one, it provides stronger support for \(H_p\), whereas when it is smaller than one, it offers stronger support for \(H_d\). 
%----------------------------------- 
The \e{LR} framework is now accepted as a logically and legally correct framework for interpreting forensic evidential analyses \parencite{aitken2021, robertson2016}, having been studied for various evidential types \parencite{bolck2017, bolck2009, davis2012, morrison2009, neumann2012, zadora2009}. The \e{LR} framework has also been supported by the relevant scientific and professional associations \parencite{expressi2011, associationofforensicscienceproviders2009,ballantyne2017, forensicscienceregulator2021,americanstatisticalassociation2019, europeannetworkofeuropeannetworkofforensicscienceinstitutes2015}.

\subsubsection{The Log-Likelihood Ratio Cost ($\cllr$)}
In numerous source-identification tasks, the system’s performance is commonly evaluated in terms of identification Accuracy and/or identification error rate. Metrics such as precision, recall, and equal error rate (EER) are typical in this regard. While these metrics offer a quick overview of performance and are intuitively easy to comprehend, they are not suitable for assessing likelihood ratio-based inference systems  \parencite[p. 93]{morrison_measuring_2011}.
%------------------------------------ 
Primarily based on categorical decisions with specific thresholds, these metrics are inadequate for evaluating such systems. Firstly, metrics relying on categorical thresholding implicitly assess how effectively the system correctly identifies the source of evidence. In other words, these metrics implicitly pertain to the strength of the hypothesis rather than the strength of the evidence \parencite[p. 93]{morrison_measuring_2011}.
%------------------------------------ 
Secondly, categorical metrics fail to capture the gradient nature of likelihood ratios \parencite[p. 93]{morrison_measuring_2011}. For instance, in a same-author comparison, if one system returns $\log_{10}(\text{\e{LR}}) = 3$ while another yields $\log_{10}(\text{\e{LR}}) = 2$, the former is superior as it supports the correct hypothesis more strongly. However, using a categorical metric would assess both systems in the same way, disregarding the differing strengths of evidence supporting the correct hypothesis.

The conventional metric employed for evaluating the performance of a system within the Likelihood Ratio Framework is the \textit{log-likelihood ratio cost} ($\cllr$). The computation of the $\cllr$ involves utilizing a set of \e{LRs} for same-source scenarios (\eg same-authorship, \classY) and a set for different-source scenarios (\eg different-authorship, \classN). The computation is performed using the following equation \parencite{BRUMMER2006230, ramos_information-theoretical_2013, van_leeuwen_introduction_2007}

\begin{equation}\label{eq:cllr}
\cllr = \frac{1}{2}\left(\frac{1}{\gamma}\sum_{i}^{\gamma}\log_2\left(1+\frac{1}{\e{LR}_i}\right)+\frac{1}{\eta}\sum_{j}^{\eta}\log_2\bigg(1+\e{LR}_j\bigg)\right) 
\end{equation}

where, $\gamma$ and $\eta$ are the number of \classYdash{} and \classNdash{cases}, and $\e{LR}_i$ and $\e{LR}_j$ are the \e{LRs} estimated for the \classYdash{} and \classNdash{cases}, respectively. $\cllr$ considers the magnitude of calculated \e{LR} values, imposing appropriate penalties based on their values. Within the $\cllr$ metric, \e{LRs} supporting counter-factual hypotheses, or contrary-to-fact \e{LRs} (with $\log_{10}(\e{LR}) < 0$ for \classYdash{cases} and $\log_{10}(\e{LR}) > 0$ for \classNdash{cases}), incur significant penalties. The extent of the penalty is proportional to the deviation of the \e{LRs} from unity. Optimal performance is attained when $\cllr$ equals 0, and performance gradually diminishes as $\cllr$ approaches and surpasses 1. Therefore, a lower $\cllr$ value corresponds to better performance. 

$\cllr$ serves as an assessment metric for the overall performance of an \e{LR}-based inference system and can be decomposed into two distinct metrics: $\cllrmin$ and $\cllrcal$. The former evaluates the discrimination performance and the latter the calibration performance of the system. The $\cllrmin$ is determined by computing the losses for the optimized \e{LRs} through the non-parametric pool-adjacent-violators algorithm. The disparity between $\cllr$ and $\cllrmin$ is expressed as $\cllrcal$; in other words, $\cllr$ equals the sum of discrimination loss ($\cllrmin$) and calibration loss ($\cllrcal$). More in-depth descriptions of $\cllr$, encompassing both $\cllrmin$ and $\cllrcal$, can be found in work by \textcite{BRUMMER2006230}, \textcite{drygajlo_methodological_2015}, and  \textcite{meuwly_guideline_2017}. The $\cllr$ has already been effectively utilized for evaluating \e{LR}-based forensic text comparison systems \parencite{ishihara2021,ishihara_weight_2023,ishihara_likelihood_2022,NiniTheoryLingAA:2023}.

\subsection{Baseline AV Methods} \label{sec-baseline_av_methods} 
For the evaluation of \ourApproach, we selected \numberOfBaselines baseline \AV methods that were covered in our literature review (see Section~\ref{sec-introduction}). Specifically, we chose \imOrg \parencite{KoppelWinter2DocsBy1:2014}, \coav \parencite{HalvaniARES:2017}, \taveer \parencite{HalvaniARES:2020}, \janithDiffVec (\janithLR \parencite{WeerasingheFeVecDiff:2021} + \janithMLP \parencite{WeerasingheFeVecDiff:2020}), \adhominem \parencite{BoenninghoffExplainableAV:2019} and \luarAV \parencite{RiveraSotoLUAR:2021}. The selection of these methods was guided by several key considerations. \adhominem, \janithDiffVec (\janithLR + \janithMLP) and \taveer were the three top approaches in the PAN AV 2020 competition \parencite{PANOverviewAV:2020}. Furthermore, \janithDiffVecLR ranked third in the PAN AV 2021 competition \parencite{PANOverviewAV:2021}, closely followed by the second-placed method. Regarding \imOrg and \coav, \textcite{HalvaniPhD:2021} demonstrated that, across ten different corpora, \imOrg achieved the highest average performance, while \coav ranked second (both in terms of Accuracy and AUC) among twelve evaluated \AV methods. \coav has also attracted attention in the \AV community. In independent studies, \textcite{HalvaniARES:2017} and \textcite{BevendorffUnmasking:2019} came to the conclusion that \coav is on par with Bagnall's well-known RNN-based \AV method \parencite{BagnallRNN:2015}, the winning approach of the PAN AV 2015 competition \parencite{PANOverviewAV:2015}. Moreover, \coav was selected as one of six baseline methods in the PAN AV 2024 competition, where it outperformed four of them \parencite[Table~5]{PANOverviewAV:2024}. Note that with respect to \coav, \taveer, \janithDiffVec (\janithLR + \janithMLP) and \adhominem, we used the \textbf{original} implementations\footnote{The original implementations were downloaded from the following official Github repositories:\newline
\adhominem $\rightarrow$ \url{https://github.com/boenninghoff/AdHominem}\newline
\janithDiffVecMLP $\rightarrow$ \url{https://github.com/janithnw/pan2020_authorship_verification}\newline
\janithDiffVecLR $\rightarrow$ \url{https://github.com/janithnw/pan2021_authorship_verification}} of the authors. Regarding \imOrg, we have used the same implementation as used in \parencite{HalvaniPhD:2021}. The hyperparameter optimization for \janithDiffVec (\janithLR + \janithMLP) and \adhominem was performed by splitting each training corpus $\CorpusTrain$ into a training subset $\CorpusTrainStar$ and a validation set $\CorpusValidation$, following an 80/20 ratio. 

\subsubsection{\luarAV Approach} \label{sec-baseline-luar} 
In the following we describe our implementation of the \luarAV method introduced in section \ref{sec-languagemodels-compression}. \luarAV extends the SBERT architecture with an enhanced sampling strategy. The method was trained by dividing the documents into 16 equally sized segments, from which sequences of 32 tokens are drawn at random. A self-attention layer processed the resulting embeddings, which were then aggregated via max pooling and passed through a final linear layer to yield a 512-dimensional author embedding \parencite{RiveraSotoLUAR:2021}. The chosen model\footnote{The original model \textbf{rrivera1849/LUAR-MUD} is available at \url{https://huggingface.co/rrivera1849/LUAR-MUD}} was trained using contrastive loss on the \emph{Reddit Million User Dataset} (MUD), which contains over 300 million Reddit comments from one million users. The texts were sampled from users contributing 100–1000 posts between July 2015 and June 2016 \parencite{khan-etal-deepMetricLearningforAccountLinking:2021}.
\\
\\
Given a verification case $\Problem = (\Dunk, \DA)$, we first used \luarAV to embed both documents into their corresponding feature vectors $\featVector_{\unknown}$ and $\featVector_{\A}$. As part of this embedding process, the texts of the known author were concatenated and segmentation was performed according to \luarAV’s 16×32 token windowing scheme \parencite{RiveraSotoLUAR:2021}. Because the original \luarAV method was evaluated using ranking-based metrics, we extended the approach with a classification scheme inspired by the \taveer method \parencite{HalvaniARES:2020}. This allows us to evaluate \luarAV's embeddings within a binary verification setting. To adapt the pretrained \luarAV model to the POSNoise masked texts, we fine-tuned the transformer once in a universal setting across all training corpora. The fine-tuning was carried out using a variant of the contrastive loss \parencite{HadsellDimReductionInvMapping:2006} in which the squared Euclidean distance was replaced by the Manhattan distance with a margin of $m = 2.0$. Given two feature vectors, $\featVector_{\unknown}$ and $\featVector_{\A}$, the corresponding loss function is formally defined as follows:
\begin{equation}
L = 
y \, \lVert \featVector_{\unknown} - \featVector_{\A} \rVert_{1} \;+\; (1 - y)\, \max\bigl(0,\, m - \lVert \featVector_{\unknown} - \featVector_{\A} \rVert_{1}\bigr)
\end{equation}
Here, $\lVert\cdot\rVert_{1}$ denotes the L1 norm, $y = 1$ if $\A = \unknown$ and $y = 0$ otherwise. 
%--------------------------------------- 
Fine-tuning was performed with a batch size of 16, a learning rate of 1e-6, five epochs, a weight decay of 0.01 and a linear warmup ratio of 0.1. In a second step, threshold-based classification was applied. For this, we first normalize the finetuned \luarAV embeddings using the L1 norm, following \textcite{HalvaniARES:2020}, which ensures that the resulting values lie within the interval $[0,2]$. The closeness between each pair of embedding vectors $(\featVector_{\unknown}, \featVector_{\A})$ is then computed via the the Manhattan distance. To transform \luarAV into a classifier, we introduce a corpus-specific decision threshold $\theta_F$, defined as the median of all pairwise distances in the training data. The resulting distance $d$ is subsequently converted into a calibrated similarity score $s \in [0,1]$ using the following piecewise linear function proposed by \textcite{HalvaniARES:2020}:
\begin{equation}
    \text{sim}(d, d_{\text{max}}, \theta_F) =
    \begin{cases}
    1 - \dfrac{d}{2\theta_F}, & \text{if } d \leq \theta_F \\
    \dfrac{1}{2} - \dfrac{d - \theta_F}{2(d_{\text{max}} - \theta_F)}, & \text{otherwise}
    \end{cases}
    \label{eq:similarity_function}
\end{equation}
Here, $d_{\text{max}}$ is set to 2 (the upper bound of the L1 norm). The sim$(\cdot)$ function guarantees that the threshold $\theta_F$ maps to the decision boundary $s = 0.5$. If the resulting similarity score exceeds 0.5, we assume same authorship (\classY), otherwise different authorship (\classN).

\section{Qualitative Data Exploration}\label{sec-appendix-a}

\definecolor{colorlow}{rgb}{0.95, 0.00, 0.00}
\newcommand{\code}[2]{%
    \begingroup\setlength{\fboxsep}{0pt}%
    \colorbox{#1}{\texttt{\hspace*{1pt}\vphantom{Ay}#2\hspace*{1pt}}}%
    \endgroup
}

This Section contains a series of examples of how an analyst can visualize the results of \ourApproach. To aid readability, we replaced the original placeholders used in the POSNoise paper (see \cite[Table~2]{HalvaniPOSNoise:2021}) within the visualizations with their corresponding POS tags, according to the following mapping: 
\begin{align*} 
\textrm{\texttt{\#}} &\rightarrow \textrm{\texttt{N} (nouns),} \\ 
\textrm{\texttt{Ø}} &\rightarrow \textrm{\texttt{V} (verbs),} \\ 
\textrm{\texttt{@}} &\rightarrow \textrm{\texttt{J} (adjectives),} \\ 
\textrm{\texttt{©}} &\rightarrow \textrm{\texttt{B} (adverbs),} \\
\textrm{\texttt{§}} &\rightarrow \textrm{\texttt{P} (proper nouns).} 
\end{align*}
Because the magnitude of $\lambda_G$ varies depending on the corpus, the examples in this Section are color-coded using the z-score of $\lambda_G$ using the mean and standard deviation of $\lambda_G$ for the $\Dunk$ being examined. Three shades of red are used, the lightest indicating $0.5 < \textrm{z-score}(\lambda_G) \leq 1$, then $1 < \textrm{z-score}(\lambda_G) \leq 2$, and the darkest indicating $\textrm{z-score}(\lambda_G) > 2$.
\newline\newline
\begin{center}\fbox{\begin{minipage}{0.95\textwidth}\code{colorlow!50}{\strut as}\allowbreak\code{colorlow!70}{\strut a}\allowbreak\code{colorlow!20}{\strut N}\allowbreak\code{colorlow!70}{\strut ,}\allowbreak\code{colorlow!70}{\strut i}\allowbreak\code{colorlow!0}{\strut ca}\allowbreak\code{colorlow!0}{\strut n't}\allowbreak\code{colorlow!0}{\strut V}\allowbreak\code{colorlow!0}{\strut too}\allowbreak\code{colorlow!20}{\strut many}\allowbreak\code{colorlow!0}{\strut of}\allowbreak\code{colorlow!0}{\strut them}\allowbreak\code{colorlow!0}{\strut wanting}\allowbreak\code{colorlow!0}{\strut to}\allowbreak\code{colorlow!70}{\strut go}\allowbreak\code{colorlow!0}{\strut the}\allowbreak\code{colorlow!0}{\strut J}\allowbreak\code{colorlow!0}{\strut N}\allowbreak\code{colorlow!0}{\strut to}\allowbreak\code{colorlow!0}{\strut V}\allowbreak\code{colorlow!0}{\strut a}\allowbreak\code{colorlow!20}{\strut N}\allowbreak\code{colorlow!0}{\strut too}\allowbreak\code{colorlow!0}{\strut .}\allowbreak\code{colorlow!0}{\strut [EOS]}\allowbreak\newline\code{colorlow!0}{\strut this}\allowbreak\code{colorlow!0}{\strut N}\allowbreak\code{colorlow!0}{\strut seems}\allowbreak\code{colorlow!70}{\strut to}\allowbreak\code{colorlow!0}{\strut be}\allowbreak\code{colorlow!0}{\strut already}\allowbreak\code{colorlow!0}{\strut B}\allowbreak\code{colorlow!0}{\strut V}\allowbreak\code{colorlow!0}{\strut a}\allowbreak\code{colorlow!70}{\strut little}\allowbreak\code{colorlow!0}{\strut more}\allowbreak\code{colorlow!70}{\strut B}\allowbreak\code{colorlow!0}{\strut in}\allowbreak\code{colorlow!0}{\strut another}\allowbreak\code{colorlow!0}{\strut N}\allowbreak\code{colorlow!50}{\strut :}\allowbreak\code{colorlow!50}{\strut P}\allowbreak\code{colorlow!0}{\strut .}\allowbreak\code{colorlow!0}{\strut [EOS]}\allowbreak\newline\code{colorlow!0}{\strut V}\allowbreak\code{colorlow!50}{\strut a}\allowbreak\code{colorlow!0}{\strut J}\allowbreak\code{colorlow!0}{\strut N}\allowbreak\code{colorlow!50}{\strut from}\allowbreak\code{colorlow!0}{\strut your}\allowbreak\code{colorlow!0}{\strut J}\allowbreak\code{colorlow!0}{\strut N}\allowbreak\code{colorlow!0}{\strut would}\allowbreak\code{colorlow!0}{\strut actually}\allowbreak\code{colorlow!0}{\strut V}\allowbreak\code{colorlow!0}{\strut N}\allowbreak\code{colorlow!0}{\strut about}\allowbreak\code{colorlow!0}{\strut your}\allowbreak\code{colorlow!0}{\strut N}\allowbreak\code{colorlow!0}{\strut in}\allowbreak\code{colorlow!0}{\strut your}\allowbreak\code{colorlow!0}{\strut J}\allowbreak\code{colorlow!0}{\strut N}\allowbreak\code{colorlow!0}{\strut ,}\allowbreak\code{colorlow!0}{\strut thus}\allowbreak\code{colorlow!0}{\strut V}\allowbreak\code{colorlow!0}{\strut the}\allowbreak\code{colorlow!0}{\strut N}\allowbreak\code{colorlow!0}{\strut of}\allowbreak\code{colorlow!0}{\strut N}\allowbreak\code{colorlow!0}{\strut you}\allowbreak\code{colorlow!70}{\strut 're}\allowbreak\code{colorlow!70}{\strut looking}\allowbreak\code{colorlow!70}{\strut for}\allowbreak\code{colorlow!0}{\strut (}\allowbreak\code{colorlow!0}{\strut at}\allowbreak\code{colorlow!0}{\strut J}\allowbreak\code{colorlow!0}{\strut as}\allowbreak\code{colorlow!0}{\strut V}\allowbreak\code{colorlow!20}{\strut in}\allowbreak\code{colorlow!50}{\strut a}\allowbreak\code{colorlow!20}{\strut J}\allowbreak\code{colorlow!0}{\strut N}\allowbreak\code{colorlow!70}{\strut )}\allowbreak\code{colorlow!20}{\strut .}\allowbreak\code{colorlow!0}{\strut [EOS]}\allowbreak\newline\code{colorlow!0}{\strut if}\allowbreak\code{colorlow!50}{\strut you}\allowbreak\code{colorlow!0}{\strut moved}\allowbreak\code{colorlow!0}{\strut to}\allowbreak\code{colorlow!0}{\strut a}\allowbreak\code{colorlow!0}{\strut J}\allowbreak\code{colorlow!0}{\strut N}\allowbreak\code{colorlow!0}{\strut (}\allowbreak\code{colorlow!20}{\strut be}\allowbreak\code{colorlow!0}{\strut it}\allowbreak\code{colorlow!0}{\strut N}\allowbreak\code{colorlow!0}{\strut or}\allowbreak\code{colorlow!0}{\strut N}\allowbreak\code{colorlow!0}{\strut )}\allowbreak\code{colorlow!70}{\strut for}\allowbreak\code{colorlow!0}{\strut a}\allowbreak\code{colorlow!0}{\strut N}\allowbreak\code{colorlow!0}{\strut ,}\allowbreak\code{colorlow!0}{\strut you}\allowbreak\code{colorlow!0}{\strut will}\allowbreak\code{colorlow!70}{\strut B}\allowbreak\code{colorlow!0}{\strut run}\allowbreak\code{colorlow!20}{\strut into}\allowbreak\code{colorlow!0}{\strut some}\allowbreak\code{colorlow!0}{\strut J}\allowbreak\code{colorlow!0}{\strut N}\allowbreak\code{colorlow!70}{\strut with}\allowbreak\code{colorlow!50}{\strut N}\allowbreak\code{colorlow!50}{\strut .}\allowbreak\code{colorlow!0}{\strut [EOS]}\allowbreak\newline\code{colorlow!0}{\strut this}\allowbreak\code{colorlow!70}{\strut is}\allowbreak\code{colorlow!0}{\strut V}\allowbreak\code{colorlow!70}{\strut -}\allowbreak\code{colorlow!0}{\strut and}\allowbreak\code{colorlow!0}{\strut -}\allowbreak\code{colorlow!0}{\strut J}\allowbreak\code{colorlow!0}{\strut N}\allowbreak\code{colorlow!0}{\strut .}\allowbreak\code{colorlow!0}{\strut [EOS]}
\end{minipage}}\end{center}
The first example is an unknown document $\Dunk$ taken from $\CorpusStack$ where the sentences above are the five sentences with the highest $\lambda_G$. Although the important tokens are highlighted following a color scheme, the interpretation of why they are important is left to the analyst using a manual analysis of the concordance lines. An exploration of the corpus reveals that it is the entire grammatical construction \quotetxt{as a N,} at the beginning of a sentence that is rare in the reference dataset but it is found in both the known and unknown documents. The use of \quotetxt{J N with N .} is only used by this author and another author in the entire corpus. Similarly, the phrase \quotetxt{you're looking for} is rare and only used by four other authors in the corpus. The combination of these three grammatical constructions is therefore already uniquely identifying this individual.

%\begin{center}\rule{0.5\linewidth}{0.5pt}\end{center}
\begin{center}\fbox{\begin{minipage}{0.95\textwidth}\code{colorlow!70}{\strut finally}\allowbreak\code{colorlow!50}{\strut ,}\allowbreak\code{colorlow!0}{\strut it}\allowbreak\code{colorlow!0}{\strut puts}\allowbreak\code{colorlow!0}{\strut an}\allowbreak\code{colorlow!0}{\strut ever}\allowbreak\code{colorlow!0}{\strut -}\allowbreak\code{colorlow!0}{\strut so}\allowbreak\code{colorlow!0}{\strut -}\allowbreak\code{colorlow!0}{\strut J}\allowbreak\code{colorlow!0}{\strut N}\allowbreak\code{colorlow!0}{\strut of}\allowbreak\code{colorlow!0}{\strut N}\allowbreak\code{colorlow!0}{\strut on}\allowbreak\code{colorlow!0}{\strut the}\allowbreak\code{colorlow!0}{\strut N}\allowbreak\code{colorlow!0}{\strut by}\allowbreak\code{colorlow!20}{\strut V}\allowbreak\code{colorlow!20}{\strut the}\allowbreak\code{colorlow!0}{\strut N}\allowbreak\code{colorlow!0}{\strut so}\allowbreak\code{colorlow!50}{\strut that}\allowbreak\code{colorlow!70}{\strut they}\allowbreak\code{colorlow!70}{\strut 're}\allowbreak\code{colorlow!70}{\strut trying}\allowbreak\code{colorlow!0}{\strut to}\allowbreak\code{colorlow!0}{\strut V}\allowbreak\code{colorlow!0}{\strut you}\allowbreak\code{colorlow!0}{\strut ,}\allowbreak\code{colorlow!0}{\strut rather}\allowbreak\code{colorlow!0}{\strut than}\allowbreak\code{colorlow!20}{\strut the}\allowbreak\code{colorlow!50}{\strut other}\allowbreak\code{colorlow!0}{\strut N}\allowbreak\code{colorlow!0}{\strut around}\allowbreak\code{colorlow!0}{\strut .}\allowbreak\code{colorlow!0}{\strut [EOS]}\allowbreak\newline\code{colorlow!50}{\strut to}\allowbreak\code{colorlow!70}{\strut begin}\allowbreak\code{colorlow!20}{\strut with}\allowbreak\code{colorlow!0}{\strut ,}\allowbreak\code{colorlow!20}{\strut it}\allowbreak\code{colorlow!50}{\strut 's}\allowbreak\code{colorlow!70}{\strut not}\allowbreak\code{colorlow!0}{\strut N}\allowbreak\code{colorlow!0}{\strut if}\allowbreak\code{colorlow!0}{\strut you}\allowbreak\code{colorlow!70}{\strut B}\allowbreak\code{colorlow!0}{\strut V}\allowbreak\code{colorlow!50}{\strut your}\allowbreak\code{colorlow!0}{\strut N}\allowbreak\code{colorlow!20}{\strut .}\allowbreak\code{colorlow!0}{\strut [EOS]}\allowbreak\newline\code{colorlow!0}{\strut it}\allowbreak\code{colorlow!70}{\strut would}\allowbreak\code{colorlow!0}{\strut be}\allowbreak\code{colorlow!20}{\strut J}\allowbreak\code{colorlow!50}{\strut to}\allowbreak\code{colorlow!20}{\strut V}\allowbreak\code{colorlow!70}{\strut that}\allowbreak\code{colorlow!0}{\strut you}\allowbreak\code{colorlow!50}{\strut 're}\allowbreak\code{colorlow!0}{\strut the}\allowbreak\code{colorlow!0}{\strut first}\allowbreak\code{colorlow!0}{\strut to}\allowbreak\code{colorlow!0}{\strut V}\allowbreak\code{colorlow!20}{\strut a}\allowbreak\code{colorlow!0}{\strut N}\allowbreak\code{colorlow!20}{\strut between}\allowbreak\code{colorlow!0}{\strut N}\allowbreak\code{colorlow!0}{\strut and}\allowbreak\code{colorlow!0}{\strut N}\allowbreak\code{colorlow!20}{\strut N}\allowbreak\code{colorlow!0}{\strut ,}\allowbreak\code{colorlow!20}{\strut for}\allowbreak\code{colorlow!0}{\strut N}\allowbreak\code{colorlow!50}{\strut ,}\allowbreak\code{colorlow!0}{\strut but}\allowbreak\code{colorlow!20}{\strut it}\allowbreak\code{colorlow!70}{\strut 's}\allowbreak\code{colorlow!0}{\strut B}\allowbreak\code{colorlow!0}{\strut J}\allowbreak\code{colorlow!0}{\strut J}\allowbreak\code{colorlow!0}{\strut other}\allowbreak\code{colorlow!0}{\strut N}\allowbreak\code{colorlow!0}{\strut on}\allowbreak\code{colorlow!0}{\strut the}\allowbreak\code{colorlow!0}{\strut N}\allowbreak\code{colorlow!0}{\strut and}\allowbreak\code{colorlow!0}{\strut then}\allowbreak\code{colorlow!50}{\strut V}\allowbreak\code{colorlow!0}{\strut or}\allowbreak\code{colorlow!0}{\strut V}\allowbreak\code{colorlow!0}{\strut them}\allowbreak\code{colorlow!0}{\strut with}\allowbreak\code{colorlow!50}{\strut your}\allowbreak\code{colorlow!50}{\strut J}\allowbreak\code{colorlow!0}{\strut N}\allowbreak\code{colorlow!0}{\strut .}\allowbreak\code{colorlow!0}{\strut [EOS]}\allowbreak\newline\code{colorlow!0}{\strut in}\allowbreak\code{colorlow!0}{\strut the}\allowbreak\code{colorlow!20}{\strut very}\allowbreak\code{colorlow!0}{\strut J}\allowbreak\code{colorlow!0}{\strut N}\allowbreak\code{colorlow!20}{\strut of}\allowbreak\code{colorlow!0}{\strut J}\allowbreak\code{colorlow!0}{\strut N}\allowbreak\code{colorlow!0}{\strut ,}\allowbreak\code{colorlow!0}{\strut you}\allowbreak\code{colorlow!70}{\strut 'd}\allowbreak\code{colorlow!70}{\strut need}\allowbreak\code{colorlow!0}{\strut to}\allowbreak\code{colorlow!0}{\strut V}\allowbreak\code{colorlow!0}{\strut P}\allowbreak\code{colorlow!70}{\strut 's}\allowbreak\code{colorlow!20}{\strut N}\allowbreak\code{colorlow!0}{\strut to}\allowbreak\code{colorlow!0}{\strut get}\allowbreak\code{colorlow!50}{\strut a}\allowbreak\code{colorlow!0}{\strut N}\allowbreak\code{colorlow!20}{\strut to}\allowbreak\code{colorlow!0}{\strut do}\allowbreak\code{colorlow!70}{\strut what}\allowbreak\code{colorlow!0}{\strut you}\allowbreak\code{colorlow!0}{\strut want}\allowbreak\code{colorlow!50}{\strut .}\allowbreak\code{colorlow!0}{\strut [EOS]}\allowbreak\newline\code{colorlow!0}{\strut you}\allowbreak\code{colorlow!70}{\strut could}\allowbreak\code{colorlow!70}{\strut V}\allowbreak\code{colorlow!0}{\strut the}\allowbreak\code{colorlow!0}{\strut J}\allowbreak\code{colorlow!0}{\strut N}\allowbreak\code{colorlow!0}{\strut under}\allowbreak\code{colorlow!0}{\strut your}\allowbreak\code{colorlow!0}{\strut N}\allowbreak\code{colorlow!20}{\strut N}\allowbreak\code{colorlow!0}{\strut to}\allowbreak\code{colorlow!0}{\strut V}\allowbreak\code{colorlow!0}{\strut an}\allowbreak\code{colorlow!20}{\strut N}\allowbreak\code{colorlow!0}{\strut of}\allowbreak\code{colorlow!50}{\strut your}\allowbreak\code{colorlow!0}{\strut N}\allowbreak\code{colorlow!0}{\strut of}\allowbreak\code{colorlow!0}{\strut N}\allowbreak\code{colorlow!50}{\strut ,}\allowbreak\code{colorlow!0}{\strut but}\allowbreak\code{colorlow!0}{\strut the}\allowbreak\code{colorlow!0}{\strut N}\allowbreak\code{colorlow!20}{\strut alone}\allowbreak\code{colorlow!0}{\strut does}\allowbreak\code{colorlow!50}{\strut n't}\allowbreak\code{colorlow!20}{\strut V}\allowbreak\code{colorlow!50}{\strut any}\allowbreak\code{colorlow!0}{\strut N}\allowbreak\code{colorlow!0}{\strut .}\allowbreak\code{colorlow!0}{\strut [EOS]}\end{minipage}}\end{center}
In this other $\Dunk$ from $\CorpusStack$, some of most important sequences highlighted by the model are the use of \quotetxt{finally,} and \quotetxt{to begin (with)} at the beginning of a sentence. Another important sequence that is hard to spot immediately and that becomes evident only after more manual investigation is the construction \quotetxt{get\ldots{}to do what} in the fourth sentence, which is only attested for this author.

%\begin{center}\rule{0.5\linewidth}{0.5pt}\end{center}
\begin{center}\fbox{\begin{minipage}{0.95\textwidth}\code{colorlow!0}{\strut i}\allowbreak\code{colorlow!20}{\strut am}\allowbreak\code{colorlow!70}{\strut a}\allowbreak\code{colorlow!70}{\strut little}\allowbreak\code{colorlow!50}{\strut J}\allowbreak\code{colorlow!0}{\strut between}\allowbreak\code{colorlow!0}{\strut my}\allowbreak\code{colorlow!0}{\strut N}\allowbreak\code{colorlow!0}{\strut [EOS]}\allowbreak\newline\code{colorlow!0}{\strut so}\allowbreak\code{colorlow!50}{\strut what}\allowbreak\code{colorlow!70}{\strut did}\allowbreak\code{colorlow!50}{\strut P}\allowbreak\code{colorlow!20}{\strut do}\allowbreak\code{colorlow!0}{\strut ?}\allowbreak\code{colorlow!0}{\strut [EOS]}\allowbreak\newline\code{colorlow!0}{\strut have}\allowbreak\code{colorlow!0}{\strut you}\allowbreak\code{colorlow!70}{\strut wanted}\allowbreak\code{colorlow!20}{\strut to}\allowbreak\code{colorlow!50}{\strut be}\allowbreak\code{colorlow!70}{\strut ?}\allowbreak\code{colorlow!0}{\strut [EOS]}\allowbreak\newline\code{colorlow!20}{\strut how}\allowbreak\code{colorlow!0}{\strut do}\allowbreak\code{colorlow!50}{\strut i}\allowbreak\code{colorlow!70}{\strut get}\allowbreak\code{colorlow!70}{\strut there}\allowbreak\code{colorlow!0}{\strut ?}\allowbreak\code{colorlow!0}{\strut [EOS]}\allowbreak\newline\code{colorlow!50}{\strut she}\allowbreak\code{colorlow!70}{\strut did}\allowbreak\code{colorlow!70}{\strut n't}\allowbreak\code{colorlow!0}{\strut know}\allowbreak\code{colorlow!20}{\strut what}\allowbreak\code{colorlow!0}{\strut to}\allowbreak\code{colorlow!0}{\strut think}\allowbreak\code{colorlow!0}{\strut at}\allowbreak\code{colorlow!0}{\strut first}\allowbreak\code{colorlow!0}{\strut .}\allowbreak\code{colorlow!0}{\strut [EOS]}\end{minipage}}\end{center}

This is an example from $\CorpusPJ$, which is the easiest of the corpora to interpret, probably because of the non-standard, unedited nature of the language used in this genre. In this example, the author uses the phrase \quotetxt{you wanted} frequently, as well as the 5-gram \quotetxt{i am a little J}, which is only found in one other author in the corpus. The most interesting pattern is the phrase \quotetxt{she did n't}, which seems completely unremarkable but it is used only by two other authors in the entire corpus and never at the beginning of a sentence. This could be taken as an instance of content interference but, in reality, there are 417 occurrences of the pronoun \quotetxt{she} in the corpus. Instead, this is an example that reflects a larger unit for this author: beginning a sentence, which in this case is a new chat message, using the pronoun and the contracted form of \quotetxt{did not} with an apostrophe and then followed by a mental verb (\quotetxt{think} in the known sample) to describe a mental state.

%\begin{center}\rule{0.5\linewidth}{0.5pt}\end{center}
\begin{center}\fbox{\begin{minipage}{0.95\textwidth}\code{colorlow!0}{\strut it}\allowbreak\code{colorlow!50}{\strut is}\allowbreak\code{colorlow!50}{\strut J}\allowbreak\code{colorlow!70}{\strut enough}\allowbreak\code{colorlow!70}{\strut to}\allowbreak\code{colorlow!70}{\strut be}\allowbreak\code{colorlow!70}{\strut B}\allowbreak\code{colorlow!50}{\strut V}\allowbreak\code{colorlow!0}{\strut in}\allowbreak\code{colorlow!20}{\strut any}\allowbreak\code{colorlow!0}{\strut J}\allowbreak\code{colorlow!0}{\strut N}\allowbreak\code{colorlow!50}{\strut .}\allowbreak\code{colorlow!0}{\strut [EOS]}\allowbreak\newline\code{colorlow!70}{\strut however}\allowbreak\code{colorlow!50}{\strut ,}\allowbreak\code{colorlow!0}{\strut this}\allowbreak\code{colorlow!20}{\strut N}\allowbreak\code{colorlow!0}{\strut of}\allowbreak\code{colorlow!0}{\strut N}\allowbreak\code{colorlow!0}{\strut does}\allowbreak\code{colorlow!0}{\strut not}\allowbreak\code{colorlow!0}{\strut V}\allowbreak\code{colorlow!0}{\strut V}\allowbreak\code{colorlow!0}{\strut the}\allowbreak\code{colorlow!0}{\strut N}\allowbreak\code{colorlow!0}{\strut ,}\allowbreak\code{colorlow!20}{\strut find}\allowbreak\code{colorlow!0}{\strut the}\allowbreak\code{colorlow!0}{\strut N}\allowbreak\code{colorlow!0}{\strut from}\allowbreak\code{colorlow!0}{\strut the}\allowbreak\code{colorlow!0}{\strut N}\allowbreak\code{colorlow!50}{\strut :}\allowbreak\code{colorlow!50}{\strut the}\allowbreak\code{colorlow!0}{\strut N}\allowbreak\code{colorlow!20}{\strut N}\allowbreak\code{colorlow!0}{\strut ,}\allowbreak\code{colorlow!20}{\strut unlike}\allowbreak\code{colorlow!0}{\strut other}\allowbreak\code{colorlow!0}{\strut N}\allowbreak\code{colorlow!50}{\strut ,}\allowbreak\code{colorlow!0}{\strut is}\allowbreak\code{colorlow!70}{\strut not}\allowbreak\code{colorlow!50}{\strut J}\allowbreak\code{colorlow!0}{\strut .}\allowbreak\code{colorlow!0}{\strut [EOS]}\allowbreak\newline\code{colorlow!20}{\strut the}\allowbreak\code{colorlow!70}{\strut first}\allowbreak\code{colorlow!0}{\strut of}\allowbreak\code{colorlow!0}{\strut these}\allowbreak\code{colorlow!0}{\strut N}\allowbreak\code{colorlow!20}{\strut is}\allowbreak\code{colorlow!0}{\strut a}\allowbreak\code{colorlow!0}{\strut set}\allowbreak\code{colorlow!0}{\strut of}\allowbreak\code{colorlow!20}{\strut V}\allowbreak\code{colorlow!0}{\strut N}\allowbreak\code{colorlow!0}{\strut V}\allowbreak\code{colorlow!0}{\strut with}\allowbreak\code{colorlow!0}{\strut a}\allowbreak\code{colorlow!0}{\strut N}\allowbreak\code{colorlow!0}{\strut of}\allowbreak\code{colorlow!20}{\strut N}\allowbreak\code{colorlow!20}{\strut N}\allowbreak\code{colorlow!0}{\strut :}\allowbreak\code{colorlow!0}{\strut N}\allowbreak\code{colorlow!0}{\strut abyssal}\allowbreak\code{colorlow!0}{\strut a}\allowbreak\code{colorlow!0}{\strut -}\allowbreak\code{colorlow!0}{\strut N}\allowbreak\code{colorlow!0}{\strut N}\allowbreak\code{colorlow!0}{\strut V}\allowbreak\code{colorlow!0}{\strut P}\allowbreak\code{colorlow!20}{\strut -}\allowbreak\code{colorlow!0}{\strut N}\allowbreak\code{colorlow!0}{\strut the}\allowbreak\code{colorlow!0}{\strut N}\allowbreak\code{colorlow!0}{\strut of}\allowbreak\code{colorlow!70}{\strut this}\allowbreak\code{colorlow!20}{\strut N}\allowbreak\code{colorlow!0}{\strut of}\allowbreak\code{colorlow!0}{\strut N}\allowbreak\code{colorlow!0}{\strut in}\allowbreak\code{colorlow!0}{\strut J}\allowbreak\code{colorlow!70}{\strut P}\allowbreak\code{colorlow!70}{\strut is}\allowbreak\code{colorlow!0}{\strut quite}\allowbreak\code{colorlow!0}{\strut J}\allowbreak\code{colorlow!0}{\strut .}\allowbreak\code{colorlow!0}{\strut [EOS]}\allowbreak\newline\code{colorlow!20}{\strut the}\allowbreak\code{colorlow!20}{\strut only}\allowbreak\code{colorlow!0}{\strut N}\allowbreak\code{colorlow!0}{\strut we}\allowbreak\code{colorlow!50}{\strut set}\allowbreak\code{colorlow!70}{\strut is}\allowbreak\code{colorlow!0}{\strut N}\allowbreak\code{colorlow!20}{\strut :}\allowbreak\code{colorlow!50}{\strut the}\allowbreak\code{colorlow!0}{\strut N}\allowbreak\code{colorlow!0}{\strut of}\allowbreak\code{colorlow!0}{\strut N}\allowbreak\code{colorlow!0}{\strut ,}\allowbreak\code{colorlow!20}{\strut even}\allowbreak\code{colorlow!0}{\strut if}\allowbreak\code{colorlow!0}{\strut it}\allowbreak\code{colorlow!50}{\strut is}\allowbreak\code{colorlow!50}{\strut J}\allowbreak\code{colorlow!0}{\strut ,}\allowbreak\code{colorlow!0}{\strut V}\allowbreak\code{colorlow!0}{\strut the}\allowbreak\code{colorlow!0}{\strut N}\allowbreak\code{colorlow!0}{\strut of}\allowbreak\code{colorlow!0}{\strut other}\allowbreak\code{colorlow!0}{\strut N}\allowbreak\code{colorlow!0}{\strut .}\allowbreak\code{colorlow!0}{\strut [EOS]}\end{minipage}}\end{center}

This is an example from $\CorpusACL$. Apart from the use of \quotetxt{however,} at the beginning of a sentence, this author uses the sequence \quotetxt{is (not) J enough to be B V}, which is only found in their writing. 

%\begin{center}\rule{0.5\linewidth}{0.5pt}\end{center}
\begin{center}\fbox{\begin{minipage}{0.95\textwidth}\code{colorlow!0}{\strut the}\allowbreak\code{colorlow!70}{\strut V}\allowbreak\code{colorlow!20}{\strut N}\allowbreak\code{colorlow!0}{\strut is}\allowbreak\code{colorlow!0}{\strut a}\allowbreak\code{colorlow!20}{\strut J}\allowbreak\code{colorlow!0}{\strut N}\allowbreak\code{colorlow!50}{\strut N}\allowbreak\code{colorlow!0}{\strut that}\allowbreak\code{colorlow!50}{\strut V}\allowbreak\code{colorlow!0}{\strut various}\allowbreak\code{colorlow!0}{\strut J}\allowbreak\code{colorlow!20}{\strut J}\allowbreak\code{colorlow!0}{\strut N}\allowbreak\code{colorlow!50}{\strut that}\allowbreak\code{colorlow!0}{\strut can}\allowbreak\code{colorlow!50}{\strut be}\allowbreak\code{colorlow!70}{\strut B}\allowbreak\code{colorlow!50}{\strut V}\allowbreak\code{colorlow!70}{\strut to}\allowbreak\code{colorlow!0}{\strut V}\allowbreak\code{colorlow!50}{\strut N}\allowbreak\code{colorlow!0}{\strut J}\allowbreak\code{colorlow!20}{\strut N}\allowbreak\code{colorlow!0}{\strut .}\allowbreak\code{colorlow!0}{\strut [EOS]}\allowbreak\newline\code{colorlow!0}{\strut the}\allowbreak\code{colorlow!0}{\strut N}\allowbreak\code{colorlow!0}{\strut V}\allowbreak\code{colorlow!0}{\strut the}\allowbreak\code{colorlow!0}{\strut N}\allowbreak\code{colorlow!20}{\strut -}\allowbreak\code{colorlow!20}{\strut N}\allowbreak\code{colorlow!0}{\strut J}\allowbreak\code{colorlow!20}{\strut N}\allowbreak\code{colorlow!0}{\strut P}\allowbreak\code{colorlow!0}{\strut ,}\allowbreak\code{colorlow!0}{\strut the}\allowbreak\code{colorlow!50}{\strut J}\allowbreak\code{colorlow!0}{\strut N}\allowbreak\code{colorlow!70}{\strut between}\allowbreak\code{colorlow!0}{\strut P}\allowbreak\code{colorlow!20}{\strut and}\allowbreak\code{colorlow!50}{\strut P}\allowbreak\code{colorlow!0}{\strut N}\allowbreak\code{colorlow!20}{\strut N}\allowbreak\code{colorlow!0}{\strut of}\allowbreak\code{colorlow!0}{\strut N}\allowbreak\code{colorlow!20}{\strut ,}\allowbreak\code{colorlow!0}{\strut J}\allowbreak\code{colorlow!0}{\strut N}\allowbreak\code{colorlow!50}{\strut ,}\allowbreak\code{colorlow!20}{\strut and}\allowbreak\code{colorlow!50}{\strut N}\allowbreak\code{colorlow!20}{\strut N}\allowbreak\code{colorlow!0}{\strut N}\allowbreak\code{colorlow!20}{\strut .}\allowbreak\code{colorlow!0}{\strut [EOS]}\allowbreak\newline\code{colorlow!0}{\strut V}\allowbreak\code{colorlow!70}{\strut N}\allowbreak\code{colorlow!0}{\strut V}\allowbreak\code{colorlow!0}{\strut by}\allowbreak\code{colorlow!50}{\strut N}\allowbreak\code{colorlow!0}{\strut ,}\allowbreak\code{colorlow!0}{\strut the}\allowbreak\code{colorlow!70}{\strut P}\allowbreak\code{colorlow!0}{\strut -}\allowbreak\code{colorlow!0}{\strut P}\allowbreak\code{colorlow!0}{\strut N}\allowbreak\code{colorlow!0}{\strut V}\allowbreak\code{colorlow!0}{\strut J}\allowbreak\code{colorlow!0}{\strut N}\allowbreak\code{colorlow!20}{\strut in}\allowbreak\code{colorlow!0}{\strut the}\allowbreak\code{colorlow!0}{\strut given}\allowbreak\code{colorlow!0}{\strut set}\allowbreak\code{colorlow!50}{\strut of}\allowbreak\code{colorlow!20}{\strut N}\allowbreak\code{colorlow!0}{\strut ,}\allowbreak\code{colorlow!0}{\strut whether}\allowbreak\code{colorlow!0}{\strut they}\allowbreak\code{colorlow!0}{\strut me}\allowbreak\code{colorlow!0}{\strut due}\allowbreak\code{colorlow!0}{\strut to}\allowbreak\code{colorlow!0}{\strut J}\allowbreak\code{colorlow!0}{\strut ,}\allowbreak\code{colorlow!0}{\strut J}\allowbreak\code{colorlow!0}{\strut ,}\allowbreak\code{colorlow!0}{\strut J}\allowbreak\code{colorlow!0}{\strut ,}\allowbreak\code{colorlow!0}{\strut or}\allowbreak\code{colorlow!50}{\strut N}\allowbreak\code{colorlow!50}{\strut N}\allowbreak\code{colorlow!20}{\strut .}\allowbreak\code{colorlow!0}{\strut [EOS]}\allowbreak\newline\code{colorlow!0}{\strut the}\allowbreak\code{colorlow!70}{\strut other}\allowbreak\code{colorlow!20}{\strut N}\allowbreak\code{colorlow!0}{\strut is}\allowbreak\code{colorlow!20}{\strut the}\allowbreak\code{colorlow!0}{\strut J}\allowbreak\code{colorlow!0}{\strut N}\allowbreak\code{colorlow!0}{\strut of}\allowbreak\code{colorlow!20}{\strut the}\allowbreak\code{colorlow!0}{\strut N}\allowbreak\code{colorlow!0}{\strut N}\allowbreak\code{colorlow!0}{\strut .}\allowbreak\code{colorlow!0}{\strut [EOS]}\allowbreak\newline\code{colorlow!0}{\strut J}\allowbreak\code{colorlow!0}{\strut N}\allowbreak\code{colorlow!20}{\strut P}\allowbreak\code{colorlow!20}{\strut and}\allowbreak\code{colorlow!50}{\strut P}\allowbreak\code{colorlow!0}{\strut are}\allowbreak\code{colorlow!20}{\strut B}\allowbreak\code{colorlow!20}{\strut V}\allowbreak\code{colorlow!0}{\strut for}\allowbreak\code{colorlow!0}{\strut V}\allowbreak\code{colorlow!0}{\strut J}\allowbreak\code{colorlow!50}{\strut J}\allowbreak\code{colorlow!0}{\strut N}\allowbreak\code{colorlow!0}{\strut of}\allowbreak\code{colorlow!70}{\strut each}\allowbreak\code{colorlow!20}{\strut N}\allowbreak\code{colorlow!0}{\strut .}\allowbreak\code{colorlow!0}{\strut [EOS]}\end{minipage}}\end{center}

This example, again from $\CorpusACL$, reveals the sequence \quotetxt{can be B V to}, which is only used by two other authors and the construction \quotetxt{the other N} at the beginning of a sentence.

%\begin{center}\rule{0.5\linewidth}{0.5pt}\end{center}
\begin{center}\fbox{\begin{minipage}{0.95\textwidth}\code{colorlow!50}{\strut this}\allowbreak\code{colorlow!50}{\strut N}\allowbreak\code{colorlow!0}{\strut V}\allowbreak\code{colorlow!0}{\strut B}\allowbreak\code{colorlow!20}{\strut J}\allowbreak\code{colorlow!0}{\strut N}\allowbreak\code{colorlow!0}{\strut ,}\allowbreak\code{colorlow!0}{\strut now}\allowbreak\code{colorlow!0}{\strut ,}\allowbreak\code{colorlow!0}{\strut they}\allowbreak\code{colorlow!20}{\strut V}\allowbreak\code{colorlow!0}{\strut like}\allowbreak\code{colorlow!0}{\strut P}\allowbreak\code{colorlow!0}{\strut so}\allowbreak\code{colorlow!50}{\strut i}\allowbreak\code{colorlow!0}{\strut ca}\allowbreak\code{colorlow!70}{\strut nt}\allowbreak\code{colorlow!0}{\strut say}\allowbreak\code{colorlow!0}{\strut they}\allowbreak\code{colorlow!20}{\strut are}\allowbreak\code{colorlow!0}{\strut all}\allowbreak\code{colorlow!0}{\strut J}\allowbreak\code{colorlow!20}{\strut but}\allowbreak\code{colorlow!0}{\strut the}\allowbreak\code{colorlow!0}{\strut J}\allowbreak\code{colorlow!0}{\strut few}\allowbreak\code{colorlow!0}{\strut i}\allowbreak\code{colorlow!50}{\strut tried}\allowbreak\code{colorlow!0}{\strut made}\allowbreak\code{colorlow!0}{\strut it}\allowbreak\code{colorlow!0}{\strut D}\allowbreak\code{colorlow!20}{\strut N}\allowbreak\code{colorlow!20}{\strut in}\allowbreak\code{colorlow!0}{\strut my}\allowbreak\code{colorlow!0}{\strut N}\allowbreak\code{colorlow!0}{\strut had}\allowbreak\code{colorlow!50}{\strut the}\allowbreak\code{colorlow!0}{\strut N}\allowbreak\code{colorlow!0}{\strut N}\allowbreak\code{colorlow!50}{\strut with}\allowbreak\code{colorlow!0}{\strut V}\allowbreak\code{colorlow!0}{\strut N}\allowbreak\code{colorlow!50}{\strut (}\allowbreak\code{colorlow!70}{\strut B}\allowbreak\code{colorlow!0}{\strut V}\allowbreak\code{colorlow!0}{\strut )}\allowbreak\code{colorlow!50}{\strut .}\allowbreak\code{colorlow!0}{\strut [EOS]}\allowbreak\newline\code{colorlow!0}{\strut D}\allowbreak\code{colorlow!20}{\strut N}\allowbreak\code{colorlow!0}{\strut D}\allowbreak\code{colorlow!0}{\strut )}\allowbreak\code{colorlow!0}{\strut J}\allowbreak\code{colorlow!20}{\strut P}\allowbreak\code{colorlow!20}{\strut in}\allowbreak\code{colorlow!0}{\strut J}\allowbreak\code{colorlow!0}{\strut N}\allowbreak\code{colorlow!70}{\strut -}\allowbreak\code{colorlow!0}{\strut D}\allowbreak\code{colorlow!20}{\strut N}\allowbreak\code{colorlow!0}{\strut ,}\allowbreak\code{colorlow!0}{\strut V}\allowbreak\code{colorlow!0}{\strut J}\allowbreak\code{colorlow!0}{\strut N}\allowbreak\code{colorlow!0}{\strut N}\allowbreak\code{colorlow!0}{\strut in}\allowbreak\code{colorlow!20}{\strut a}\allowbreak\code{colorlow!0}{\strut J}\allowbreak\code{colorlow!0}{\strut N}\allowbreak\code{colorlow!0}{\strut ,}\allowbreak\code{colorlow!0}{\strut over}\allowbreak\code{colorlow!0}{\strut V}\allowbreak\code{colorlow!20}{\strut N}\allowbreak\code{colorlow!0}{\strut D}\allowbreak\code{colorlow!0}{\strut )}\allowbreak\code{colorlow!0}{\strut P}\allowbreak\code{colorlow!0}{\strut P}\allowbreak\code{colorlow!0}{\strut over}\allowbreak\code{colorlow!0}{\strut N}\allowbreak\code{colorlow!50}{\strut -}\allowbreak\code{colorlow!0}{\strut D}\allowbreak\code{colorlow!20}{\strut N}\allowbreak\code{colorlow!0}{\strut ,}\allowbreak\code{colorlow!0}{\strut J}\allowbreak\code{colorlow!50}{\strut for}\allowbreak\code{colorlow!50}{\strut a}\allowbreak\code{colorlow!0}{\strut N}\allowbreak\code{colorlow!0}{\strut under}\allowbreak\code{colorlow!0}{\strut S}\allowbreak\code{colorlow!0}{\strut D}\allowbreak\code{colorlow!0}{\strut ,}\allowbreak\code{colorlow!0}{\strut comes}\allowbreak\code{colorlow!0}{\strut with}\allowbreak\code{colorlow!0}{\strut N}\allowbreak\code{colorlow!0}{\strut and}\allowbreak\code{colorlow!20}{\strut V}\allowbreak\code{colorlow!0}{\strut J}\allowbreak\code{colorlow!0}{\strut N}\allowbreak\code{colorlow!0}{\strut D}\allowbreak\code{colorlow!0}{\strut )}\allowbreak\code{colorlow!0}{\strut P}\allowbreak\code{colorlow!0}{\strut P}\allowbreak\code{colorlow!70}{\strut ,}\allowbreak\code{colorlow!0}{\strut V}\allowbreak\code{colorlow!0}{\strut N}\allowbreak\code{colorlow!50}{\strut -}\allowbreak\code{colorlow!0}{\strut J}\allowbreak\code{colorlow!20}{\strut but}\allowbreak\code{colorlow!0}{\strut a}\allowbreak\code{colorlow!0}{\strut N}\allowbreak\code{colorlow!20}{\strut J}\allowbreak\code{colorlow!50}{\strut for}\allowbreak\code{colorlow!50}{\strut a}\allowbreak\code{colorlow!0}{\strut N}\allowbreak\code{colorlow!0}{\strut plus}\allowbreak\code{colorlow!0}{\strut N}\allowbreak\code{colorlow!0}{\strut .}\allowbreak\code{colorlow!0}{\strut [EOS]}\allowbreak\newline\code{colorlow!0}{\strut 1}\allowbreak\code{colorlow!0}{\strut )}\allowbreak\code{colorlow!0}{\strut the}\allowbreak\code{colorlow!0}{\strut N}\allowbreak\code{colorlow!20}{\strut -}\allowbreak\code{colorlow!0}{\strut N}\allowbreak\code{colorlow!0}{\strut N}\allowbreak\code{colorlow!20}{\strut ,}\allowbreak\code{colorlow!20}{\strut N}\allowbreak\code{colorlow!0}{\strut N}\allowbreak\code{colorlow!50}{\strut ,}\allowbreak\code{colorlow!50}{\strut P}\allowbreak\code{colorlow!0}{\strut P}\allowbreak\code{colorlow!0}{\strut P}\allowbreak\code{colorlow!70}{\strut ,}\allowbreak\code{colorlow!0}{\strut and}\allowbreak\code{colorlow!0}{\strut a}\allowbreak\code{colorlow!0}{\strut N}\allowbreak\code{colorlow!0}{\strut for}\allowbreak\code{colorlow!0}{\strut less}\allowbreak\code{colorlow!0}{\strut than}\allowbreak\code{colorlow!0}{\strut D}\allowbreak\code{colorlow!20}{\strut N}\allowbreak\code{colorlow!0}{\strut .}\allowbreak\code{colorlow!0}{\strut [EOS]}\allowbreak\newline\code{colorlow!0}{\strut now}\allowbreak\code{colorlow!20}{\strut ,}\allowbreak\code{colorlow!0}{\strut the}\allowbreak\code{colorlow!0}{\strut N}\allowbreak\code{colorlow!0}{\strut set}\allowbreak\code{colorlow!0}{\strut is}\allowbreak\code{colorlow!0}{\strut J}\allowbreak\code{colorlow!20}{\strut ,}\allowbreak\code{colorlow!20}{\strut but}\allowbreak\code{colorlow!0}{\strut i}\allowbreak\code{colorlow!0}{\strut V}\allowbreak\code{colorlow!0}{\strut the}\allowbreak\code{colorlow!0}{\strut N}\allowbreak\code{colorlow!20}{\strut and}\allowbreak\code{colorlow!0}{\strut N}\allowbreak\code{colorlow!0}{\strut of}\allowbreak\code{colorlow!0}{\strut the}\allowbreak\code{colorlow!0}{\strut J}\allowbreak\code{colorlow!0}{\strut .}\allowbreak\code{colorlow!0}{\strut [EOS]}\allowbreak\newline\code{colorlow!50}{\strut but}\allowbreak\code{colorlow!0}{\strut definitely}\allowbreak\code{colorlow!0}{\strut J}\allowbreak\code{colorlow!0}{\strut !}\allowbreak\code{colorlow!0}{\strut [EOS]}\allowbreak\newline\code{colorlow!0}{\strut it}\allowbreak\code{colorlow!20}{\strut was}\allowbreak\code{colorlow!0}{\strut J}\allowbreak\code{colorlow!0}{\strut as}\allowbreak\code{colorlow!20}{\strut N}\allowbreak\code{colorlow!0}{\strut could}\allowbreak\code{colorlow!0}{\strut be}\allowbreak\code{colorlow!0}{\strut in}\allowbreak\code{colorlow!20}{\strut P}\allowbreak\code{colorlow!0}{\strut .}\allowbreak\code{colorlow!0}{\strut [EOS]}\end{minipage}}\end{center}

Finally, the last example is from $\CorpusYelp$, and it is a $\Dunk$ where the most obvious evidence is constituted by punctuation mark usage, such as dashes and commas, and by the contracted form of \quotetxt{can't} without an apostrophe. This token is marked as very important because the overarching sequence that identifies this author is the use of \quotetxt{can't} in combination with \quotetxt{so} with an intervening pronoun, a sequence only found in one other author in the corpus.

\section{Acknowledgments}
This research work was supported by the National Research Center for Applied Cybersecurity ATHENE. ATHENE is funded jointly by the German Federal Ministry of Education and Research and the Hessian Ministry of Higher Education, Research and the Arts. 

We thank Dmitry Nikolaev, Łukasz Dębowski, Petar Milin, Ricardo Bermúdez-Otero, and Richard Zimmermann for their useful comments and feedback.

\section{Declarations}

\paragraph{Ethical Approval}
Ethical approval was not required, since this research analyzed only existing, publicly available text corpora and involved no recruitment, intervention, or access to non-public personal data.

\paragraph{Informed Consent}
This article does not contain any studies with human participants performed by any of the authors.

\paragraph{Competing Interests}
The authors declare no competing interests.

\paragraph{Author Contributions}
The contributions of the authors (indicated by their initials) can be grouped as follows. 
\begin{itemize}[noitemsep]
    \item Conceptualization: AN
    \item Methodology: AN, OH, LG, VG
    \item Implementation: AN, LG, VG, ST, OH
    \item Validation: LG, VG
    \item Formal analysis: LG, VG, SI
    \item Investigation: AN, LG, OH
    \item Resources: OH, LG, ST
    \item Data Curation: OH, LG
    \item Writing - original draft: AN, OH, LG, VG
    \item Writing - review and editing: AN, OH, LG, ST, VG, SI
    \item Visualization: LG, OH
    \item Supervision: AN
    \item Project Administration: AN
\end{itemize}

\paragraph{Data Availability}
Some of the datasets analyzed during the current study are available at \url{https://github.com/AndreaNini/LambdaG}, together with the source code to reproduce the results. The missing datasets are available on reasonable request via email.

\clearpage

\printbibliography

\listoffigures
\end{document}